\documentclass{article}

\usepackage{microtype}
\usepackage{graphicx}
\usepackage{subcaption}
\usepackage{booktabs} 
\usepackage{enumitem} 
\usepackage{tikz}
\usetikzlibrary{calc,positioning,fit,backgrounds,arrows.meta}
\usepackage{multirow}
\usepackage[most]{tcolorbox}

\usepackage{hyperref}

\usepackage[preprint]{icml2026}

\usepackage{amsmath}
\usepackage{amssymb}
\usepackage{mathtools}
\usepackage{amsthm}
\usepackage{bm}

\newcommand{\probP}{\mathbb{P}}
\newcommand{\probQ}{\mathbb{Q}}

\usepackage[capitalize,noabbrev]{cleveref}

\theoremstyle{plain}
\newtheorem{theorem}{Theorem}[section]

\theoremstyle{definition}

\theoremstyle{remark}

\usepackage[textsize=tiny]{todonotes}

\icmltitlerunning{Kolmogorov-Arnold Energy Models}

\begin{document}

\twocolumn[
  \icmltitle{Kolmogorov-Arnold Energy Models: Fast, Interpretable Generative Modeling}



  \icmlsetsymbol{equal}{*}

  \begin{icmlauthorlist}
    \icmlauthor{Prithvi Raj}{equal,yyy}
  \end{icmlauthorlist}

  \icmlaffiliation{yyy}{}

  \icmlcorrespondingauthor{Prithvi Raj}{pr478@cantab.ac.uk}

  \icmlkeywords{Generative Models, Energy-Based Models, Latent Variable Models, Kolmogorov-Arnold Representation Theorem, Interpretability}

  \vskip 0.3in
]



\printAffiliationsAndNotice{}  

\begin{abstract}
Generative models typically rely on either simple latent priors (e.g., Variational Autoencoders, VAEs), which are efficient but limited, or expressive iterative samplers (e.g., Diffusion and Energy-based Models), which are costly and opaque. We introduce a new unsupervised model, the Kolmogorov-Arnold Energy Model (KAEM), to bridge this trade-off and provide new opportunities for interpretability. Based on a novel adaptation of the Kolmogorov-Arnold Representation Theorem, KAEM imposes a univariate latent prior, enabling fast and exact inference via the inverse transform method. On small datasets, we show that importance sampling becomes a tractable, unbiased, and single-pass posterior inference method. For settings requiring exploration, we propose a population-based strategy that decomposes the posterior into a sequence of annealed distributions, serving as a new remedy for poor mixing in Energy-based Models. KAEM attains competitive Fréchet Inception Distance among latent-prior models on SVHN, CIFAR10, and CelebA while sampling in a single forward pass at lower cost than iterative EBMs, and exposing an interpretable prior built from 1D densities.
\end{abstract}

\section{Introduction}

Latent variable models learn a data distribution by pairing a low-dimensional prior with a neural network decoder. The learning objective is to maximize the marginal likelihood of the observed data, which requires integrating over all latent configurations. Variational Autoencoders (VAEs) \citep{kingma2022autoencodingvariationalbayes} sidestep the intractable integral by optimizing a lower bound, incurring approximation and amortization gaps that bias the model away from the true maximum likelihood solution \citep{cremer2018inference}. Generative Adversarial Networks (GANs) \citep{goodfellow2014generativeadversarialnetworks} avoid the integral through adversarial training, producing sharp samples but forgoing any explicit likelihood and remaining prone to training instability. Both also restrict the latent prior to a fixed distribution such as an isotropic Gaussian, which cannot capture multimodal latent structure.

Latent energy-based models (EBMs) \citep{pang2020learninglatentspaceenergybased} address these limitations by replacing the fixed prior with a learned, data-dependent one and targeting the marginal likelihood directly. However, parameterizing the prior as a neural network renders it difficult to sample, requiring Langevin Monte Carlo (LMC) \citep{Brooks2011,Rossky1978,Roberts2002}. This is an iterative and gradient-based Markov chain Monte Carlo (MCMC) method, adding computational overhead and mixing poorly in the multimodal latent distributions that arise in generative modeling. Tuning compounds the difficulty: step sizes that are too small yield slow exploration and poor mixing, while larger ones accumulate discretization bias.

The neural prior is also opaque, offering no way to interpret what the model has learned about the latent structure or to embed domain knowledge into the prior, which is a missed opportunity. Latent EBMs learn the prior jointly with the generator by maximum likelihood, which can be viewed as an application of empirical Bayes in which the prior aligns with the posterior, converging on whatever latent structure best explains observations through the generator. If the prior were interpretable, this learned structure would be recoverable, offering a window into the low-dimensional manifolds that underlie the data \citep{Champion_2019} and a natural site for embedding domain knowledge or physical constraints to reduce data dependence and improve generalization, similar to the work of \citet{Karniadakis2021}.

Realizing this potential requires latent priors that (i) admit efficient inference, (ii) expose interpretable structure, and (iii) remain open to informed design.

We address these requirements by proposing the Kolmogorov-Arnold Energy Model (KAEM), which revises the latent EBM using a collection of univariate energy functions, drawing on Kolmogorov-Arnold Networks (KANs) \citep{liu2024kankolmogorovarnoldnetworks} and deep spline networks \citep{9264754}. The univariate structure enables interpretability alongside exact inference via the inverse transform method \citep{devroye1986nonuniform}, producing samples in a single forward pass rather than the iterative methods used for sampling EBMs and diffusion models \citep{ho2020denoisingdiffusionprobabilisticmodels}.

Since inverse transform is a single, static function applied to inputs in \( [0,1] \), after training, the entire prior sampling process can be replaced by 16-bit Look-Up Tables (LUTs), enabling fast FPGA acceleration at inference time. Example High-Level Synthesis (HLS) code is provided in Appendix~\ref{app:code}, demonstrating how low-latency generation enables real-time steering of samples toward higher or lower quantiles of the latent densities. By choosing the values of uniform noise serving as input to KAEM, users can tune generations with an almost instantaneous refresh rate.

By reinterpreting its inner functions as pushforwards between probability spaces, we also show that KAEM can be formed as a strict interpretation of the Kolmogorov-Arnold Representation Theorem (KART) \citep{Arnold2009}. Where standard neural architectures require manual choices of depth and width, KART determines the number and arrangement of univariate functions directly from latent dimensionality. On simple datasets, KAEM's low-dimensional latent space enables efficient training with importance sampling (IS) \citep{goertzel1949quota}, a technique typically avoided due to high variance under prior-posterior mismatch and the curse of dimensionality \citep{dupuis2004importance}, both of which KAEM can mitigate. Moreover, the trained priors are 1D densities, one per latent dimension, that can be plotted, interpreted, and potentially reused with other generators.

Unlike standard latent EBMs, which require iterative LMC for both prior and posterior sampling, KAEM confines LMC to posterior inference solely during training. Following \citet{pang2020learninglatentspaceenergybased}, we use the Unadjusted Langevin Algorithm (ULA) with a small iteration count, since non-convergent short-run MCMC has been shown to provide sufficient gradient signal for learning energy-based models \citep{nijkamp2019learningnonconvergentnonpersistentshortrun, nijkamp2019anatomymcmcbasedmaximumlikelihood}. As a result, KAEM targets the marginal likelihood directly, avoiding the approximation and amortization gaps of VAEs \citep{cremer2018inference}, and bypassing the need for an encoder, thus reducing storage and parameter overhead.

To improve mixing in multimodal posterior landscapes, we augment ULA with population-based sampling \citep{PT, Marinari1992, Hukushima1996} across a sequence of power posteriors \citep{2d186535-310b-3428-8c6f-e4ef298483d7} and a training criterion derived from Thermodynamic Integration \citep{CALDERHEAD20094028, Annis2019}. Previous works integrated diffusion to improve convergence and multimodal exploration in EBMs \citep{cui2024learninglatentspacehierarchical, zhang2023persistentlytraineddiffusionassistedenergybased}, but replacing the single prior with a chain of conditional models makes generation iterative and scatters the latent structure across denoising steps. Our thermodynamic criterion is only applied during training and scales in parallel over power posteriors, preserving interpretable structure and inference speed.

We evaluate KAEM against a VAE, the neural-prior latent EBM of \citet{pang2020learninglatentspaceenergybased}, and a denoising diffusion probabilistic model (DDPM) \citep{ho2020denoisingdiffusionprobabilisticmodels}. With regards to image fidelity metrics, KAEM outperformed the VAE on every dataset, while remaining competitive against the neural EBM and outperforming it by a wide margin on SVHN. The DDPM led SVHN and CIFAR10 operating in pixel space and sampling with $700 \times$ greater inference cost. KAEM also sampled at $1.4\times$ lower cost than the neural latent EBM, and is the only model whose prior decomposes into per-dimension densities that can be inspected. We present KAEM as an initial step toward further adaptations of KART in machine learning, with the far-reaching hope of advancing toward a broader conceptualization: \emph{``The Kolmogorov-Arnold Representation Theorem Is All You Need.''}

\begin{figure*}[t]
\centering
\begin{tikzpicture}[
    font=\scriptsize,
    box/.style={
        draw=gray!65,
        fill=white,
        rounded corners=2pt,
        minimum height=7mm,
        minimum width=2.0cm,
        align=center,
        line width=0.5pt
    },
    flow/.style={
        -Stealth,
        line width=0.65pt,
        draw=gray!80
    },
    update/.style={
        dashed,
        -Stealth,
        line width=0.75pt,
        draw=gray!75
    },
    info/.style={
        dotted,
        -Stealth,
        line width=0.6pt
    },
    sectionlabel/.style={
        font=\bfseries\scriptsize,
        anchor=north west
    }
]

\node[box, fill=red!18] (data) at (0,0) {Data\\$\mathbf{x}$};
\node[box, fill=blue!22] (post) at (2.8,0) {Posterior\\$p(\mathbf z\mid\mathbf x)$};
\node[box, fill=blue!32] (samples) at (5.8,0) {Posterior\\$\mathbf z$};

\node[box, fill=blue!38, minimum width=2.6cm] (gradprior) at (9.8,0.8)
{$\nabla_f \log p_f(\mathbf z)$};

\node[box, fill=purple!45, minimum width=2.8cm] (gradgen) at (9.8,-0.8)
{$\nabla_\Phi \log p_\Phi(\mathbf x\mid\mathbf z)$};

\node[box, fill=gray!10] (uniform) at (7.4,-5.9)
{$\mathbf u\sim\mathrm{Unif}(0,1)^Q$};

\node[
    draw=blue!55,
    fill=blue!8,
    rounded corners=3pt,
    inner sep=2pt,
    minimum width=3.8cm,
    minimum height=4.0cm
] (prior) at (3.6,-4.5) {};

\node[
    font=\bfseries\tiny,
    text=blue!90!black
] at ([yshift=16mm]prior.center)
{KAEM Prior};

\node at ([yshift=2mm]prior.center)
{
    \includegraphics[
        width=2.6cm,
        keepaspectratio
    ]{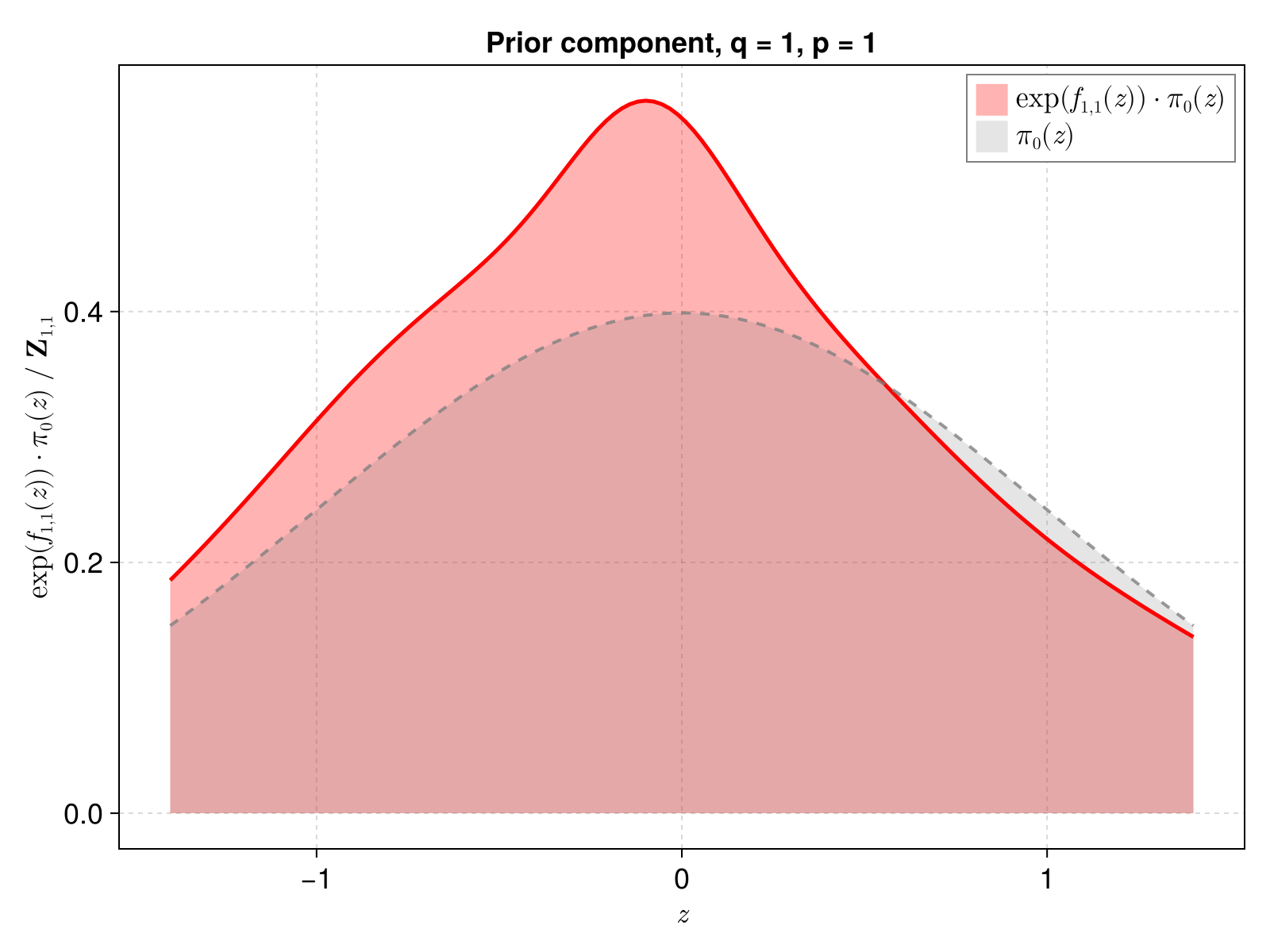}
};

\node[
    font=\tiny
] at ([yshift=-16mm]prior.center)
{
    $\displaystyle
    p_{q,p}(z) = \frac{\exp(f_{q,p}(z))}{Z_{q,p}} \pi_0(z),
    $
};

\node[box, fill=blue!28] (latent) at (7.4,-4.5) {Prior\\$\mathbf z$};
\node[box, fill=purple!42] (gen) at (9.8,-4.5) {Generator\\$G_\Phi$};
\node[box, fill=red!18] (output) at (12.4,-4.5) {Sample\\$\tilde{\mathbf x}$};

\begin{scope}[on background layer]
    \node[
        draw=gray!25,
        fill=blue!5,
        rounded corners=4pt,
        inner sep=4pt,
        fit={(data) (post) (samples) (gradprior) (gradgen) (12.4, 1.4)}
    ] (trainpanel) {};
    
    \node[
        draw=gray!25,
        fill=blue!3,
        rounded corners=4pt,
        inner sep=4pt,
        fit={(uniform) (prior) (latent) (gen) (output) (0, -6.5)}
    ] (genpanel) {};

    \node[sectionlabel] at ([xshift=2mm,yshift=-2mm]trainpanel.north west) {Training};
    \node[sectionlabel] at ([xshift=2mm,yshift=-2mm]genpanel.north west) {Generation};
\end{scope}

\draw[flow] (data) -- (post);
\draw[flow] (post) -- node[midway, above, yshift=3.0mm, font=\tiny]
{IS / ULA / Thermo} (samples);

\draw[flow] (samples.east) -- ++(1.0,0) |- ([yshift=2mm]gradprior.west);
\draw[flow] (samples.east) -- ++(1.0,0) |- ([yshift=2mm]gradgen.west);

\draw[flow] (prior) -- node[above,font=\tiny] {ITS} (latent);

\draw[flow] (latent) -- (gen);
\draw[flow] (gen) -- (output);

\draw[flow]
    (uniform.west)
    -- ++(-0.3,0)
    -| ($(prior.east)!0.5!(latent.west)$);

\draw[info, blue!85!black] (latent.north) |- ([yshift=-2mm]gradprior.west);
\draw[info, red!80!black] (data.south) |- ([yshift=-2mm]gradgen.west);
\draw[info, purple!85!black] (output.north) |- (gradgen.east);

\draw[update, blue!85!black] (gradprior.east) -- ++(2.9,0)
|- ([yshift=-4mm]genpanel.south)
-| node[pos=0.25, below, font=\tiny] {Update $f$} (prior.south);

\draw[update, purple!85!black] (gradgen.south) -- (gen.north)
node[pos=0.5, left, font=\tiny] {Update $\Phi$};

\end{tikzpicture}
\caption{
\textbf{KAEM architecture overview.}
\emph{Training (top):} posterior samples are obtained using importance sampling (IS), unadjusted Langevin dynamics (ULA), or thermodynamic integration, and used to update both the KAEM prior parameters $f$ and generator parameters $\Phi$.
\emph{Generation (bottom):} latent variables are sampled exactly from the learned KAEM prior via inverse transform sampling (ITS), then mapped to data space through the generator.
}
\label{fig:kaem_architecture}
\end{figure*}

\section{Preliminaries}
\label{sec:prelims}

\paragraph{The Kolmogorov-Arnold Representation Theorem (KART).} Presented by \citet{Arnold2009}, this forms the basis of KAEM. A structured proof of the theorem has been provided by \citet{dzhenzher2022structuredproofkolmogorovssuperposition}. 
\begin{theorem}
For any integer $n_z > 1$, there are continuous univariate functions \( \Phi_q: \mathbb{R} \to \mathbb{R} \) and \( \psi_{q,p} : [0,1] \to \mathbb{R} \), such that any continuous multivariate function, $g: [0, 1]^{n_z} \to \mathbb{R}$, can be represented as a superposition of those functions:
\begin{equation}
g(u_1, \dots, u_{n_z}) = \sum_{q=1}^{2n_z+1} \Phi_q \left( \sum_{p=1}^{n_z} \psi_{q,p}(u_p) \right),
\label{eq:KAT}
\end{equation}
\end{theorem}

\paragraph{Latent variable generative models.}
Let $\boldsymbol{x}\in\mathcal{X}$ denote observed data vectors and $\boldsymbol{z}\in\mathcal{Z}$ latent vectors.
A latent variable generative model specifies
\begin{equation}
p_{\Phi,f}(\boldsymbol{x},\boldsymbol{z})
\;=\;
p_\Phi(\boldsymbol{x}\mid \boldsymbol{z})\,p_f(\boldsymbol{z}),
\label{eq:joint}
\end{equation}
where $p_\Phi(\boldsymbol{x}\mid \boldsymbol{z})$ is a neural generator (decoder) and $p_f(\boldsymbol{z})$ is a latent prior parameterized by $f$.
The log-marginal likelihood is the learning objective for Maximum Likelihood Estimation (MLE):
\begin{align}
\log p_{\Phi,f}(\boldsymbol{x})
&\;=\;
\int \log p_{\Phi,f}(\boldsymbol{x}, \boldsymbol{z})\,p_{\Phi,f}(\boldsymbol{z}\mid\boldsymbol{x})\,d\boldsymbol{z} \notag \\
&\;=\; \mathbb{E}_{p_{\Phi,f}(\boldsymbol{z}\mid\boldsymbol{x})}\left[ \log p_{\Phi,f}(\boldsymbol{x}, \boldsymbol{z}) \right]
\label{eq:marginal}
\end{align}

VAEs circumvent the intractability of Eq.~\ref{eq:marginal} by instead maximizing the evidence lower bound (ELBO). This incurs both an approximation and an amortization gap \citep{cremer2018inference}; the true maximum likelihood solution is not the optimization target. We instead estimate the marginal likelihood directly, using importance sampling or ULA for posterior inference, with thermodynamic integration to improve the quality of the estimate (Sec.~\ref{sec:sampling}).

In Appendix~\ref{sec:MLE-lr}, we prove that the learning gradient does not depend on the sampling distribution:
\begin{align}
    \nabla_{f, \Phi} 
    \left[ \log p_{f, \Phi} (\boldsymbol{x}) \right] 
    \;=\; &\mathbb{E}_{p_{f, \Phi}(\boldsymbol{z} \mid \boldsymbol{x})} \Bigg[
        \nabla_{f} 
        \left[ \log p_{f}\left(\boldsymbol{z}\right) \right] \Bigg] \notag \\ \;+\; 
        &\mathbb{E}_{p_{f, \Phi}(\boldsymbol{z} \mid \boldsymbol{x})} \Bigg[
        \nabla_{\Phi } 
        \left[ \log p_{ \Phi } \left(\boldsymbol{x} \mid \boldsymbol{z}\right) \right]
    \Bigg].
    \label{eq:learning-grad}
\end{align} 

\paragraph{Monte Carlo estimator}

To approximate arbitrary integrals such as the expectation in Eq.~\ref{eq:learning-grad}, one can draw $N$ samples from the target distribution and take an unbiased Monte Carlo estimator. For some arbitrary function, $\rho$, (e.g., the log-likelihood), the expectation with respect to an arbitrary density, $p$, can be estimated as: 

\begin{align}
    \mathbb{E}_{p(z)}[\rho(z)] \;=\; \int_{\mathcal{Z}}\rho(z) \; p(z) \; dz \; &\approx \; \frac{1}{N} \sum_{s=1}^{N} \rho(z^{(s)}), \notag \\ \text{where} \quad \{z^{(s)}\}_{1:N} &\; \sim \; p(z).
    \label{eq:mc-estimate}
\end{align}

\paragraph{Top-down generator}
The latent-to-data generative process is realized using a neural network $G_\Phi:\mathcal{Z}\to\mathcal{X}$, (parameterized by $\Phi$),
which maps a latent sample $z$ to a reconstruction (mean image or data sample)
\begin{equation}
\tilde{\boldsymbol{x}} \;=\; \mu_\Phi(\boldsymbol{z}) \;=\; G_\Phi(\boldsymbol{z}) + \boldsymbol{\varepsilon} ,
\qquad
\boldsymbol{\varepsilon} \sim \mathcal{N}(\mathbf{0},\sigma_{\text{noise}}\mathbf{I}),
\label{eq:decoder_mean}
\end{equation}
In this work we model the conditional likelihood with an isotropic Gaussian observation model,
equivalent to a pixel-wise $\ell_2$ reconstruction objective:
\begin{equation}
p_\Phi(\boldsymbol{x}\mid \boldsymbol{z})
\;=\;
\mathcal{N}\!\big(\boldsymbol{x};\,\mu_\Phi(\boldsymbol{z}),\,\sigma_x^2 \mathbf{I}\big),
\label{eq:gaussian_likelihood}
\end{equation}
where $\sigma_x^2$ is a specified variance, $\sigma_{\text{noise}}^2$ captures observation noise, and $\mathbf{I}$ is the identity matrix over pixels and channels.
For an image $x\in\mathbb{R}^{H\times W\times C}$, the corresponding log-likelihood is
\begin{equation}
\log p_\Phi(\boldsymbol{x}\mid \boldsymbol{z})
\;\propto\;
-\frac{1}{2\sigma_x^2}\,\|\boldsymbol{x}-\mu_\Phi(\boldsymbol{z})\|_2^2
\label{eq:gaussian_loglik}
\end{equation}
Up to an additive constant, maximizing $\log p_\Phi(\boldsymbol{x}\mid \boldsymbol{z})$ is equivalent to minimizing the pixel-wise squared error. The gradient needed for the log-likelihood component of Eq.~\ref{eq:learning-grad} can then be found by autodifferentiation, which we used Enzyme \citep{moses2020enzyme} for.

\paragraph{Latent energy-based prior.}
Latent EBMs \citep{pang2020learninglatentspaceenergybased} define the prior by an energy $E_f$ and base reference prior, e.g., $\pi_0=\mathcal{N}(0,1)$:
\begin{equation}
p_f(\boldsymbol{z})\propto \exp(E_f(z))\cdot\pi_0(z).
\label{eq:ebmprior}
\end{equation}
Training requires sampling from the posterior
\begin{equation}
p_{\Phi,f}(\boldsymbol{z}\mid \boldsymbol{x})
\propto p_\Phi(\boldsymbol{x}\mid \boldsymbol{z})\,p_f(\boldsymbol{z}),
\label{eq:posterior}
\end{equation}
which is often multimodal and difficult to sample from.

\paragraph{Unadjusted Langevin Algorithm (ULA).}
Following \citet{pang2020learninglatentspaceenergybased}, we use ULA \citep{Roberts2002, Rossky1978} for posterior sampling. Given a target $\pi(\boldsymbol{z})$, ULA generates proposals via a gradient-informed random walk:
\begin{equation}
\boldsymbol{z}^{i+1} = \boldsymbol{z}^{i} + \eta\,\nabla_{\boldsymbol{z}} \log \pi(\boldsymbol{z}^{i}) + \sqrt{2\eta}\,\boldsymbol{\xi}^{i}, \quad \boldsymbol{\xi}^{i} \sim \mathcal{N}(0,I),
\label{eq:ula}
\end{equation}
with step size $\eta > 0$. In latent EBMs, $\pi$ is either the prior~\eqref{eq:exp-tilt}, posterior~\eqref{eq:posterior} or power posterior~\eqref{eq:powerposterior}. ULA incurs $O(\eta)$ discretization bias but avoids the computational overhead of Metropolis-Hastings correction, requiring only a single gradient evaluation per step. In KAEM's low-dimensional latent space, this trade-off is favorable because the bias is small for moderate step sizes and the reduced cost per step allows more iterations within a fixed compute budget. Following \citet{pang2020learninglatentspaceenergybased}, we adopt a small iteration count, (40 iterations at 0.01 step for posterior sampling), following the guidance of \citet{nijkamp2019learningnonconvergentnonpersistentshortrun, nijkamp2019anatomymcmcbasedmaximumlikelihood} regarding short-run MCMC.

\paragraph{Power posteriors.}
To improve mixing, KAEM samples from \emph{power posteriors} \citep{2d186535-310b-3428-8c6f-e4ef298483d7}:
\begin{equation}
p_{\Phi,f,t}(\boldsymbol{z}\mid \boldsymbol{x})
\propto
p_\Phi(\boldsymbol{x}\mid \boldsymbol{z})^t\,p_f(\boldsymbol{z}),
\qquad
t\in[0,1],
\label{eq:powerposterior}
\end{equation}
where $t=0$ recovers the prior and $t=1$ recovers the posterior.
A population of chains at $\{t_k\}_{k=0}^K$ with swap moves (parallel tempering) improves exploration.
Thermodynamic Integration connects expectations under $p_{t_k}$ to log-normalizer differences; details are deferred to Sec.~\ref{sec:sampling}.

\section{Method}
\label{sec:method}

\subsection{Architecture}
\label{sec:arch}

\paragraph{KAEM prior.} The inner functions, $\psi_{q,p}$, of Theorem~\ref{eq:KAT} can be conveniently translated to the framework of inverse transform sampling (ITS) \citep{devroye1986nonuniform}. We interpret $\psi_{q,p}$ as the application of an inverse cumulative density function (CDF) to a uniform random variable to recover a sample from a univariate latent density, $p_{q,p}$: 
\begin{equation}
z = \psi_{q,p}(u_p) = F^{-1}_{\pi_{q,p}}(u_p) \quad \Leftrightarrow \quad z \sim p_{q,p}(z).
\label{eq:inv-cdf}
\end{equation}
Here, \( F^{-1}_{\pi_{q,p}} : [0,1] \to \mathcal{Z} \) is the generalized inverse CDF associated with a trained measure, $\pi_{q,p}$ in the latent space. Since $F^{-1}_{\pi_{q,p}}$ is a measurable, monotone, increasing function, it can be viewed as a measurable transformation between two probability spaces and does not invalidate the deterministic structure of KART. We formalize this transformation in Appendix~\ref{sec:measureable}, and show in Appendix~\ref{sec:flow-connection} that ITS computes a normalizing flow~\citep{pmlr-v37-rezende15} over $[0,1]^Q$. We adapt the formulation of \citet{pang2020learninglatentspaceenergybased} to the univariate case, and parameterize the latent density via exponential tilting of a base prior $\pi_0$:
\begin{equation}
p_{q,p}(z) = \frac{\exp(f_{q,p}(z))}{Z_{q,p}} \pi_0(z),
\label{eq:exp-tilt}
\end{equation}
where $f_{q,p}(z)$ is a learned energy function and $Z_{q,p}$ is the finite normalizing constant (partition function).

We parameterize the functions $f_{q,p}$ using Radial Basis Functions (RBFs) \citep{li2024kolmogorovarnoldnetworksradialbasis}. Since fixed-grid RBF KANs are defined over a predetermined domain, we adopt the periodic least-squares grid updating of \citet{liu2024kankolmogorovarnoldnetworks} to adapt their support as training progresses. For ULA-based training, where posterior samples may extend far beyond the initial domain, we prespecify a larger ITS domain and use the Wavelet-KAN parameterization of \citet{bozorgasl2024wavkanwaveletkolmogorovarnoldnetworks}. In this case, the centers are treated as learnable parameters, eliminating the need for least-squares.

Given the independence of the prior's parameters, the log-prior can be modeled as a sum over the univariate cases:
\begin{align}
\log p_{f} \left(\boldsymbol{z}\right) &\;=\; \sum_{q=1}^{Q} \sum_{p=1}^{P} \log p_{q,p}(z_{q,p}) \notag \\ &\;=\; \sum_{q=1}^{Q} \sum_{p=1}^{P} f_{q,p}(z_{q,p}) + \log \pi_0(z_{q,p}) - \log Z_{q,p},
\label{eq:log-prior}
\end{align}
where $P=n_z,Q=2n_z+1$ are taken throughout the study based on the structure of KART in Eq.~\ref{eq:KAT}. Following the approach of \citet{pang2020learninglatentspaceenergybased} we adopt the contrastive divergence (CD) criterion to train the univariate prior. This is derived under Appendix~\ref{app:prior-grad}:
\begin{align}
    &\mathbb{E}_{p_{f, \Phi}(\boldsymbol{z} \mid \boldsymbol{x})} \Bigg[
        \nabla_{f} 
        \left[ \log p_{f}\left(\boldsymbol{z}\right) \right] 
    \Bigg] \notag \\ =\; &\mathbb{E}_{p_{f, \Phi}(\boldsymbol{z} \mid \boldsymbol{x})} \Bigg[ \nabla_{f} \sum_{q=1}^{2n_z+1} \sum_{p=1}^{n_z} f_{q,p}(z) \Bigg] \notag \\ -\; &\mathbb{E}_{p_{f}(\boldsymbol{z})} 
    \Bigg[ \nabla_{f} \sum_{q=1}^{2n_z+1} \sum_{p=1}^{n_z} f_{q,p}(z) \Bigg].
    \label{eq:simplified-log-prior-grad}
\end{align}

We approximate these expectations with Monte Carlo estimators after first drawing samples from the prior using ITS, and from the posterior using ULA or population-based sampling. Alternatively, posterior sampling can be circumvented entirely using the importance sampling (IS) estimator. Similar to the log-likelihood in Eq.~\ref{eq:gaussian_loglik}, the gradients can be found with autodifferentiation.

\paragraph{KAEM mixture-of-univariate prior.} The form in Eq.~\ref{eq:exp-tilt} ignores inter-dimensional dependencies in the latent space and results in an axis-aligned joint distribution, but KAEM remains flexible to other parameterizations.
For efficiency and scalability, we can also parameterize the latent as $\boldsymbol{z}\in\mathbb{R}^{Q}$ (we write $\boldsymbol{z}_q$ for the $q$-th coordinate) and define a mixture prior per coordinate, where $f_{q,p}$ is realized as univariate KAN or spline functions.
\begin{equation}
p_f(\boldsymbol{z})
=
\prod_{q=1}^Q p_{f,q}(\boldsymbol{z}_q),
\qquad
p_{f,q}(z)
=
\sum_{p=1}^P \alpha_{q,p}\,p_{q,p}(z),
\label{eq:kaem_mix}
\end{equation}
where $\alpha_{q,p}\ge 0$ are mixture proportions satisfying $\sum_p\alpha_{q,p}=1$. They could be learned as unnormalized logits $\theta_{q,p}$ passed through a softmax, $\alpha_{q,p}=\exp(\theta_{q,p})/\sum_{p'}\exp(\theta_{q,p'})$, to enforce the non-negativity and unit-sum constraints. However, in this study, they are fixed uniformly as $1/n_z$, guided by \cite{rivera2024trainvae}.

The joint remains an axis-aligned product: each $\boldsymbol{z}_q$ is drawn independently from a multimodal 1D mixture. The mixture broadens per-coordinate expressivity but does not introduce inter-dimensional correlations, which are deferred to the generator, as in the standard VAE setup \citep{kingma2022autoencodingvariationalbayes}. Sampling becomes more efficient than in the previous formulation \ref{eq:exp-tilt}, as it proceeds component-wise using ITS for discrete variables, \citep{DEVROYE200683}. However, this violates KART, since $F^{-1}_{\pi_{q,p}}$ now corresponds to a discrete rather than a continuous CDF.

A CD learning gradient for the mixture prior can be formulated by first viewing the energy as a log-sum-exp coupling: 
\begin{align}
\log p_f(\boldsymbol{z})
    &= \sum_{q=1}^Q f_{\text{mix}} - \log \bm{Z_q} \notag \\
    &= \sum_{q=1}^Q \mathrm{lse}_{p}
    \Big(\log \alpha_{q,p} + f_{q,p}(z_q) + \log \pi_0(z_q) \Big) \notag \\
    &- \log \bm{Z_{q}},
\label{eq:logprior_mix}
\end{align}

where $\mathrm{lse}$ denotes the log-sum-exp function and $\bm{Z_{q}}$ is a multivariate partition that can be avoided in training via the same contrastive divergence framework derived for the univariate prior in Appendix~\ref{app:prior-grad}:
\begin{align}
    \mathbb{E}_{p_{f, \Phi}(\boldsymbol{z} \mid \boldsymbol{x})} &\Bigg[
        \nabla_{f} 
        \left[ \log p_{f}\left(\boldsymbol{z}\right) \right]
    \Bigg] \;= \notag \\ \mathbb{E}_{p_{f, \Phi}(\boldsymbol{z} \mid \boldsymbol{x})} \Bigg[ \nabla_{f} \sum_{q=1}^{Q} f_{\text{mix}}(z) \Bigg] &- \mathbb{E}_{p_{f}(\boldsymbol{z})} 
\Bigg[ \nabla_{f} \sum_{q=1}^{Q} f_{\text{mix}}(z) \Bigg].
\label{eq:mix-log-prior-grad}
\end{align}

\paragraph{Generator} Having sampled the latent variable such that $z_{q,p} = F^{-1}_{\pi_{q,p}}(u_p)$, generation proceeds via Eq.~\ref{eq:decoder_mean}, where $G_\Phi:\mathcal{Z}\to\mathcal{X}$ is a neural network parameterized by $\Phi$:
\begin{equation}
G_\Phi(\boldsymbol{z}), \quad \text{where} \quad\boldsymbol{z}
\;=\;
\{z_{q,p}\}_{1:Q,1:P}
\end{equation}

In this study, we primarily use convolutional neural networks (CNNs) \citep{LeCun2015Deep} due to their spatial inductive bias: sequences of convolutional layers implement shift-equivariant transformations that are well-suited for images. However, we also consider using Kolmogorov-Arnold Networks (KANs) for simple datasets and strict adherence to KART (Eq.~\ref{eq:KAT}). This is discussed under Appendix~\ref{app:gen_adherent}. 

The log-likelihood in Eq.~\ref{eq:gaussian_loglik} can be differentiated to train the generator, as per Eq.~\ref{eq:learning-grad} for the posterior expectations. These expectations can be approximated either with the standard Monte Carlo estimator in Eq.~\ref{eq:mc-estimate}, or with the importance sampling (IS) estimator, which circumvents the need to sample from the posterior.

\subsection{Sampling}
\label{sec:sampling}

\paragraph{Normalization and inverse transform.} Since $Z_{q,p}$ is one-dimensional, it can be computed efficiently by Gauss-Kronrod quadrature, \citep{gauss-quad}:
\begin{equation}
Z_{q,p} = \int_{z_{\text{min}}}^{z_{\text{max}}} \exp(f_{q,p}(z)) \pi_0(z) dz \approx \sum_{i=1}^{N_{\text{quad}}} w_i \, H(z_i^{\text{node}}), 
\label{eq:normalization}
\end{equation}
where $H(z) = \exp(f_{q,p}(z)) \pi_0(z)$, and $\{z_i^{\text{node}}, w_i\}_{i=1}^{N_{\text{quad}}}$ are the quadrature nodes and weights for the interval $\left[z_{\text{min}}, z_{\text{max}}\right]$. The availability of $Z_{q,p}$ enables inverse transform sampling (ITS) from the prior, and the smoothness of $H$, together with its bounded support, makes this approximation accurate. 

The multivariate latent product density, (Eq.~\ref{eq:kaem_mix}), is exact-sampleable since each $p_{q,p}$ is univariate and normalized. Sampling from the mixture proceeds by selecting $p^\star\sim\mathrm{Cat}(\alpha_{q,1:P})$ and then sampling $z_q\sim p_{q,p^\star}$ via ITS. Algorithm~\ref{alg:its} summarizes the procedure. To sample from Eq.~\ref{eq:exp-tilt}, one simply has to skip component-selection and sample from all densities, $p_{1:Q,1:P}$.

\begin{algorithm}[t]
\caption{Exact sampling from KAEM mixture prior by inverse transform sampling (ITS)}
\label{alg:its}
\begin{algorithmic}[1]
\REQUIRE Weights $\alpha_{q,1:P}$, energies $f_{q,1:P}$,
base density $\pi_0$,
Gauss-Legendre nodes and weights $\{z_i,w_i\}_{i=1}^{N}$.
\ENSURE Sample $z\in\mathbb{R}^Q$ from $p_f(\boldsymbol{z})$.
\FOR{$q=1$ to $Q$}
    \STATE \textbf{Sample component} $p^\star\sim\mathrm{Categorical}(\alpha_{q,1:P})$.
    \STATE \textbf{Find} $Z_{q,p}\leftarrow \sum_{i=1}^{N} w_i
    \exp(f_{q,p^\star}(z_i))\pi_0(z_i)$.
    \STATE \textbf{CDF table}:
    $C_0\leftarrow 0$ and
    $C_i\leftarrow C_{i-1} + w_i
    \exp(f_{q,p^\star}(z_i))\pi_0(z_i)$.
    \STATE \textbf{Normalize:} $C_i \leftarrow C_i/Z_{q,p} \quad \forall i$.
    \STATE \textbf{Sample} $u\sim\mathrm{Unif}(0,1)$.
    \STATE \textbf{Find bin:} Smallest $j$ such that $C_j\ge u$.
    \STATE \textbf{Interpolate:}
    $t\leftarrow \frac{u-C_{j-1}}{C_j-C_{j-1}}$,
    $z_q\leftarrow (1-t)z_{j-1} + tz_j$.
\ENDFOR
\STATE \textbf{return} $z$.
\end{algorithmic}
\end{algorithm}

\paragraph{Importance sampling (IS)} We can estimate expectations with respect to
$p_{f,\Phi}(\boldsymbol{z} \mid \boldsymbol{x})$
using importance sampling (IS) \citep{910e8e7d-2e70-317c-adf6-e21cd50755a1} with the prior as proposal. Algorithm~\ref{alg:importance_sampling} summarizes the procedure. More details are provided under Appendix~\ref{sec:sampling-theory}, including theoretical depth regarding the importance weights, and a summary of residual resampling, which we adopt for variance reduction and to focus learning on samples that are more representative of the posterior distribution.

IS is used to estimate the log-marginal likelihood and its gradient in Eq.~\ref{eq:learning-grad}. However, unlike ULA (Eq.~\ref{eq:ula}), IS fails when the prior is poorly aligned with the posterior, making it unsuitable for complex datasets such as RGB images. In cases where IS fails, we adopt ULA to sample from $p_{f,\Phi}(\boldsymbol{z} \mid \boldsymbol{x})$ and estimate the posterior expectation in Eq.~\ref{eq:marginal} using the standard Monte Carlo estimator in Eq.~\ref{eq:mc-estimate}.

\begin{algorithm}[t]
\caption{Posterior expectation via importance sampling}
\label{alg:importance_sampling}
\begin{algorithmic}[1]
\REQUIRE Observation $\boldsymbol{x}^{(s)}$, prior $p_{f}(\boldsymbol{z})$,
likelihood $p_{\Phi}(\boldsymbol{x} \mid \boldsymbol{z})$,
number of samples $N$,
ESS threshold $\gamma\in(0,1)$.
\ENSURE Approximation of
$\mathbb{E}_{p_{f,\Phi}(\boldsymbol{z} \mid \boldsymbol{x})}[\rho(\boldsymbol{z})]$.

\STATE \textbf{Sample proposal:}
Draw $\{\boldsymbol{z}^{(s)}\}_{1:N}\sim p_{f}(\boldsymbol{z})$ using inverse transform sampling (Algorithm~\ref{alg:its}).

\STATE \textbf{Find importance weights using the likelihood:}
\begin{equation*}
w^{(s)} \leftarrow p_{\Phi}(\boldsymbol{x}\mid \boldsymbol{z}^{(s)}),
\qquad
w_{\mathrm{norm}}^{(s)} \leftarrow
\frac{w^{(s)}}{\sum_{r=1}^{N} w^{(r)}}
\end{equation*}
implemented stably via a softmax over $\log w^{(s)}$.

\STATE \textbf{Effective sample size (ESS):}
\begin{equation*}
ESS \leftarrow \frac{1}{\sum_{1:N} (w_{\mathrm{norm}}^{(s)})^2}.
\end{equation*}

\IF{$ESS < \gamma\,N$}
    \STATE \textbf{Resample:}
    Resample $\{\boldsymbol{z}^{\prime \; (s)}\}$ according to $w_{\mathrm{norm}}^{(s)}$
    (we use residual resampling; Sec.~\ref{sec:residual-resampling}).
    \STATE Set $w_{\mathrm{norm}}^{(s)} \leftarrow 1/N$.
\ENDIF

\STATE \textbf{Estimate expectation:}
\begin{equation*}
\mathbb{E}_{p_{f,\Phi}}[\rho(\boldsymbol{z})]
\;\approx\;
\sum_{s=1}^N w_{\mathrm{norm}}^{(s)}\,\rho(\boldsymbol{z}^{(s)}) \;\approx\;
\frac{1}{N}\sum_{s=1}^N\rho(\boldsymbol{z}^{\prime \; (s)}).
\end{equation*}

\STATE \textbf{return} estimated expectation.
\end{algorithmic}
\end{algorithm}

\paragraph{Thermodynamic criterion} To mitigate the poor convergence of EBMs with multimodal posterior landscapes, we introduce a training criterion that facilitates posterior annealing. Details on the temperature schedule adopted in this study, along with practical guidance on hyperparameter selection, are provided in Sec.~\ref{sec:thermo-help}.

We inform our derivations using the Thermodynamic Integral, derived in Sec.~\ref{sec:TI-derived}. In place of the log-marginal likelihood of Eq.~\ref{eq:marginal}, this uses the power posterior in Eq.~\ref{eq:powerposterior}:
\begin{equation}
\log \left( p_{f, \Phi} (\boldsymbol{x}) \right) = \int_0^1
\mathbb{E}_{p_{f, \Phi}(\boldsymbol{z} \mid \boldsymbol{x},t)}\left[ \log \left( p_{\Phi} \left(\boldsymbol{x}\mid\boldsymbol{z} \right) \right) \right] \, dt
\label{eq:thermo}
\end{equation}

The exact calculation of the integral in Eq.~\ref{eq:thermo} is feasible given that the bounds are \( t=0 \), corresponding to the expectation being taken with respect to the prior, and \( t=1 \), where it is taken with respect to the posterior. However, the integral can also be expressed exactly without collapsing the temperatures. This allows greater flexibility in how the integral evolves through the temperature schedule, $\{t_k\}_{k=0}^{N_t}$. In particular, the temperature schedule is discretized as:
\begin{equation}
    \boldsymbol{t}=\{ t_0, t_1, \ldots t_{N_{\text{t}}} \}, \quad \quad t_k \in [0,1]
    \label{eq:tempsarray}
\end{equation}

Following \citet{CALDERHEAD20094028}, we obtain the training objective via trapezoidal discretization:
\begin{align}
    \log p_{f, \Phi}(\boldsymbol{x}) \approx \frac{1}{2} \sum_{k=1}^{N_t} \Delta t_k \left( E_{k-1} + E_k \right),
    \label{eq:SS-KAEM}
\end{align}
where $E_k = \mathbb{E}_{p_{f,\Phi}(\boldsymbol{z} \mid \boldsymbol{x}, t_k)}\left[ \log p_\Phi(\boldsymbol{x} \mid \boldsymbol{z}) \right]$, estimated via Monte Carlo using samples from population-based ULA, and $\Delta t_k = t_k - t_{k-1}$. \citet{CALDERHEAD20094028} also showed that the discretization bias can be corrected using Kullback-Leibler (KL) divergence between adjacent power posteriors, providing the exact Riemann integral, (Appendix~\ref{sec:disc-ti}). In Appendix~\ref{sec:TI-learning-grad}, we show this form reduces to a sum of log-partition ratios known as the Steppingstone estimator (SE) by \citet{Annis2019}. While exact, neither form provides gradients to learn the prior, but since the SE gradient equals the MLE gradient in Eq.~\ref{eq:learning-grad}, (shown under Appendix~\ref{sec:TI-learning-grad}), the prior can instead be learned by adding the original posterior expectation of the log-prior from Eq.~\ref{eq:learning-grad}, estimated using the final temperature samples at $t_{N_t} = 1$.

While SE is exact, it produces importance-weighted gradients, (Appendix~\ref{sec:SS-IS}), that concentrate on single samples when $\Delta t_k$ is large. Eq.~\ref{eq:SS-KAEM} is preferred for learning the generator, with $O(\Delta t_k^2)$ bias diminished by increasing $N_t$.

\paragraph{Population-based ULA} In their implementation of Thermodynamic Integration, \citet{CALDERHEAD20094028} estimated the Thermodynamic Integral with the discretization in Sec.~\ref{sec:disc-ti}, using population-based Markov-chain Monte Carlo (MCMC) and Parallel Tempering (PT) \citep{PT, Marinari1992, Hukushima1996} to sample from all power posteriors.

Specifically, parallel chains were simultaneously maintained at each temperature, (or Replica), using Metropolis-Hastings (MH) \citep{13ab5b5e-0237-33fb-a7a8-6f6e4e0d4e0f}, to drive local moves for each power posterior. They also proposed a geometric path between the prior and posterior, enabling global swaps between adjacent temperatures. This approach allowed higher temperatures to leverage the efficient mixing of lower temperatures, thereby facilitating a more thorough exploration of the posterior landscape.

We adopt a similar strategy but use ULA (Eq.~\ref{eq:ula}) to drive local moves within each Replica. Annealing with power posteriors can further improve mixing against multimodal posterior landscapes. Mixing can be further improved by proposing swaps between adjacent temperatures, where each sample independently accepts or rejects the exchange:
\begin{align}
    r^{(s)} = \frac{ p_{\Phi} \left(\boldsymbol{x}\mid\boldsymbol{z}^{(s, \; i, \; t_{k+1})} \right)^{t_{k}} p_{\Phi} \left(\boldsymbol{x}\mid\boldsymbol{z}^{(s, \; i, \; t_{k})}\right)^{t_{k+1}} }
    { p_{\Phi} \left(\boldsymbol{x}\mid\boldsymbol{z}^{(s, \; i, \; t_{k})}\right)^{t_{k}} p_{\Phi} \left(\boldsymbol{x}\mid\boldsymbol{z}^{(s, \; i, \; t_{k+1})} \right)^{t_{k+1}} },
\label{eq:global-swap}
\end{align}
where ${(s, \; i, \; t_{k})}$ indexes the sample, ULA iteration, and temperature respectively. For each sample $s$, the proposed swap $\boldsymbol{z}^{(s, \; i, \; t_{k+1})} \leftrightarrow \boldsymbol{z}^{(s, \; i, \; t_{k})}$ is accepted with probability $\min \left( 1, r^{(s)} \right)$. In this study, we adopt the deterministic even-odd (DEO) swap scheme \citep{Syed_2021}, which alternates between proposing swaps for all even-indexed pairs $(t_0 \leftrightarrow t_1, t_2 \leftrightarrow t_3, \dots)$ and all odd-indexed pairs $(t_1 \leftrightarrow t_2, t_3 \leftrightarrow t_4, \dots)$ every iteration, yielding a non-reversible parallel tempering scheme that outperforms its reversible counterparts, as shown by \citet{Syed_2021}.

\subsection{Implementation}
\label{sec:implementation}

Compilation and differentiation were implemented using Julia's Reactant \citep{Reactantjl2025} and Enzyme \citep{moses2020enzyme, moses2021enzyme_gpu, moses2022enzyme_parallel} packages, which optimize and lower Julia code to Multi-Level Intermediate Representation (MLIR) \citep{lattner2020mlircompilerinfrastructureend}. For the NIST experiments, univariate energy functions were realized with Gaussian Radial Basis Functions (RBF KANs) \citep{li2024kolmogorovarnoldnetworksradialbasis}, adopted instead of the cubic B-spline bases used by \citet{liu2024kankolmogorovarnoldnetworks}, as RBFs are Reactant-compatible and more efficient for GPU implementation. They have also been proven to provide an optimal solution to the regularized approximation/interpolation problem by \citet{58326}. Since importance sampling draws proposals from the prior's finite domain, fixed RBF centres are sufficient, and the grid updating scheme of \citet{liu2024kankolmogorovarnoldnetworks} is applied periodically to adapt their domains.

\section{Experiments}

\subsection{Setup}

\paragraph{Scope} We show that importance sampling becomes a viable and efficient training strategy for simple NIST \citep{mnist,fmnist} datasets, due to the low dimensionality of the latent space. We also demonstrate that the model remains effective when strictly adhering to KART with independent energy-based priors, thereby validating the use of KART as a source of structural inductive bias. Additionally, we plot a single learned distribution to demonstrate KAEM's interpretability. Architectures and hyperparameter details are provided in Table~\ref{tab:hp-nist}.

We then evaluate KAEM with its mixture prior on three RGB image datasets at different resolutions: SVHN ($32\times32$) \citep{37648}, CIFAR10 ($32\times32$) \citep{krizhevsky2009learning}, and CelebA ($64\times64$) \citep{liu2015deeplearningfaceattributes}. 

Against these datasets, we compare KAEM trained using single-posterior ULA sampling with the MLE criterion against a population-based ULA sampling strategy using the thermodynamic criterion. These comparisons assess the practicality of using mixture densities and component-wise ITS for scaling KAEM and using annealing to improve training for high-dimensional RGB images. We further benchmark against three baselines spanning the principal alternatives to a univariate-prior latent EBM:
\begin{itemize}
    \setlength{\itemsep}{2pt}
    \item a VAE \citep{kingma2022autoencodingvariationalbayes}, which uses a fixed Gaussian prior with amortized inference;
    \item the neural latent EBM of \citet{pang2020learninglatentspaceenergybased}, which replaces KAEM's univariate factorization with a neural-network prior trained by contrastive divergence;
    \item a denoising diffusion probabilistic model \citep{ho2020denoisingdiffusionprobabilisticmodels}, which forgoes the latent prior entirely and generates by iterative denoising in pixel space.
\end{itemize}
Hyperparameters and architectures are provided in Table~\ref{tab:hp-cnn} and Appendix~\ref{app:arch} respectively.


\paragraph{Benchmarks} Benchmarks were collected for KAEM, the VAE, neural latent EBM, and DDPM baselines in FP32 on an NVIDIA RTX 4060 GPU paired with an Intel Core i9-13900H host (14 cores, 20 threads), using Julia's BenchmarkTools package \citep{BenchmarkTools}. Compilation and reverse differentiation were performed using Julia's Reactant \citep{Reactantjl2025} and Enzyme \citep{moses2020enzyme, moses2021enzyme_gpu, moses2022enzyme_parallel} packages. They reflect reverse autodifferentiation of the training criterion in Eqs.~\ref{eq:learning-grad} and~\ref{eq:SS-KAEM}, or the prior sampling process during inference.

\paragraph{Image metrics} We report the Fréchet Inception Distance (FID) \citep{heusel2018gans} and the Kernel Inception Distance (KID) \citep{binkowski2021demystifying}, both computed from a finite sample of $N$ generated images. The estimators $\text{FID}_N$ and $\text{KID}_N$ are biased relative to their infinite-sample limits, $\mathbb{E}[\text{FID}_N] \neq \text{FID}_\infty$, with a gap that depends on the generator being evaluated, so comparisons made at any fixed $N$ are unreliable \citep{chong2020effectively}. Following \citet{chong2020effectively}, we compute each metric at sample sizes $N \in \{2000, 4000, \ldots, 20{,}000\}$ and extrapolate to $N = \infty$ via linear regression, yielding $\overline{\text{FID}}_\infty$ and $\overline{\text{KID}}_\infty$. These are effectively unbiased in $N$.


This removes finite-sample bias but leaves feature extraction untouched. Both metrics evaluate samples in the feature space of an InceptionV3 network \citep{szegedy2015rethinking} trained on ImageNet, so the score systematically favours ImageNet-like distributions, which cannot be addressed by drawing more samples. Quantifying sample fidelity therefore remains an inexact science, and our conclusions are also informed by our subjective perception of the images. No image metrics were applied to the NIST datasets, whose non-RGB content lies outside this distribution.

\subsection{Results}
\label{sec:experiment}

\paragraph{MNIST \& FMNIST}

KAEM was trained on MNIST and FMNIST \citep{mnist,fmnist} for 10 epochs with 50,000 training examples, using importance sampling with a batch and sample size of $N=100$ (5,000 parameter updates). No MCMC was required for either prior or posterior sampling. The architecture adhered strictly to KART, using the KAN generator in Appendix~\ref{app:gen_adherent} with RBF energy functions \citep{li2024kolmogorovarnoldnetworksradialbasis} and independent priors.

KAEM consistently generated diverse samples under Gaussian, uniform, and uninformed reference priors (Appendix~\ref{app:images-nist}). Importance sampling is generally impractical for posterior inference in generative models due to variance that grows with prior-posterior mismatch \citep{dupuis2004importance}. Its viability in this setting indicates that KAEM's learned prior maintains sufficient overlap with the posterior in the low-dimensional latent space, rendering MCMC unnecessary for simple datasets.

\begin{figure}[t]
    \centering
    \includegraphics[width=0.77\linewidth]{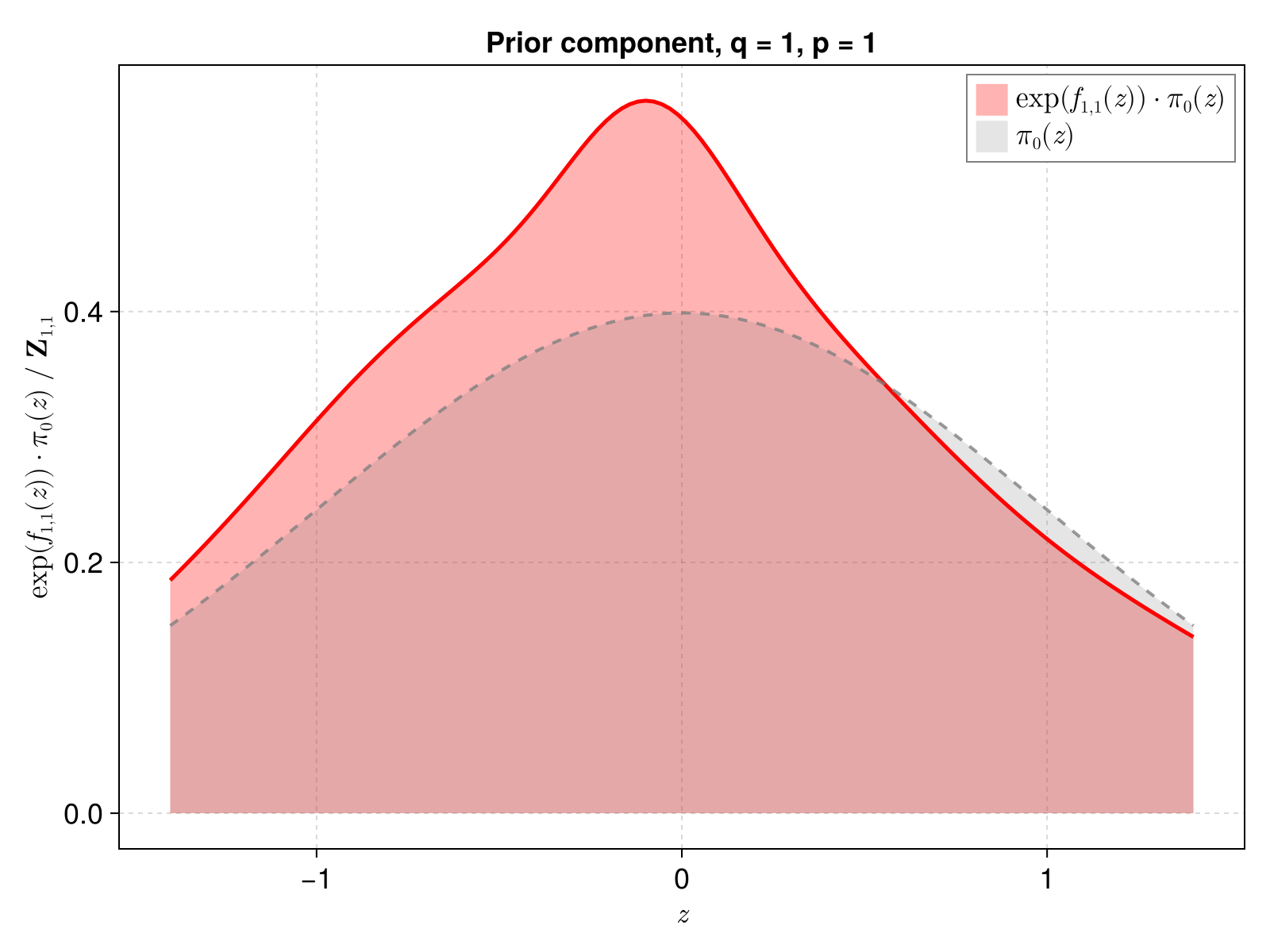}
    \caption{Learned prior component $p_{q=1,p=1}(z)$ (shaded) and Gaussian reference prior $\pi_0(z)$ (dashed) after training on FMNIST. The partition function $Z_{q,p}$ is computed by quadrature, normalizing the density up to a small discretization bias.}
    \label{fig:density-plot}
\end{figure}

Since KAEM's energy functions are univariate, each component of the learned prior is a 1D density and can be plotted as a curve. Fig.~\ref{fig:density-plot} shows a single prior component after training on FMNIST with a Gaussian reference. The learned density (shaded) departs from the reference prior (dashed) to capture structure in the data. Such inspection is infeasible for neural EBM priors.

\paragraph{SVHN, CIFAR10, and CelebA}

All four model classes were trained with Adam for 80,000 parameter updates (100 epochs at batch size 50 over 40,000 training examples). The VAE and DDPM use a single learning rate of $1\times10^{-4}$, the value reported by \citet{ho2020denoisingdiffusionprobabilisticmodels}. For stability in contrastive divergence, we adopted the two-rate schedule of \citet{pang2020learninglatentspaceenergybased} for the neural latent EBM and KAEM, $2\times10^{-5}$ for the energy network and $1\times10^{-4}$ for the generator. The 80,000-update budget is broadly comparable to the original schedule of \citet{pang2020learninglatentspaceenergybased} but remains an order of magnitude below reference DDPM schedules. We used the same CNN decoder architecture for KAEM, the VAE, and the neural latent EBM, with the DDPM UNet constructed at matching depth and channel widths, so that the convolutional generator architecture was held constant across all four model classes. Sizes were bounded by the VAE encoder's VRAM footprint, and architecture tables are provided in Appendix~\ref{sec:arch}. For KAEM, posterior sampling used 40 steps of ULA with a standard Gaussian reference prior $\pi_0 = \mathcal{N}(0,1)$, and annealing used $N_t = 10$ power posteriors with the schedule of Sec.~\ref{sec:optimal-temp}. Remaining hyperparameters are in Tab.~\ref{tab:hp-cnn}.

\begin{table*}[t]
    \centering
    \scriptsize
    \caption{Effectively unbiased image quality metrics (lower is better). Bold marks the best result among latent-prior models for each dataset and metric. The pixel-space DDPM is reported as a reference (italicised), as it does not share the latent-prior architecture.}
    \label{tab:metrics}
    \renewcommand{\arraystretch}{1.15}
    \setlength{\tabcolsep}{6pt}
    \begin{tabular}{llccc|ccc}
        \toprule
        & & \multicolumn{3}{c}{$\overline{\textbf{FID}}_\infty$} & \multicolumn{3}{c}{$\overline{\textbf{KID}}_\infty$} \\
        \cmidrule(lr){3-5} \cmidrule(lr){6-8}
        \textbf{Dataset} & \textbf{Model}
        & Value & Std. Err. & $R^2$
        & Value & Std. Err. & $R^2$ \\
        \midrule
        \multirow{5}{*}{SVHN}
        & KAEM (MLE)
        & \textbf{43.23} & 0.06 & 0.996
        & \textbf{0.0316} & $9.04e^-5$ & 0.017 \\
        & KAEM (Thermo $N_t=10$)
        & 108.74 & 0.129 & 0.965
        & 0.1018 & 0.00020 & 0.067 \\
        & VAE
        & 107.80 & 0.066 & 0.991
        & 0.1059 & 0.00021 & 0.356 \\
        & Neural Latent EBM
        & 63.69 & 0.110 & 0.986
        & 0.0515 & 0.00011 & 0.042 \\
        \cmidrule(lr){2-8}
        & \emph{DDPM (ref.)}
        & \emph{20.00} & \emph{0.062} & \emph{0.997}
        & \emph{0.0133} & \emph{0.00010} & \emph{0.540} \\
        \midrule
        \multirow{5}{*}{CIFAR10}
        & KAEM (MLE)
        & 134.55 & 0.094 & 0.996
        & 0.124 & 0.00029 & 0.044 \\
        & KAEM (Thermo $N_t=10$)
        & 225.34 & 0.109 & 0.988
        & 0.2263 & 0.00022 & 0.482 \\
        & VAE
        & 181.05 & 0.145 & 0.984
        & 0.1710 & 0.00016 & 0.701 \\
        & Neural Latent EBM
        & \textbf{120.28} & 0.130 & 0.992
        & \textbf{0.1067} & 0.00015 & 0.004 \\
        \cmidrule(lr){2-8}
        & \emph{DDPM (ref.)}
        & \emph{57.88} & \emph{0.107} & \emph{0.996}
        & \emph{0.0526} & \emph{0.00014} & \emph{0.150} \\
        \midrule
        \multirow{5}{*}{CelebA}
        & KAEM (MLE)
        & 96.88 & 0.149 & 0.996
        & 0.110 & 0.0002 & 0.146 \\
        & KAEM (Thermo $N_t=10$)
        & 108.04 & 0.079 & 0.988
        & 0.1129 & 0.00011 & 0.369 \\
        & VAE
        & 136.57 & 0.112 & 0.969
        & 0.1461 & 0.00012 & 0.094 \\
        & Neural Latent EBM
        & \textbf{85.57} & 0.147 & 0.970
        & \textbf{0.0865} & 0.00027 & 0.000 \\
        \cmidrule(lr){2-8}
        & \emph{DDPM (ref.)}
        & \emph{174.05} & \emph{0.443} & \emph{0.854}
        & \emph{0.1706} & \emph{0.00049} & \emph{0.002} \\
        \bottomrule
    \end{tabular}
\end{table*}

\begin{figure}[t]
    \centering
    \includegraphics[width=0.95\linewidth]{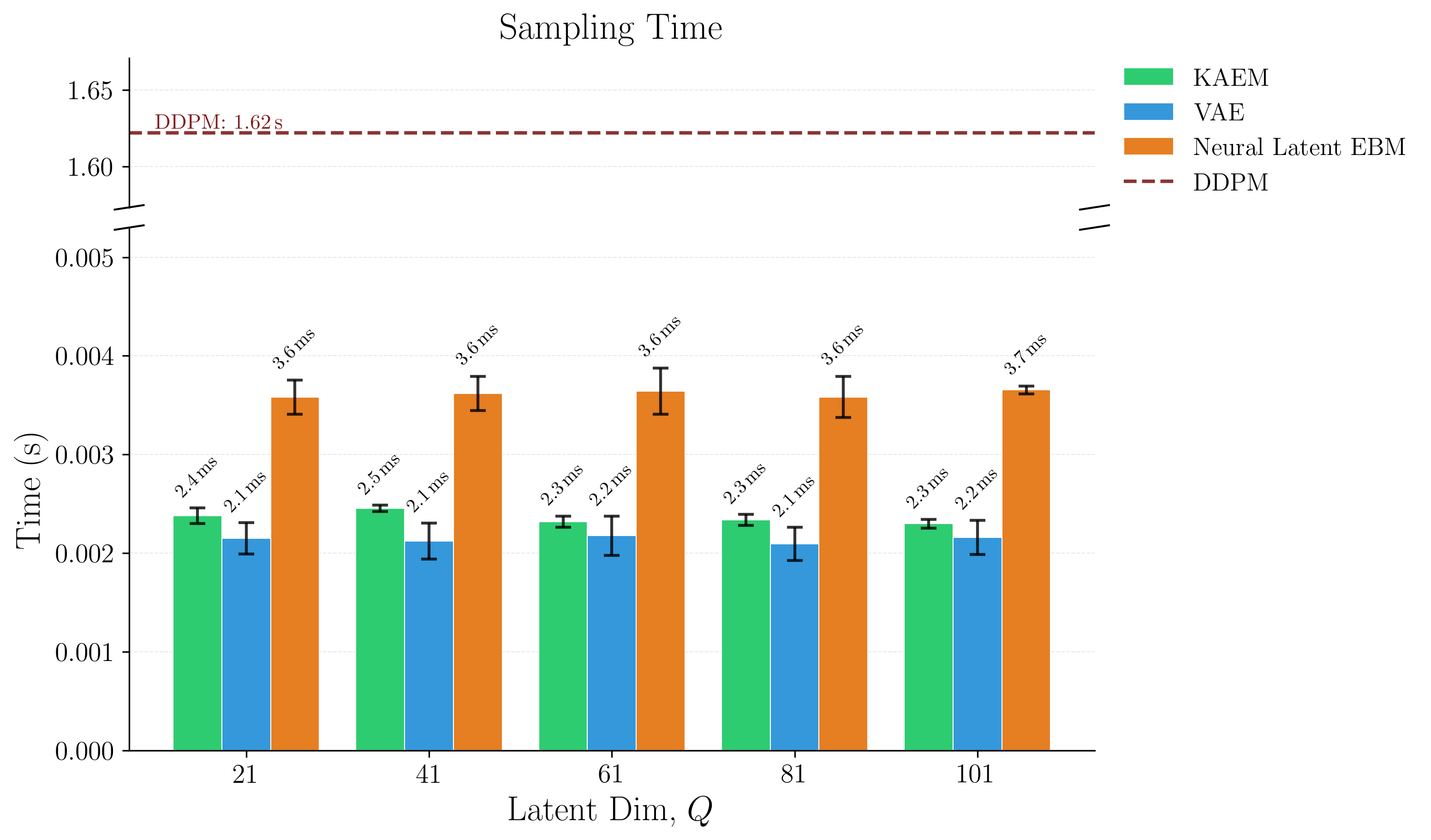}
    \caption{Sampling (inference) times of KAEM, VAE, and the neural latent EBM \citep{pang2020learninglatentspaceenergybased} across latent size, $Q$, with DDPM shown as a constant reference (independent of $Q$). KAEM uses inverse transform sampling (ITS) from the EBM prior; VAE decodes from $\boldsymbol{z} \sim \mathcal{N}(\mathbf{0}, \mathbf{I})$; the neural latent EBM samples its prior with $60$ ULA steps; DDPM denoises with $T_s = 1000$ steps in pixel space. KAEM-ITS uses 200 quadrature nodes; the small gap to VAE can be closed by optimizing or caching the ITS computational kernel, (since the partition remains the same per generation and does not need recomputing).}
    \label{fig:sampling-time}
\end{figure}

\begin{figure*}[t]
    \centering
    \begin{subfigure}[t]{0.25\textwidth}
        \centering
        \includegraphics[width=\linewidth]{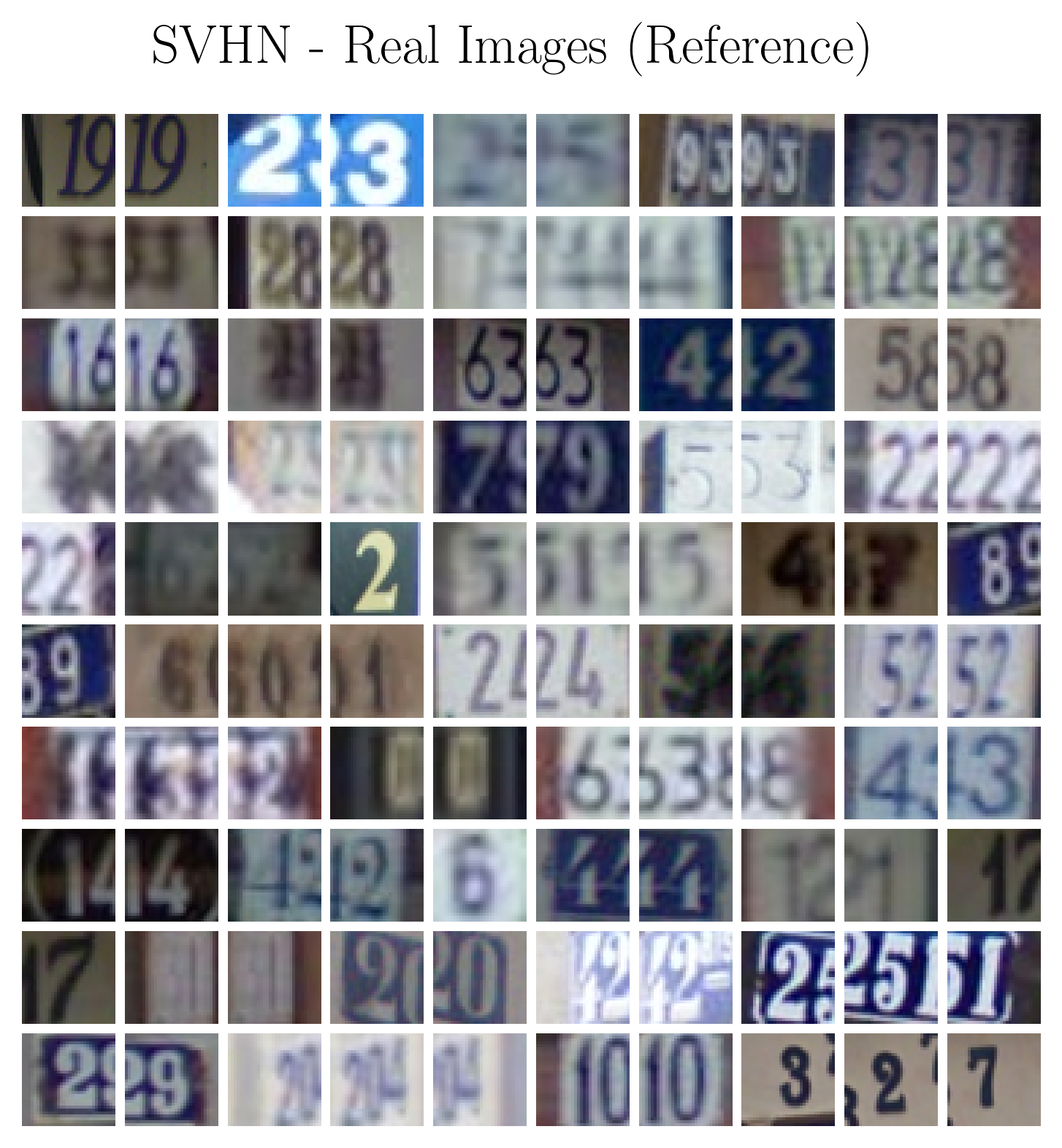}
        \caption{Real}
    \end{subfigure}\hfill
    \begin{subfigure}[t]{0.25\textwidth}
        \centering
        \includegraphics[width=\linewidth]{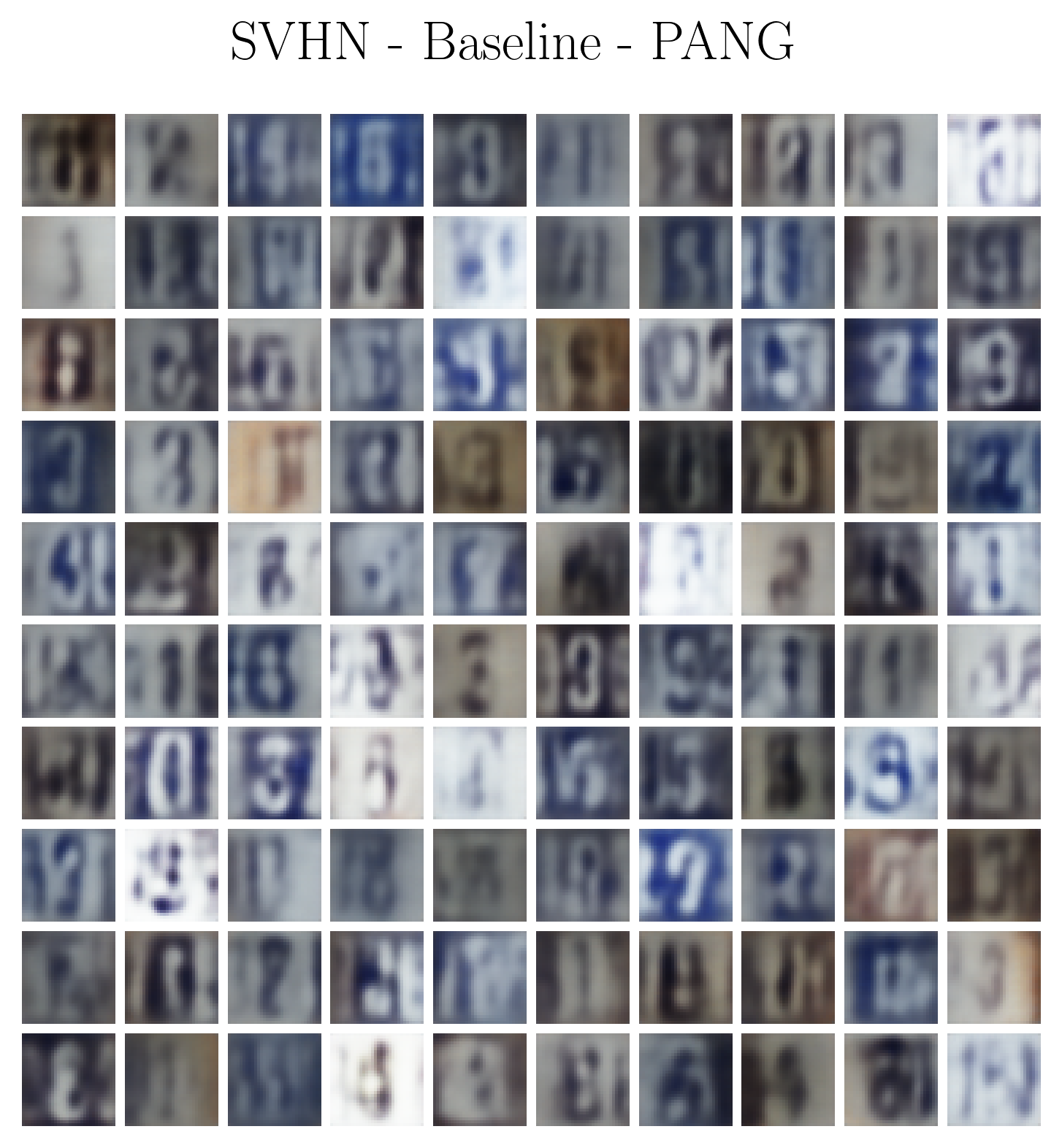}
        \caption{Neural EBM}
    \end{subfigure}\hfill
    \begin{subfigure}[t]{0.25\textwidth}
        \centering
        \includegraphics[width=\linewidth]{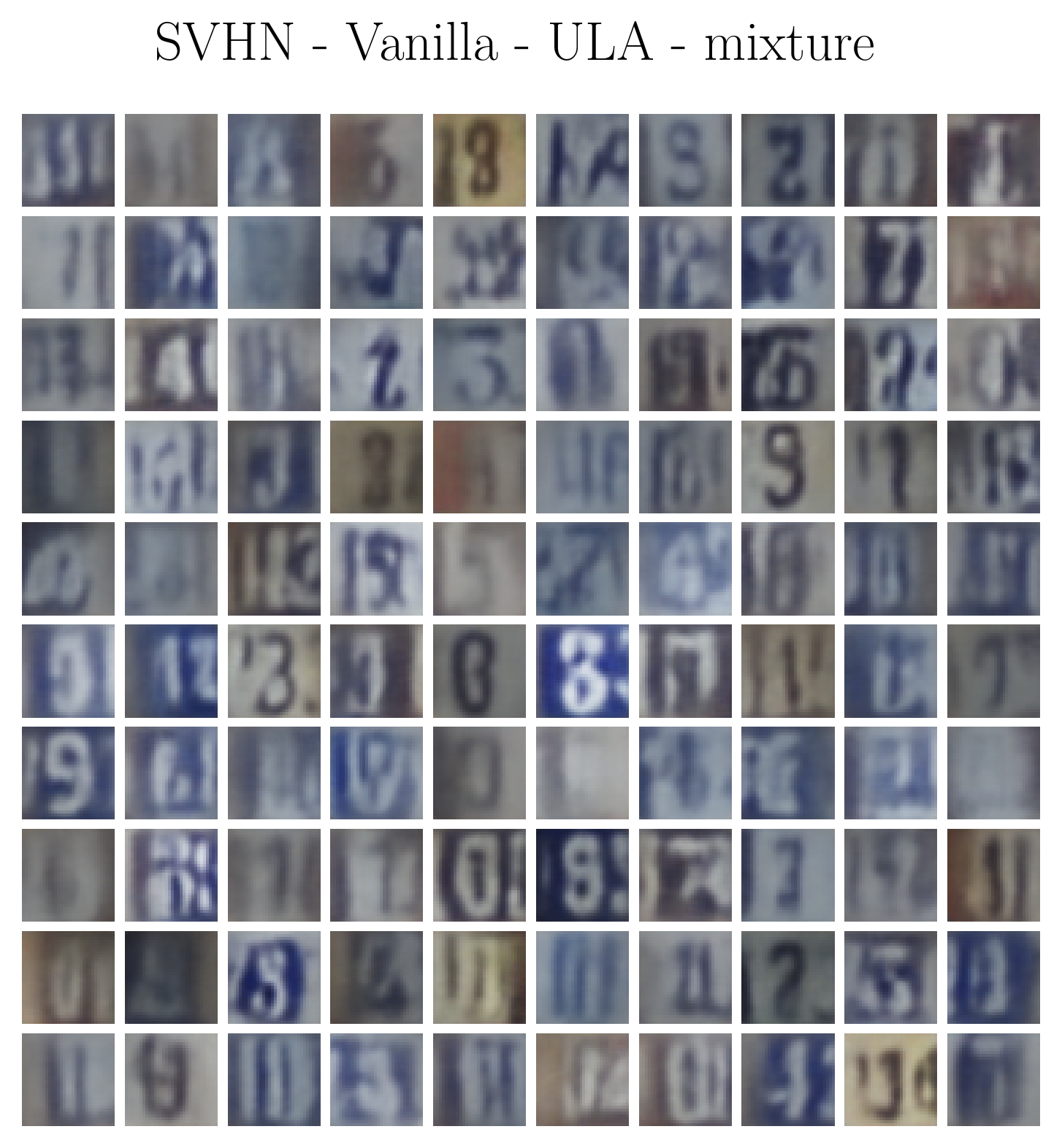}
        \caption{MLE}
    \end{subfigure}\hfill
    \begin{subfigure}[t]{0.25\textwidth}
        \centering
        \includegraphics[width=\linewidth]{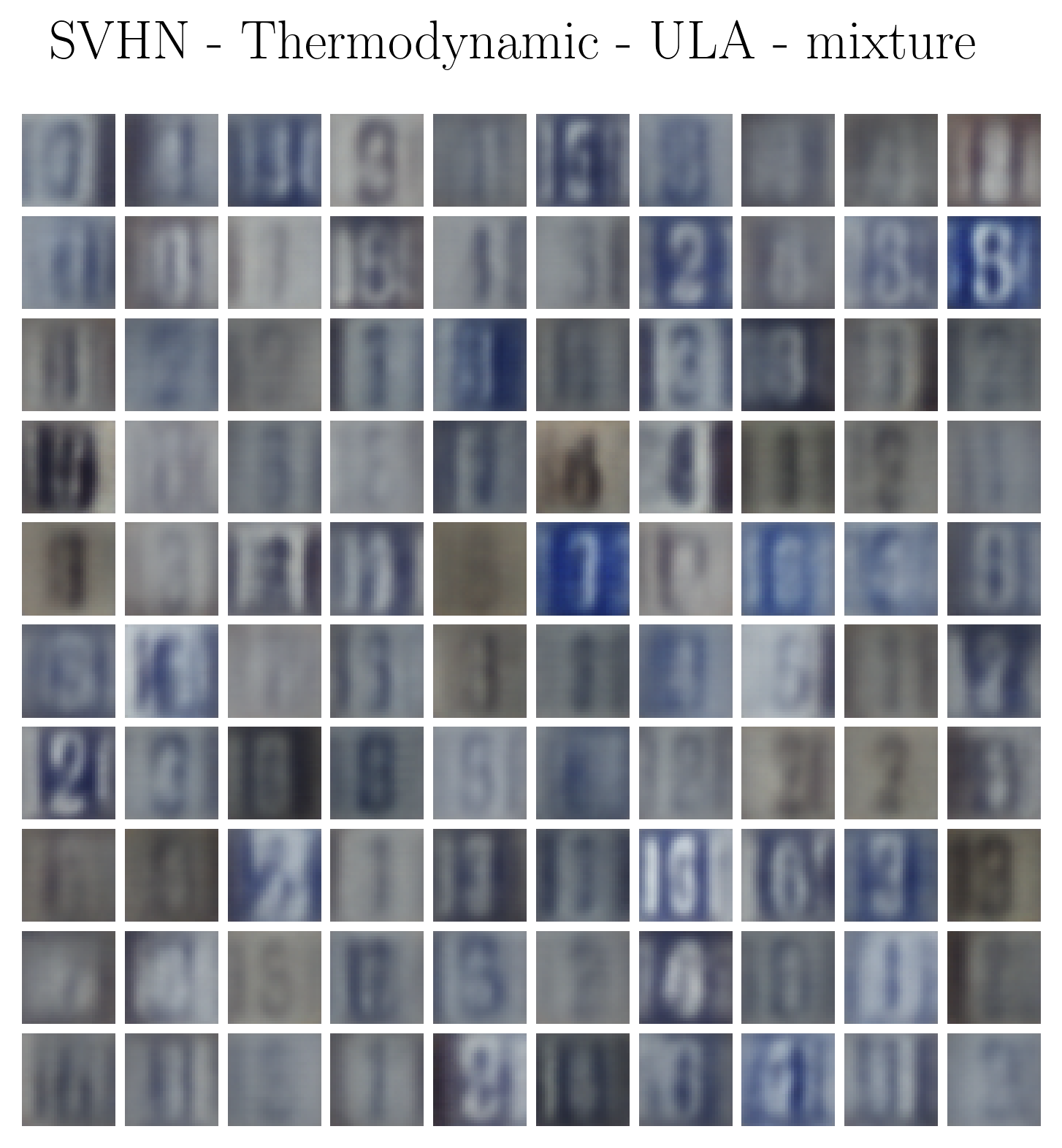}
        \caption{$N_t=10$}
    \end{subfigure}\hfill
    \caption{SVHN samples generated by the Neural EBM \citep{pang2020learninglatentspaceenergybased}, KAEM trained without annealing (MLE), and KAEM with thermodynamic training using \(N_t=10\). All other samples can be found under Appendix \ref{app:images-cif}.}
    \label{fig:celeba-main}
\end{figure*}
Among latent-prior models, the neural latent EBM of \citet{pang2020learninglatentspaceenergybased} attained the lowest $\overline{\text{FID}}_\infty$ and $\overline{\text{KID}}_\infty$ on all datasets except SVHN (Tab.~\ref{tab:metrics}). Thermodynamic training degraded all datasets but subjectively improved CelebA. DDPM led SVHN and CIFAR10 by a wide margin in pixel space, but was the weakest of the four classes on CelebA. Visual inspection in Appendix~\ref{app:images-cif} is subjective but mostly consistent with these rankings excluding CelebA.

Sampling speed separated the models more sharply than fidelity (Fig.~\ref{fig:sampling-time}). The neural prior requires ULA for both prior sampling at inference time, which scales linearly with the number of steps and is sensitive to step size. KAEM produces latent-prior samples while avoiding iteration. DDPM dominated SVHN and CIFAR10 but was the weakest on CelebA, where its reference training budget exceeded ours by an order of magnitude and our smaller UNet may have lacked capacity to compensate at $64\times64$.

KAEM completed a single ITS pass in $2.3$~ms, comparable to the VAE ($1.8$~ms) and faster than the neural latent EBM ($3.2$~ms with 60 ULA steps), and roughly $700\times$ faster than DDPM ($1.6$~s with $1000$ steps). Importance-sampling training matched a VAE step in cost. KAEM ULA-based training time was slower and grew with latent size, $Q$ (Appendix~\ref{app:bench}).

\begin{figure*}[t]
    \centering
        \begin{subfigure}[t]{0.48\textwidth}
        \centering
        \includegraphics[width=\linewidth]{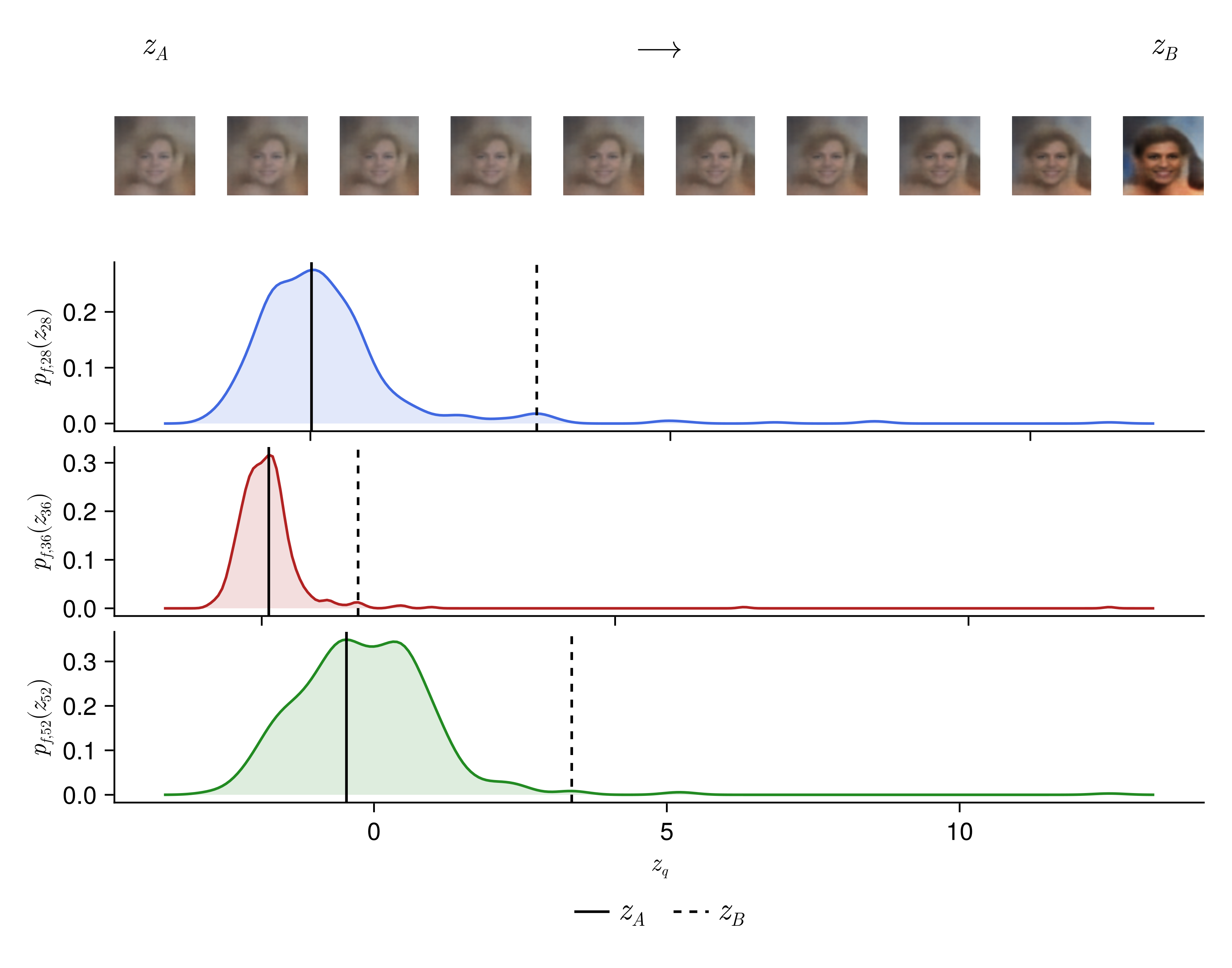}
        \caption{MLE}
    \end{subfigure}\hfill
    \begin{subfigure}[t]{0.48\textwidth}
        \centering
        \includegraphics[width=\linewidth]{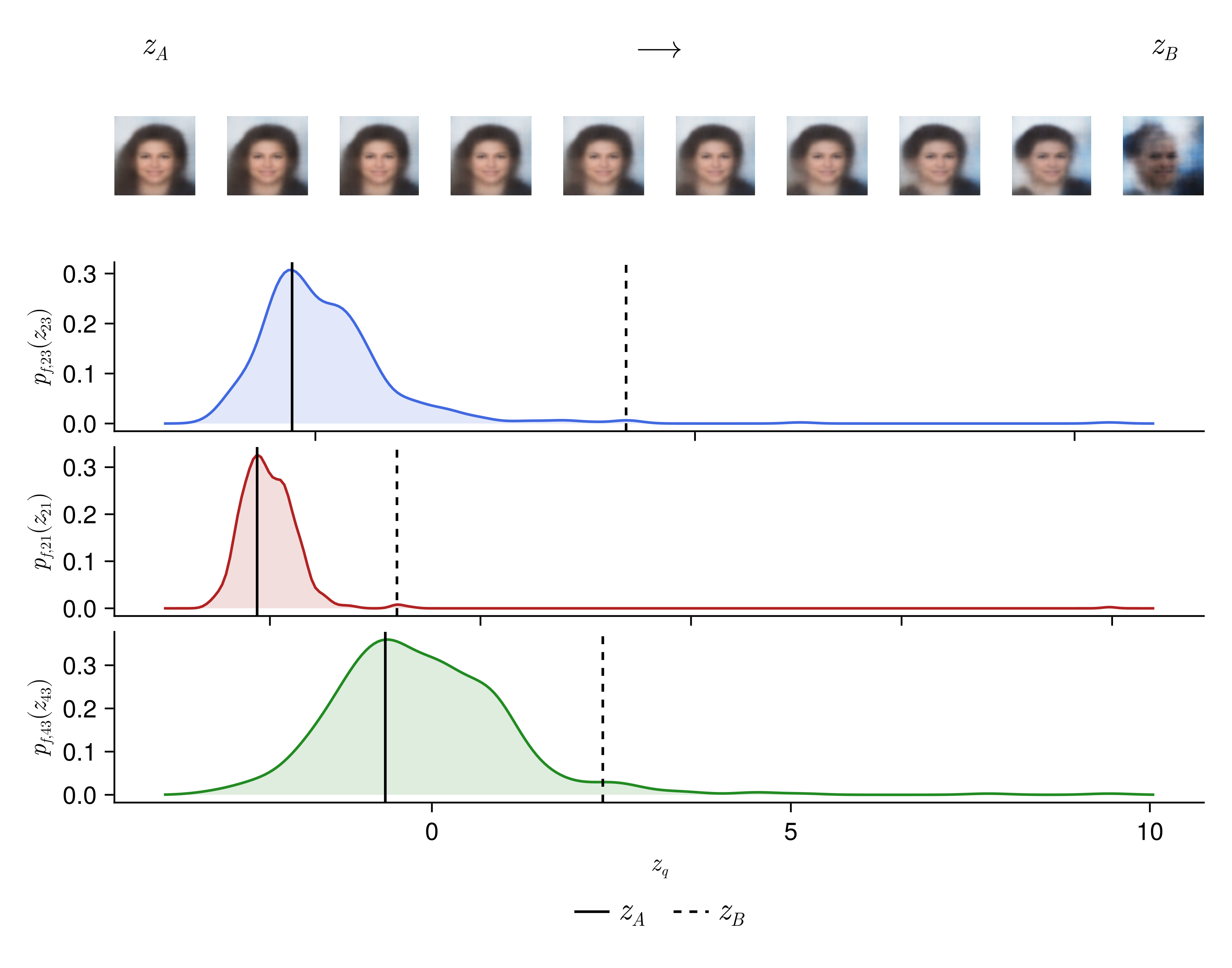}
        \caption{Thermo $N_t=10$}
    \end{subfigure}\hfill
    \caption{Interpolation between high and low density modes on CelebA. Top: decoded images from $z_A$ to $z_B$. Below: learned prior densities $p_{f,q}(z_q)$ for three dimensions, with $z_A$ (solid) and $z_B$ (dashed) marked as the latent samples that were interpolated between.}
    \label{fig:slerp}
\end{figure*}

KAEM's univariate factorization rendered each per-dimension prior as a 1D density that can be plotted, which the neural network priors of related work do not allow. Interpolation on CelebA (Fig.~\ref{fig:slerp}) demonstrates smooth transitions between visually distinct samples while remaining consistent with the learned density. In the MLE example, the highest-density region corresponds to smoother, lower-frequency facial appearances with softer, more averaged features. As the interpolation moves toward lower-density regions, the generated faces progressively exhibit sharper details and stronger local contrast. In contrast, for the Thermodynamic setting, the low-density region reflects lower-fidelity facial reconstructions. These observations suggest that the structure of the learned densities captures meaningful variations in the generated samples, although the nature of these variations is not immediately apparent or consistent.

The contrast between KAEM-MLE and KAEM-Thermo reflects the bias-variance trade-off in posterior estimation. MLE estimated the gradient from a single ULA chain. Thermodynamic integration estimated over $N_t = 10$ annealed chains, reducing variance and improving mixing on multimodal posteriors at the cost of the discretization bias of Sec.~\ref{sec:thermo-help}. \citet{ghosh2023how} showed that gradient variance can aid generalization, with the useful level set by the architecture and loss landscape. In this experiment, a single chain may have mixed adequately, so the variance reduction and discretization bias may have detracted more than the mixing improved. Ablations over $N_t$ and the temperature schedule would be needed to separate these effects.

More VRAM would allow $N_t > 10$ temperatures in parallel, which \citet{CALDERHEAD20094028} suggest can improve sample quality. We also observed improved quality with increasing $N_t$, but the training cost was prohibitive in our constrained setting. Adjacent directions include alternative annealing schedules, longer ULA chains, samplers that adapt to the varying geometries of power posteriors (e.g., autoMALA \citep{bironlattes2024automalalocallyadaptivemetropolisadjusted}), and amortized posterior inference following the information bottleneck framework that \citet{kong2021unsupervised} derived for latent EBMs.

\section{Conclusions}

We presented KAEM, a latent energy-based model whose prior is built from univariate energy functions guided by the Kolmogorov-Arnold Representation Theorem. The univariate factorization admits exact, single-pass prior sampling via the inverse transform method, eliminating the iterative MCMC that conventional latent EBMs rely on at inference time. The neural latent EBM of \citet{pang2020learninglatentspaceenergybased} attained the best $\overline{\text{FID}}_\infty$ and $\overline{\text{KID}}_\infty$ among latent-prior models on all datasets except SVHN, with KAEM placing a close second throughout and outperforming on SVHN, while sampling at $\sim\!1.4\times$ lower cost. DDPM in pixel space led SVHN and CIFAR10 with $700\times$ slower inference, and was the weakest of the four model classes on CelebA with our limited training budget. Unlike the neural-network priors of related work, KAEM's prior decomposes into per-dimension 1D densities that can be plotted and related to the generative process in an insightful manner. We view KAEM as an initial step toward generative models whose latent priors are legible, transferable, and amenable to informed design.

\subsection{Future work}

\paragraph{Symbolic regression} \citet{liu2024kankolmogorovarnoldnetworks} previously demonstrated that symbolic regression can replace KAN univariate functions with closed-form mathematical expressions. The same approach can be applied to KAEM's latent prior to recover closed-form algebraic representations of latent densities. We believe this could be a promising step toward establishing a meaningful connection between the latent representation and the underlying generative process.

\paragraph{Expressivity and multimodal generation} Through other experiments, we found that the choice of prior architecture and its encoded inductive biases significantly affect sample quality. Future work could investigate this across different KAN bases and datasets, as well as explore how domain knowledge can guide the selection of univariate functions.

Additionally, it is often desirable to enrich the latent space with more expressive distributions. For example, Normalizing Flows \citep{papamakarios2021normalizingflowsprobabilisticmodeling} apply a sequence of reversible transformations that progressively map a simple base distribution into a complex one. This might be achievable in KAEM by introducing more latent distributions, or revising Eq.~\ref{eq:kaem_mix} into a Mixture of Experts framework \citep{cai2024surveymixtureexperts} where $\alpha_{q,p}$ could be realized as a more flexible kernel to enable different inductive biases. More broadly, latent generative models with energy-based priors have been shown to extend naturally to multimodal data generation \citep{yuan2024learningmultimodallatentgenerative}, suggesting that KAEM's framework could be adapted for joint modeling across data modalities.

\subsection{Edge applications}

Given the low-latency high-level synthesis (HLS) code provided in Appendix~\ref{app:code}, KAEM is a promising candidate for real-time processing of sensory data in edge and resource-constrained environments. Future work could investigate its application to tasks such as unsupervised anomaly detection, where deviations from normal sensor behaviour could be identified without requiring labelled data.

The capacity for multimodal generation, building on the approach of \citet{yuan2024learningmultimodallatentgenerative}, could further enable applications involving missing-modality reconstruction, and simulation of plausible sensory observations. Evaluating KAEM on real-world streaming sensor data, under practical compute and latency constraints, would provide a useful next step towards assessing its suitability for deployment in autonomous and embedded systems.

\paragraph{Expanding the role of importance sampling} For SVHN, CIFAR10, and CelebA, IS proved infeasible as a posterior sampler, necessitating ULA for the RGB image datasets. While IS is limited by estimator variance, ULA is constrained by mixing difficulties when sampling multimodal distributions (addressed here through annealing).

The failure of IS on more complex datasets likely stems from prior-posterior mismatch, which renders the IS estimator ineffective. Furthermore, IS restricts the prior to its initial domain, since proposals are generated via ITS using finite-domain KAN energy functions. In contrast, ULA can explore beyond this domain.

To address prior-posterior mismatch, future work could explore incorporating domain knowledge to shape the prior's energy functions and align them with the posterior before training begins, or develop methods for selecting appropriate initial domains and adapting them dynamically. Additionally, KAEM could learn domain-specific priors that transfer across generator architectures. In this setting, IS may serve as an efficient strategy for rapidly retraining new generators using priors already aligned with the posterior.

It could also potentially be treated as a low-cost fine-tuning strategy applied after initial training with ULA or population-based ULA. This approach may be particularly advantageous in resource-constrained environments, since IS introduces no additional parameter overhead. For example, IS could be used in edge applications to reduce the discrepancy between training conditions and real-world sensor data after deployment.

\paragraph{Stochastic gradient sampling and annealing} In our experiments, we found that using more temperatures or replicas than $N_t=10$ substantially improved sample fidelity, consistent with the observations of \citet{CALDERHEAD20094028}, albeit at a significant computational cost. Beyond increasing the number of replicas, a promising direction is to investigate more effective temperature schedules than the dynamic power-law schedule adopted in this study and described in Appendix~\ref{sec:thermo-help}. For example, an optimal schedule can be constructed by minimizing the Kullback-Leibler (KL) divergence between adjacent power posteriors. As discussed by \citet{CALDERHEAD20094028}, the discretization error associated with thermodynamic integration is related to the KL divergence between neighboring power posteriors, suggesting a potential means of analytically choosing the temperature ladder, by deriving the required temperature spacing using identities from Appendix~\ref{app:derivations}.

Another direction is to reduce the cost of posterior sampling itself. Since KAEM trains by the Unadjusted Langevin Algorithm (ULA), it would be natural to investigate stochastic-gradient variants such as stochastic gradient Langevin dynamics (SGLD) \citep{31044823104568}. SGLD could replace the full ULA sampling loop per parameter update with an single unbiased estimate obtained from a minibatch of the dataset, which could substantially reduce the cost of each instance of posterior sampling. However, the stochasticity introduced by minibatch gradients also introduces discretization and sampling error, making the choice of minibatch size, step-sizes, and temperature ladder more important. Evaluating this trade-off, and determining whether SGLD can maintain the fidelity of ULA while reducing wall-clock cost, is a promising direction for future work.

\bibliography{refs}

\bibliographystyle{icml2026}

\newpage
\appendix
\onecolumn

\section{Codebase}
\label{app:code}

This study was implemented in Julia: \url{https://github.com/PritRaj1/julia-kaem}. A finalized JAX package is also available: \url{https://github.com/PritRaj1/thermo-ebms}. High-Level Synthesis (HLS) for FPGA is provided at \url{https://github.com/PritRaj1/kaem-hls}. 

\section{Benchmarks}
\label{app:bench}

Benchmarks were collected for KAEM, the VAE, neural latent EBM \citep{pang2020learninglatentspaceenergybased}, and DDPM baselines in FP32 on an NVIDIA RTX 4060 GPU paired with an Intel Core i9-13900H host (14 cores, 20 threads), using Julia's BenchmarkTools package \citep{BenchmarkTools}. Compilation and reverse differentiation were performed using Julia's Reactant \citep{Reactantjl2025} and Enzyme \citep{moses2020enzyme, moses2021enzyme_gpu, moses2022enzyme_parallel} packages. All benchmarks are based on the hyperparameters and architectures used for SVHN ($32\times32$), and reflect a reverse autodifferentiation of the training criteria in Eqs.~\ref{eq:learning-grad} and~\ref{eq:SS-KAEM}, or the prior sampling process during inference. The DDPM baseline operates in pixel space and is not parameterized by a latent dimension; its cost is therefore reported as constant across the latent-dimension axis. Note that garbage collection (GC) is zero in all cases, as Reactant compiles away all intermediate allocations.

\begin{figure}[H]
    \centering
    \includegraphics[width=\linewidth]{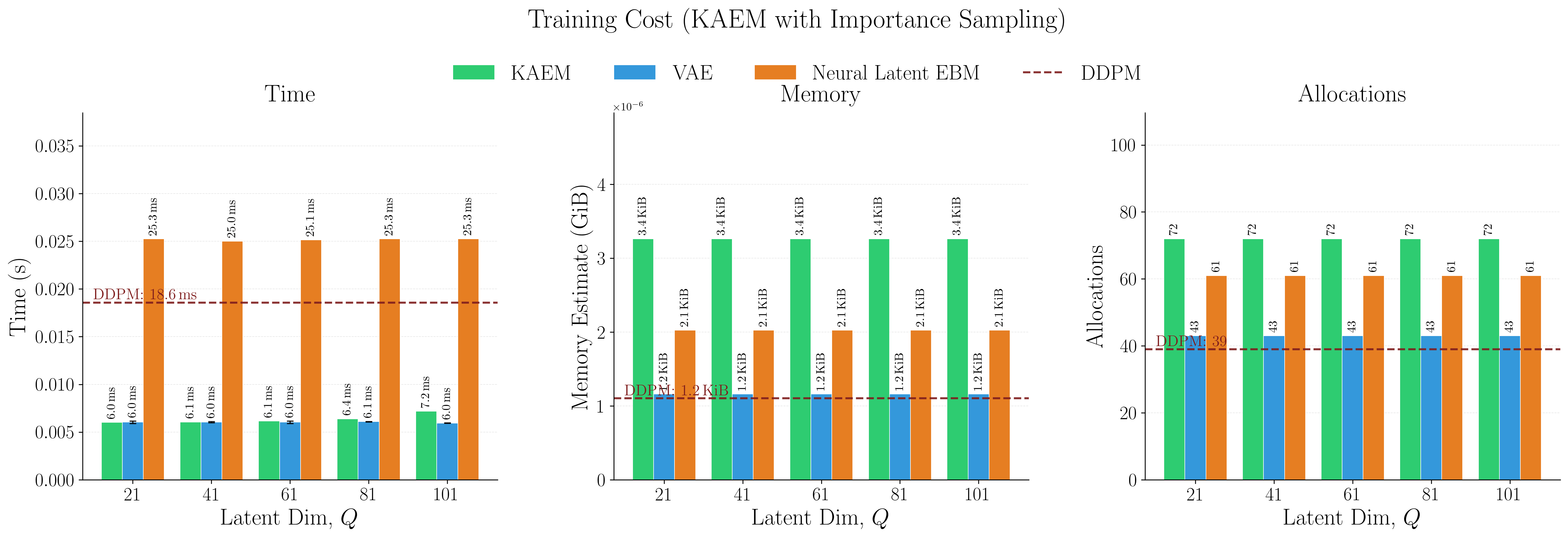}
    \caption{Training cost comparison (single training step) between KAEM (MLE with importance sampling), VAE, and the neural latent EBM \citep{pang2020learninglatentspaceenergybased} across latent dimensions $Q \in \{21, 41, 61, 81, 101\}$, with DDPM as a constant reference (it operates in pixel space and is independent of $Q$). Metrics shown include execution time (with error bars), memory usage, and allocations.}
    \label{fig:bench-training-importance}
\end{figure}

\begin{figure}[H]
    \centering
    \includegraphics[width=\linewidth]{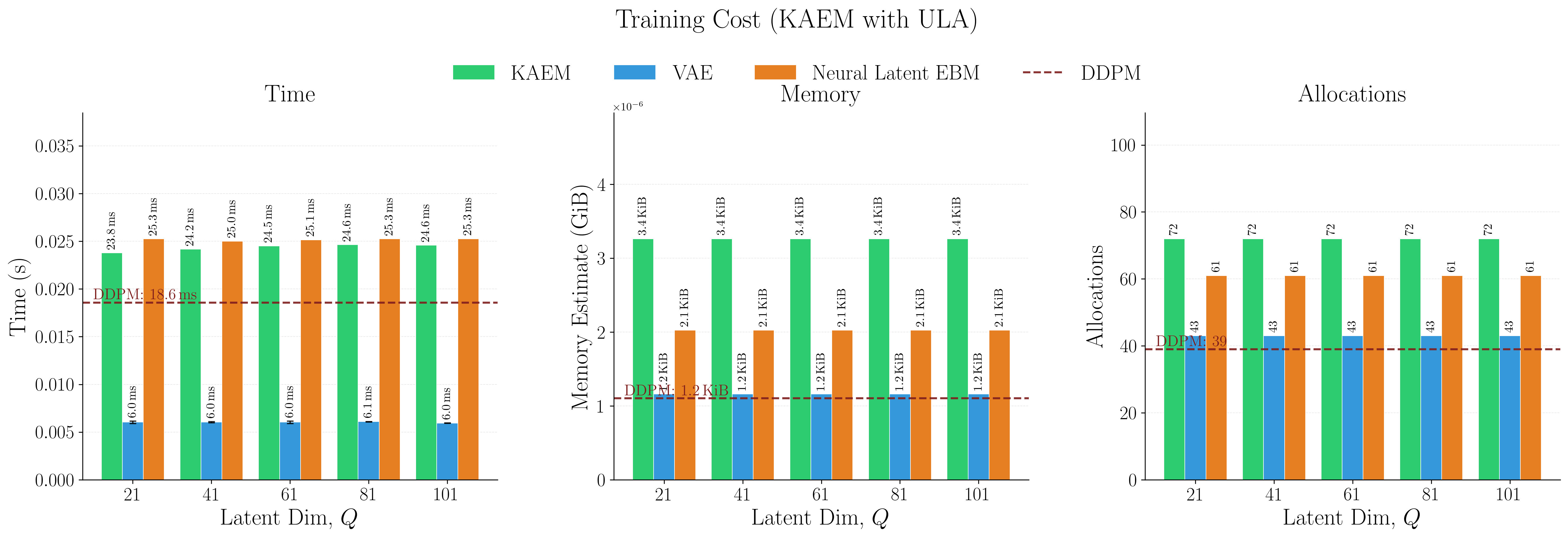}
    \caption{Training cost comparison (single training step) between KAEM (MLE with 40 ULA iterations), VAE, and the neural latent EBM \citep{pang2020learninglatentspaceenergybased} across latent dimensions $Q \in \{21, 41, 61, 81, 101\}$, with DDPM as a constant reference. Metrics shown include execution time (with error bars), memory usage, and allocations.}
    \label{fig:bench-training}
\end{figure}

\begin{figure}[H]
    \centering
    \includegraphics[width=\linewidth]{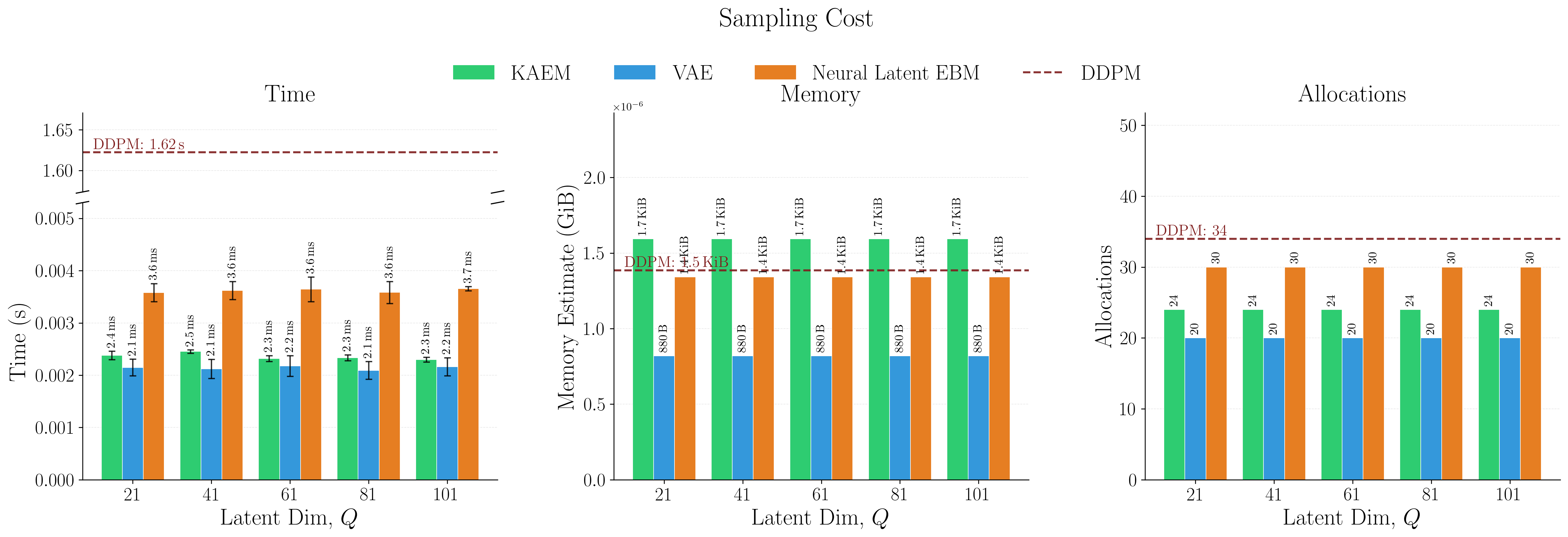}
    \caption{Sampling cost comparison between KAEM (MLE), VAE, and the neural latent EBM \citep{pang2020learninglatentspaceenergybased} across latent dimensions, with DDPM as a constant reference. KAEM uses inverse transform sampling (ITS) from the EBM prior; VAE decodes from $\boldsymbol{z} \sim \mathcal{N}(\mathbf{0}, \mathbf{I})$; the neural latent EBM samples its prior with $60$ ULA steps; DDPM denoises with $T_s = 1000$ steps in pixel space.}
    \label{fig:bench-sampling}
\end{figure}

\begin{figure}[H]
    \centering
    \includegraphics[width=\linewidth]{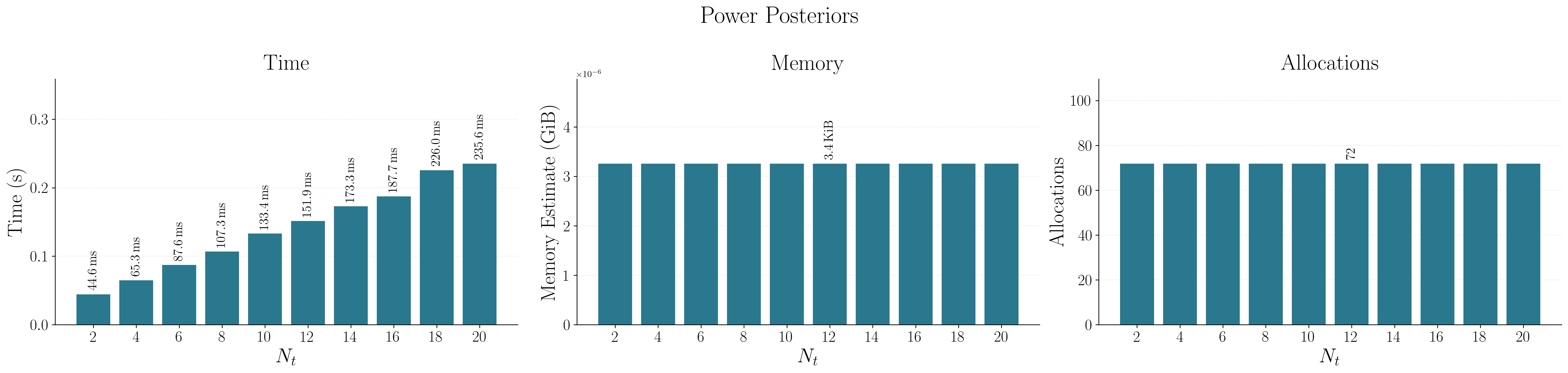}
    \caption{KAEM thermodynamic training cost scaling with the number of power posteriors $N_t \in \{2, 4, 6, 8, 10, 12, 14, 16, 18, 20\}$. The red dashed line indicates the reference cost at $Q=41$ with $N_t=1$ (single posterior, no thermodynamic integration), demonstrating the computational overhead of using multiple temperature replicas.}
    \label{fig:bench-temperatures}
\end{figure}

\section{Interpreting the Kolmogorov-Arnold Theorem}

\subsection{Inner functions as pushforwards}
\label{sec:measureable}

In Eq.~\ref{eq:inv-cdf}, \( F^{-1}_{\pi_{q,p}} : [0,1] \to \mathcal{Z} \) is the generalized inverse CDF associated with a parameterized latent measure, $\pi_{q,p}$ in the latent space, \(\mathcal{Z} \subset \mathbb{R}\). We have therefore interpreted \(u_p \sim \text{Unif}( 0,1)\) in Eq.~\ref{eq:KAT} as a uniform random variable in the probability space, $\left( [0,1], \mathcal{B}([0,1]), \mu \right)$, where \(\mathcal{B}\left([0,1]\right)\) is the Borel \(\sigma\)-algebra on \([0,1]\) and $\mu$ is the uniform probability measure on $[0,1]$. This is valid under standard regularity assumptions on $F$, since it is a measurable, monotone, increasing function, and thus $F_{\pi_{q,p}}^{-1}(u) = \inf \left\{ z \in \mathcal{Z} : F_{\pi_{q,p}}(z) \geq u \right\}$ is a pushforward (change of variables) between the probability spaces:
\begin{equation}
    \left([0,1], \mathcal{B}\left([0,1]\right), \mu\right) \; \underset{F_{\pi_{q,p}}^{-1}}{\longrightarrow} \; \left(\mathcal{Z}, \mathcal{B}\left(\mathcal{Z}\right), \pi_{q,p}\right).
\end{equation}
The pushforward measure $\pi_{q,p}$ is then defined on every measurable set $A \in \mathcal{B}(\mathcal{Z})$ by
\begin{equation}
    \pi_{q,p}(A) \;=\; \big(F^{-1}_{\pi_{q,p}}\big)_{\!*}\,\mu(A) \;=\; \mu\!\left(F_{\pi_{q,p}}(A)\right),
    \label{eq:pushforward}
\end{equation}
and, since $\mu$ admits the uniform Lebesgue density $\mathbf{1}_{[0,1]}$ and $F_{\pi_{q,p}}$ is differentiable almost everywhere with derivative denoted $F'_{\pi_{q,p}}$, the change-of-variables identity for probability densities simplifies:
\begin{equation}
    p_{q,p}(z) \;=\; \mathbf{1}_{[0,1]}\!\big(F_{\pi_{q,p}}(z)\big)\cdot|F'_{\pi_{q,p}}(z)| \;=\; F'_{\pi_{q,p}}(z),
    \label{eq:cov-density}
\end{equation}
recovering the defining relation between a CDF and its density. 

Under the exponential tilting of Eq.~\ref{eq:exp-tilt}, the latent measure $\pi_{q,p}$ against a base measure $\pi_0$ is described by the Radon-Nikodym derivative:
\begin{equation}
d\pi_{q,p}(z) = \frac{\exp(f_{q,p}(z))}{Z_{q,p}} d\pi_0(z),
\label{eq:radon}
\end{equation}
where $f_{q,p}(z)$ is our learned energy function and $Z_{q,p}$ is the finite normalizing constant (partition function) that ensures $\pi_{q,p}$ induces a valid probability density, $p_{q,p}$, for each latent feature. The inspiration for this exponential tilting form can be traced to the energy-based prior introduced by \citet{pang2020learninglatentspaceenergybased}, adapted to the univariate case. The base prior $\pi_0$ serves as a reference that guides the form of the learned $\pi_{q,p}$, allowing for flexible reweighting via the energy function $f_{q,p}$. For $\pi_{q,p}$ to define a valid density, which we denote $p_{q,p}$, we require exponential Lebesgue-integrability:

\begin{align}
\exp(f_{q,p}(z)) &\; \in \; L^1 \left( \pi_0 \right) \notag \\
\text{such that} \quad Z_{q,p} &\; = \; \int_{\mathcal{Z}} \exp(f_{q,p}(z)) \pi_0(dz)  < \infty.
\end{align}

As a result, the model retains a coherent probabilistic interpretation of Eq.~\ref{eq:KAT} that is consistent with the energy-based framework proposed by \citet{pang2020learninglatentspaceenergybased}, which is well suited to maximum likelihood training.

\subsection{Connection to normalizing flows}
\label{sec:flow-connection}

Eq.~\ref{eq:pushforward} defines a one-dimensional normalizing flow~\citep{pmlr-v37-rezende15} with base distribution $\mathrm{Uniform}(0,1)$, invertible map $T_{q,p} := F^{-1}_{\pi_{q,p}}$, and density given by the change-of-variables identity of Eq.~\ref{eq:cov-density}. For the mixture prior of Eq.~\ref{eq:kaem_mix}, Algorithm~\ref{alg:its} applies one such 1D map per coordinate,
\begin{equation}
    T(\bm{u}) \;=\; \big(T_{q,p^{\star}_q}(u_q)\big)_{q=1}^{Q},
    \qquad p^{\star}_q \sim \mathrm{Cat}(\alpha_{q,1:P}),
\end{equation}
so the joint density is the product of the per-coordinate densities, as in Eq.~\ref{eq:kaem_mix}.

KAEM and the rational-quadratic spline flow of \citet{durkan2019neuralsplineflows} take opposite approaches. Spline flows parameterize the invertible map directly as a monotonic spline and compute the density as $|T'|$, while KAEM parameterises the density via the exponentially-tilted KAN energy of Eq.~\ref{eq:radon} and computes $T_{q,p} = F^{-1}_{\pi_{q,p}}$ by inverting the CDF numerically (Algorithm~\ref{alg:its}). Dependence between coordinates, which spline flows introduce through coupling or autoregressive layers, is handled by KAEM's non-invertible CNN decoder of Sec.~\ref{app:gen_adherent}.

\subsection{Kolmogorov-Arnold Network generator}
\label{app:gen_adherent}

Having sampled the intermediary latent variable with ITS such that $z_{q,p} = F^{-1}_{\pi_{q,p}}(u_p)$, the generator can be realized as:
\begin{align}
\boldsymbol{\tilde{x}} &\;=\; G_\Phi(\boldsymbol{z}) + \boldsymbol{\varepsilon} ,  \quad \text{where} \quad \boldsymbol{z} \sim p_{f}(\boldsymbol{z}), \quad \boldsymbol{\varepsilon} \sim \mathcal{N}(\mathbf{0},\sigma_{\text{noise}}\mathbf{I}),
\end{align}
where $\boldsymbol{\tilde{x}}=\{\tilde{x}_o\}_{o=1}^{n_x}$ represents a generated data sample and $n_x$ is used to specify the dimension of $\boldsymbol{\tilde{x}}$. The architecture may be made strictly adherent to KART using a Kolmogorov-Arnold Network (KAN):
\begin{align}
    \tilde{x}_o &= \sum_{q=1}^{2n_{z}+1} \Phi_{o,q}\left( \sum_{p=1}^{n_z}z_{q,p}\right) + \varepsilon_o, \quad
    \boldsymbol{\tilde{x}} = \begin{pmatrix}
    \sum_{q=1}^{2n_{z}+1} \Phi_{1,q}\left( \sum_{p=1}^{n_z}z_{q,p} \right)  \\[0.5em]
    \sum_{q=1}^{2n_{z}+1} \Phi_{2,q}\left( \sum_{p=1}^{n_z}z_{q,p} \right) \\[0.5em]
    \vdots \\[0.5em]
    \sum_{q=1}^{2n_{z}+1} \Phi_{n_x,q}\left( \sum_{p=1}^{n_z}z_{q,p} \right) 
    \end{pmatrix} + \boldsymbol{\varepsilon}.
    \label{eq:generator}
\end{align}
The functions, \(\Phi_{o,q}: \mathbb{R} \to \mathbb{R}\), are realizable as basis or KAN functions similar to $f_{q,p}: \mathcal{Z} \to \mathbb{R}$.

Deepening this generator, as guided by \citet{liu2024kankolmogorovarnoldnetworks} for KANs, does not replace KART as the inner part of KAEM:
\begin{align} 
    x_o = &\sum_{l_n=1}^{L_n} h_{o,l_n} \Bigg( \dots h_{l_3, l_2} \Bigg(\sum_{l_1=1}^{L_1} h_{l_2, l_1}
    \Bigg(\sum_{q=1}^{2n_{z}+1} \Phi_{l_1,q} \Bigg( \sum_{p=1}^{n_z} z_{q,p} \Bigg) \Bigg) \Bigg) \Bigg) + \varepsilon_o. 
     \label{eq:KAN}
\end{align}    
Here, each $h$ represents a hidden layer introduced between the outer sum of Eq.~\ref{eq:KAT}, and the output feature space to allow for more expressivity when required. In fact, each successive instance of $h$ can be interpreted as a distinct outer sum of Eq.~\ref{eq:KAT}, where the inner sum corresponds to the previous layer. This is valid provided that:  
\begin{align}
L_1 &= 2 \left( 2n_z+1 \right) + 1, \quad L_2 = 2L_1 + 1, \quad \dots \quad L_n = 2L_{n-1} + 1.
\end{align}
Thus, a smooth transition between the latent and data space can be achieved by progressively doubling widths, eliminating the need for arbitrary choices while maintaining compliance with KART. Strict adherence to KART can be achieved for hidden layer $h_{l_k,l_{k-1}}$ by applying either a sigmoid or affine transformation after layer $h_{l_{k-2},l_{k-3}}$ to shift its range back to \([0,1]^{L_{k-1}}\). However, using sigmoid may lead to vanishing gradients and we suggest that function domains are less critical to KART than its structural bias and should not be prioritized over maintaining gradient flow and avoiding saturation.

\section{Importance sampling}
\label{sec:sampling-theory}

\subsection{Theory}
\label{sec:importance-sampling-theory}

Importance Sampling (IS) is a Monte Carlo method that enables the estimation of expectations with respect to one probability measure using samples drawn from another. For a detailed exposition, see \citet{910e8e7d-2e70-317c-adf6-e21cd50755a1}.

\paragraph{The Radon-Nikodym derivative} IS involves rewriting the expectation of an arbitrary function or density, \( \rho(\boldsymbol{z}) \), under the target measure, \( \probP(\boldsymbol{z}) \), in terms of a more tractable proposal measure, \( \probQ(\boldsymbol{z}) \). Specifically, we use the Radon-Nikodym derivative \( \frac{d\probP}{d\probQ} \) to rewrite the expectation of an arbitrary function of $\boldsymbol{z}$, denoted as $\rho(\boldsymbol{z})$:
\begin{equation}
    \mathbb{E}_{\probP}[\rho(\boldsymbol{z})] = \int_{\mathcal{\bar{Z}}} \rho(\boldsymbol{z}) \, \probP(d\boldsymbol{z})=\int_{\mathcal{\bar{Z}}} \rho(\boldsymbol{z}) \cdot \frac{d\probP}{d\probQ} \, \probQ(d\boldsymbol{z}).
    \label{eq:arb-expectation}
\end{equation}
Here, \( \frac{d\probP}{d\probQ} \) acts as a weight that adjusts the contribution of each sample to account for the difference between the target and proposal measures. This requires that the proposal measure \( \probQ \) is absolutely continuous with respect to the target measure \( \probP \), i.e. \( \probQ(A)=0 \) implies \( \probP(A)=0 \), for every measurable subset \( A \in \mathcal{Z} \).

In the context of KAEM, the target density is the posterior, \( p_{f, \Phi}(\boldsymbol{z} \mid \boldsymbol{x}) \), which is proportional to the product of the likelihood and the prior:
\begin{equation}
    p_{f, \Phi}(\boldsymbol{z} \mid \boldsymbol{x}) \propto p_{\Phi}(\boldsymbol{x} \mid \boldsymbol{z}) \cdot p_{f}(\boldsymbol{z}).
    \label{eq:propto-bayes}
\end{equation}
The exact normalization constant, (the marginal likelihood \( p_{f, \Phi}(\boldsymbol{x}) \)), is intractable, but IS circumvents the need to compute it explicitly. Instead, the expectation of \( \rho(\boldsymbol{z}) \) under the posterior can be expressed as:
\begin{equation}
\mathbb{E}_{p_{f, \Phi}(\boldsymbol{z} \mid \boldsymbol{x})}[\rho(\boldsymbol{z})] = \int_{\mathcal{\bar{Z}}} \rho(\boldsymbol{z}) w(\boldsymbol{z}) \, q(\boldsymbol{z}) d\boldsymbol{z},
\label{eq:importance-expec}
\end{equation}
where the importance weights \( w(\boldsymbol{z}) \) are given by:
\begin{equation}
w(\boldsymbol{z}) = \frac{p_{f, \Phi}(\boldsymbol{z} \mid \boldsymbol{x})}{q(\boldsymbol{z})}=\frac{p_{\Phi}(\boldsymbol{x} \mid \boldsymbol{z}) \cdot p_{f}(\boldsymbol{z})}{q(\boldsymbol{z})}.
\label{eq:bayes-weight}
\end{equation}

\paragraph{Proposal distribution and importance weights} A practical choice for the proposal density \( q(\boldsymbol{z}) \) is the prior, \( p_{f}(\boldsymbol{z}) \), given the availability of draws from the prior using the method outlined in Algorithm~\ref{alg:its}. This simplifies the importance weights, as the prior has a tractable form and covers the support of the posterior. This yields:
\begin{equation}
    w(\boldsymbol{z}^{(s)}) = p_{\Phi}(\boldsymbol{x} \mid \boldsymbol{z}^{(s)}) \propto \exp(\log(p_{\Phi}(\boldsymbol{x} \mid \boldsymbol{z}^{(s)
    }) )).
    \label{eq:weight-likelihood}
\end{equation}
This reflects that the likelihood now directly informs how much weight to assign to each sample \( \boldsymbol{z} \). Intuitively, samples that better explain the observed data \( \boldsymbol{x} \), (as measured by the likelihood), are given higher importance. Using the definition of softmax, the normalized weights are given by:
\begin{equation}
    \frac{w(\boldsymbol{z}^{(s)})}{\sum_{r=1}^{N}w(\boldsymbol{z}^{(r)})} = \text{softmax}_s\Big( \log(p_{\Phi}(\boldsymbol{x} \mid \boldsymbol{z}^{(s)
    }) ) \Big).
\end{equation}

\paragraph{Resampling} The suitability of these weights depends on how well \( q(\boldsymbol{z}) \) matches the posterior. In KAEM, there is likely to be significant mismatch due to the complexity of the likelihood used in Eq.~\ref{eq:gaussian_loglik}. Consequently, the weights will exhibit high variance, resulting in many samples contributing minimally. This reduces the Effective Sample Size (ESS) and introduces bias into the estimates. 

To address this, the weights are resampled before proceeding with the posterior expectation. This is accomplished by using one of the methods outlined by \citet{douc2005comparisonresamplingschemesparticle}, whenever the ESS falls below a certain threshold determined by $\text{ESS} < \gamma * N$, where $\gamma$ is a hyperparameter controlling resample rate and $N$ is the number of samples. Resampling redistributes the sample weights, creating a new population with uniformly distributed weights while preserving the statistical properties of the original distribution. As training progresses and the prior becomes a better match for the posterior, the need for resampling decreases.

After resampling, the corresponding weights will be uniformly distributed and a standard Monte Carlo estimator can be applied instead of the IS estimator.

\paragraph{Importance Sampling estimator} To estimate the expectation, we draw \( N \) independent samples \( \{\boldsymbol{z}^{(s)}\}_{s=1}^{N} \sim q(\boldsymbol{z}) = p_f(\boldsymbol{z}) \) using the procedure outlined in Algorithm~\ref{alg:its} and compute a weighted sum:
\begin{equation}
    \mathbb{E}_{p_{f, \Phi}(\boldsymbol{z} \mid \boldsymbol{x})}[\rho(\boldsymbol{z})] \approx \sum_{s=1}^{N} \rho(\boldsymbol{z}^{(s)}) \cdot w_{\text{norm}}(\boldsymbol{z}^{(s)}) \approx \frac{1}{N} \sum_{s=1}^{N} \rho(\boldsymbol{z}^{(s)}_{\text{resampled}}).
\end{equation}

The estimator is unbiased, implying that its expectation matches the true posterior expectation. As \( N \to \infty \), the approximation converges almost surely to the true value, provided that \( q(\boldsymbol{z}) \) sufficiently covers the posterior to ensure a well-conditioned Radon-Nikodym derivative. This estimator is used for the log-marginal likelihood and its gradient in Eqs. \ref{eq:learning-grad}, since it provides a means of estimating expectations with respect to the posterior. 

\subsection{Residual resampling}
\label{sec:residual-resampling}

In this study, we adopted residual resampling to filter the population of latent samples whenever the effective sample size (ESS) criterion in Algorithm~\ref{alg:importance_sampling} is satisfied. Given a set of normalized weights $\{w_{\text{norm}}(\boldsymbol{z}^{(s)})\}_{s=1}^{N}$, the method deterministically replicates samples according to the integer part of $N w_{\text{norm}}$, and uses multinomial sampling to allocate the remaining draws according to the residual mass.

\begin{algorithm}[t]
\caption{Residual resampling (ESS-triggered)}
\label{alg:residual-resampling}
\begin{algorithmic}[1]
\REQUIRE Latent samples $\{z^{(s)}\}_{s=1}^N$ with normalized likelihood-based weights $\{w_s\}_{s=1}^N$ ($\sum_{s=1}^N w_s=1$);
threshold fraction $\gamma\in(0,1]$.
\ENSURE Resampled $\{\tilde z^{(i)}\}_{i=1}^N$ and weights $\{\tilde w_i\}_{i=1}^N$.

\STATE \textbf{ESS gating:} $\mathrm{ESS}\leftarrow\Big(\sum_{s=1}^N w_s^2\Big)^{-1}$.
\IF{$\mathrm{ESS}\ge \gamma N$}
  \FOR{$i=1$ to $N$}
    \STATE $\tilde z^{(i)} \leftarrow z^{(i)}$
    \STATE $\tilde w_i \leftarrow w_i$
  \ENDFOR
  \STATE \textbf{return} $\{(\tilde z^{(i)},\tilde w_i)\}_{i=1}^N$ \COMMENT{no resampling}
\ENDIF

\STATE \textbf{Integer replication and residual mass:}
\FOR{$s=1$ to $N$}
  \STATE $r_s \leftarrow \lfloor N w_s \rfloor$
  \STATE $m_s \leftarrow N w_s - r_s$ \COMMENT{residual mass}
\ENDFOR
\STATE $R \leftarrow N - \sum_{s=1}^N r_s$ \COMMENT{\# residual draws}

\STATE \textbf{Initialize index list:} $\mathcal{I}\leftarrow [\,]$.
\FOR{$s=1$ to $N$}
  \STATE Append $s$ to $\mathcal{I}$ exactly $r_s$ times
\ENDFOR
\COMMENT{$|\mathcal{I}| = N-R$}

\IF{$R>0$}
  \STATE \textbf{Normalize residual weights:}
  \FOR{$s=1$ to $N$}
    \STATE $\bar w_s \leftarrow m_s / R$
  \ENDFOR

  \STATE \textbf{Build CDF:} $C_0\leftarrow 0$
  \FOR{$s=1$ to $N$}
    \STATE $C_s \leftarrow C_{s-1} + \bar w_s$
  \ENDFOR

  \STATE \textbf{Draw remaining $R$ indices (multinomial):}
  \FOR{$i=1$ to $R$}
    \STATE Sample $u\sim\mathrm{Unif}(0,1)$
    \STATE Find smallest $k^\star$ such that $C_{k^\star}\ge u$
    \STATE Append $k^\star$ to $\mathcal{I}$
  \ENDFOR
\ENDIF

\STATE \textbf{Resample and reset weights:}
\FOR{$i=1$ to $N$}
  \STATE $\tilde z^{(i)} \leftarrow z^{(\mathcal{I}_i)}$
  \STATE $\tilde w_i \leftarrow 1/N$
\ENDFOR
\STATE \textbf{return} $\{(\tilde z^{(i)},\tilde w_i)\}_{i=1}^N$.
\end{algorithmic}
\end{algorithm}

\section{Guidance on annealing}
\label{sec:thermo-help}

\subsection{Power-law temperature schedule}
\label{sec:optimal-temp}

\begin{figure}[H]
    \centering
    \includegraphics[width=0.7\linewidth]{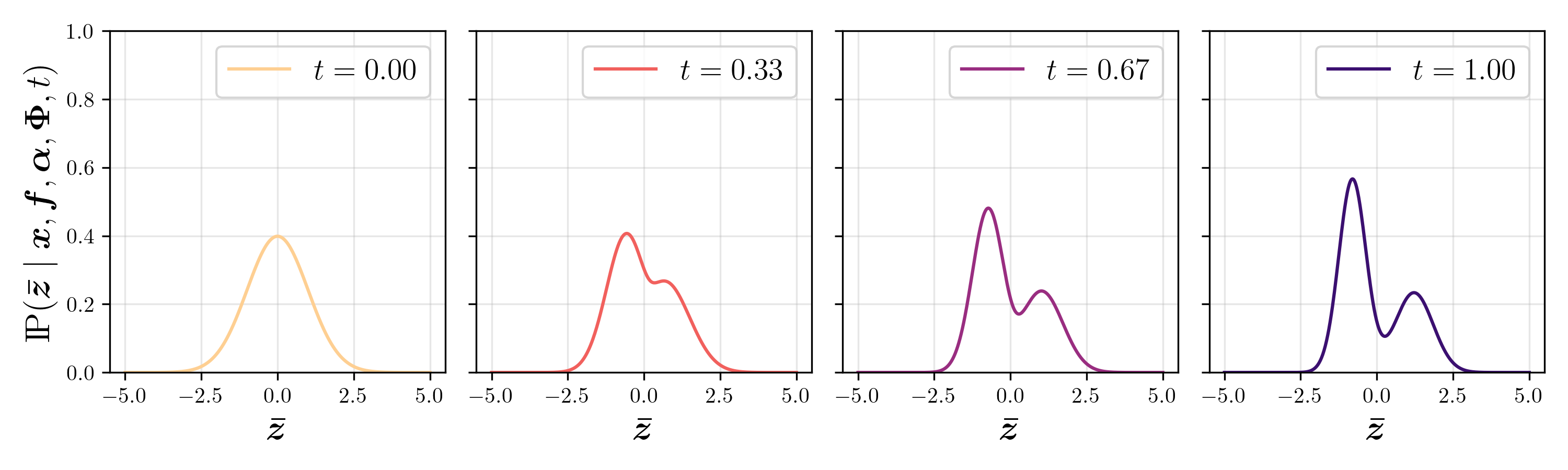}
    \caption{Demonstration of power posterior evolution with $t$. A simple Gaussian prior is sculpted into forms that resemble a more complex, multi-modal posterior as $t$ is incremented from 0 to 1.}
    \label{fig:power-posterior}
\end{figure}

Given the form of Eq.~\ref{eq:powerposterior}, setting $t=0$ results in the power posterior density assuming the form of the prior. As $t$ increments, the power posterior gradually adopts a shape more closely resembling the true Bayesian posterior. At $t=1$, the power posterior converges fully on the original posterior. The effect of tempering is illustrated in Fig.~\ref{fig:power-posterior}, starting with a Gaussian prior that is entirely uninformed of $\boldsymbol{x}$. As $t$ increases from 0 to 1, the contribution of the likelihood term in Eq.~\ref{eq:powerposterior} progressively grows, ultimately leading to the posterior at $t=1$. This final posterior can be interpreted as the prior updated by information from the samples of $\boldsymbol{x}$.

\citet{CALDERHEAD20094028} demonstrated that uniformly scheduling the discretized temperatures in Eq.~\ref{eq:tempsarray} results in unequal contributions of adjacent power posteriors to the Thermodynamic Integral. SE is an application of importance sampling, (as derived in Appendix~\ref{sec:SS-IS}), and its success relies on selecting each $t_k$ to ensure sufficient overlap of adjacent power posteriors. Similarly, the trapezoid rule in Eq.~\ref{eq:SS-KAEM} benefits through reducing discretization error. This overlap is quantified by the KL divergence term in Eq.~\ref{eq:disc}. Maintaining small KL divergences between adjacent power posteriors can be achieved by clustering temperatures in regions where \( p_{f, \Phi}(\boldsymbol{z} \mid \boldsymbol{x}, t_k) \) changes rapidly. To accomplish this, we follow \citet{CALDERHEAD20094028} in adopting a power-law relationship to schedule the temperatures of Eq.~\ref{eq:tempsarray}:

\begin{figure}[H]
    \centering
    \begin{minipage}[c]{0.49\linewidth}
        \vspace*{\fill}
        \begin{equation}
            t_k = \left( \frac{k}{N_t} \right)^p, \quad k = 0, 1, \ldots, N_t, 
            \label{eq:power-law}
        \end{equation}\\[0.5em]
        The effect of varying \( p \) on the schedule between 0 and 1 is visualized in Fig.~\ref{fig:schedule}. When \( p > 1 \), spacing becomes denser near \( t = 0 \), while \( p < 1 \) creates denser spacing near \( t = 1 \). The impact of varying \( p \) on the discretized Thermodynamic Integral of Eq.~\ref{eq:disc} is illustrated in Fig.~\ref{fig:1x3_trapezoids}. The plots demonstrate that for \( p > 1 \), the trapezoidal bins are clustered closer to \( t = 0 \), concentrating the discrete approximation error toward \( t = 1 \).
        \vspace*{\fill}
    \end{minipage}
    \hfill
    \begin{minipage}[c]{0.45\linewidth}
        \centering
        \includegraphics[width=\linewidth]{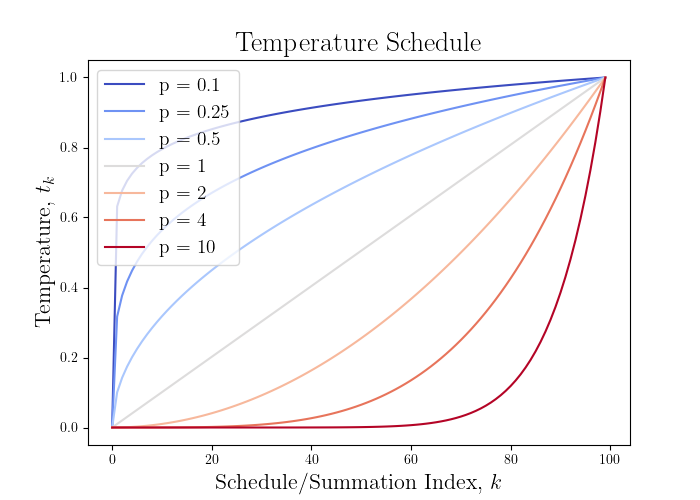}
        \captionsetup{justification=centerlast}
        \caption{Power-law schedule.}
        \label{fig:schedule}
    \end{minipage}
\end{figure}

\begin{figure}[H]
    \centering
    
    \begin{subfigure}{0.3\linewidth}
        \centering
        \includegraphics[width=\linewidth]{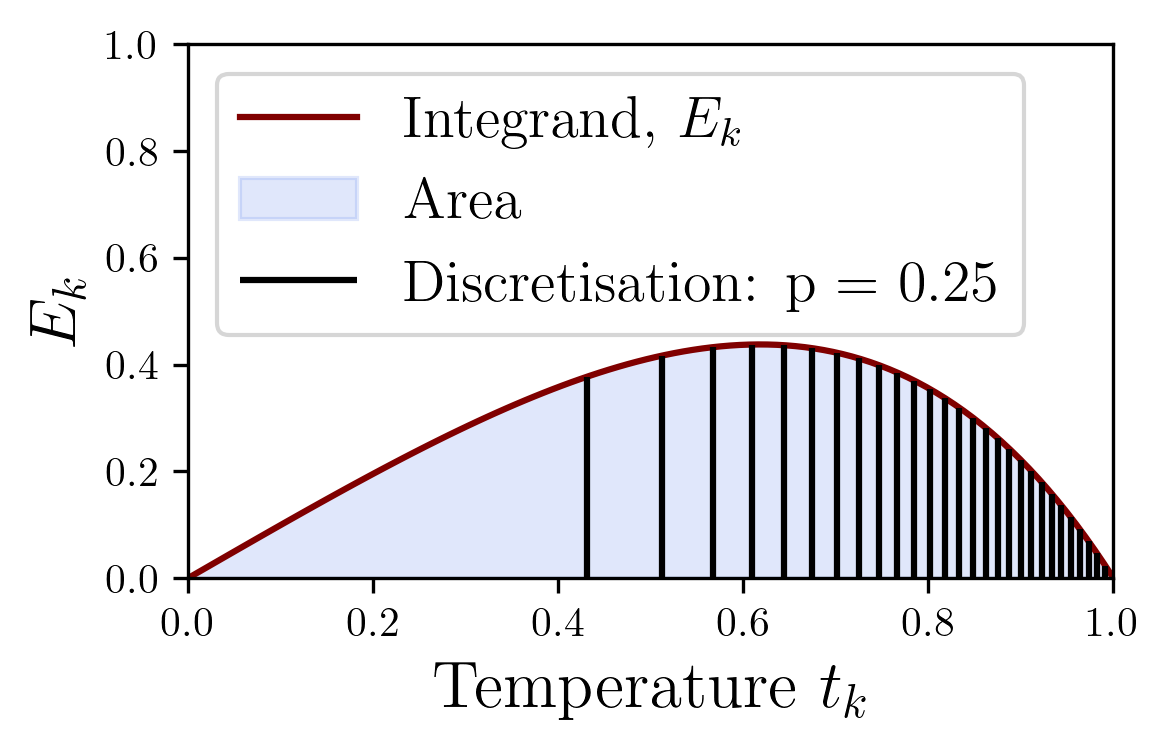}
        \caption{$p=0.25$ schedule has more bins clustered toward $t=1$.}

        \label{fig:tsuba}
    \end{subfigure}\hfill
    \begin{subfigure}{0.3\linewidth}
        \centering
        \includegraphics[width=\linewidth]{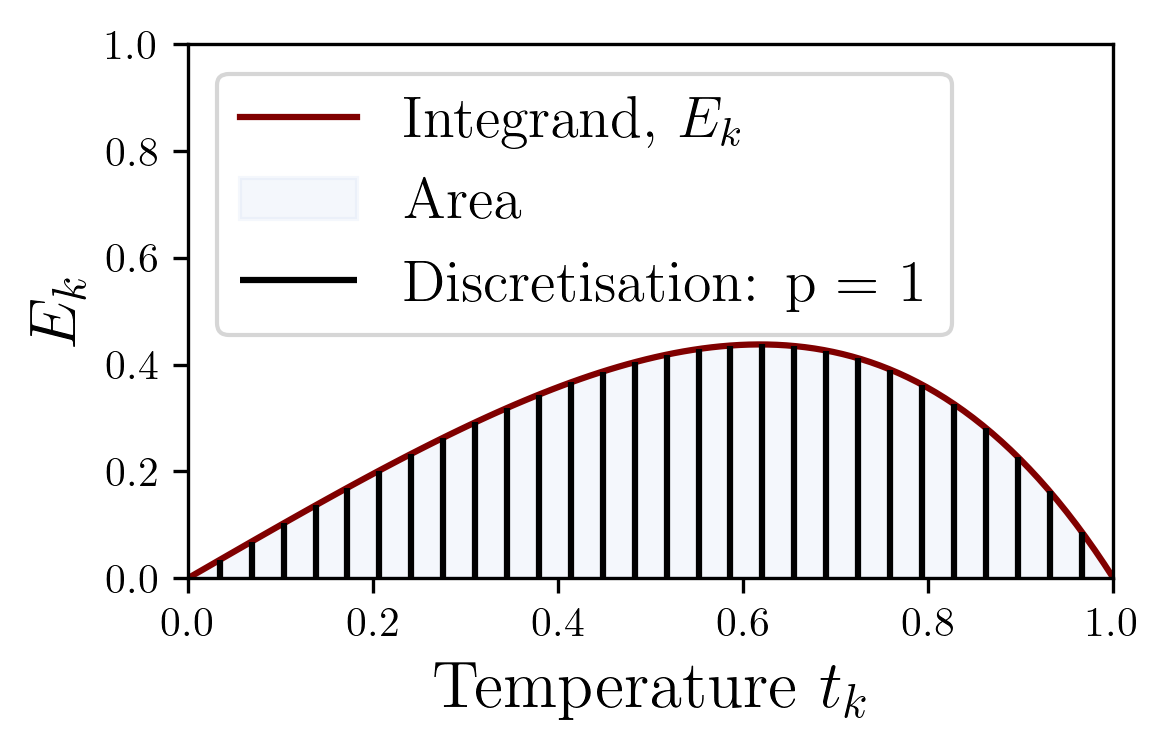}
        \caption{$p=1$ schedule is uniformly distributed between the bounds.}
        \label{fig:tsubb}
    \end{subfigure}\hfill
    \begin{subfigure}{0.3\linewidth}
        \centering
        \includegraphics[width=\linewidth]{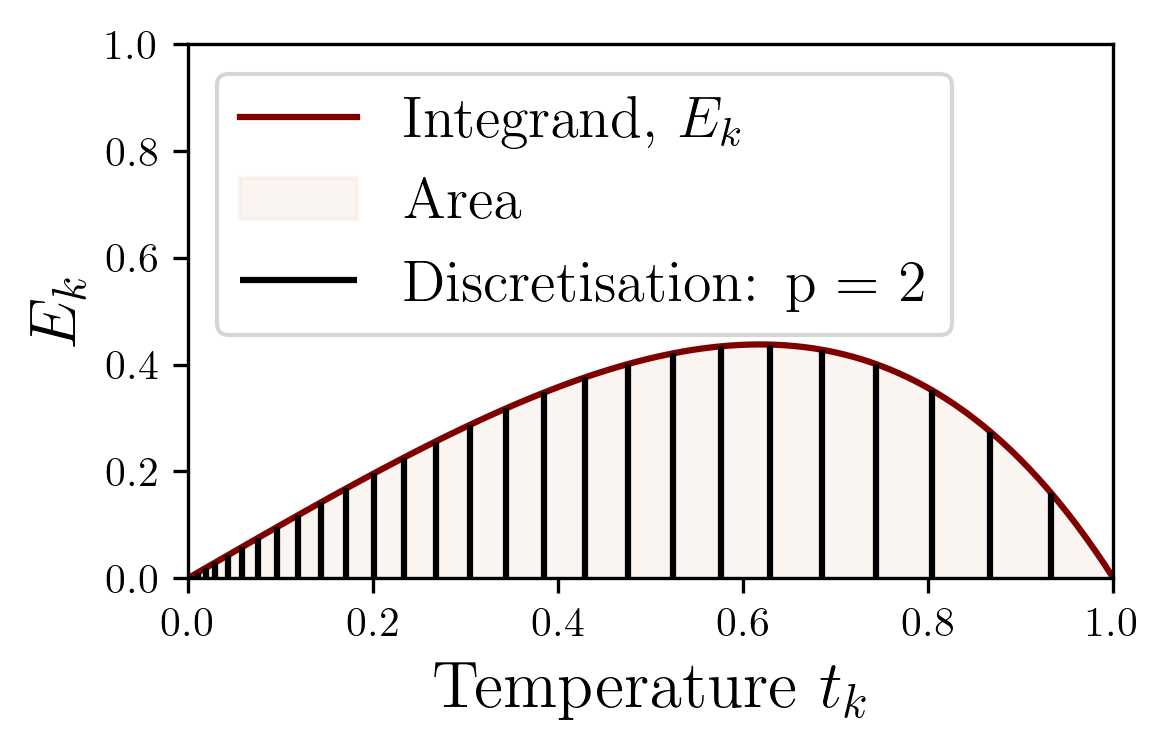}
        \caption{$p=2$ schedule has more bins clustered toward $t=0$.}
        \label{fig:tsubc}
    \end{subfigure}
    \caption{Effect of power-law schedule clusters on integral approximation error. Evaluation points can be skewed toward a particular bound by choosing $p$. Increasing \( p \) provides more tempering, concentrating more evaluation points on smoother likelihoods. Tempering helps to mitigate practical challenges associated with the non-smoothness of the univariate functions in Eq.~\ref{eq:KAT}. }
    \label{fig:1x3_trapezoids}
\end{figure}

The study by \citet{CALDERHEAD20094028} derives an analytic term for an optimal tempering distribution in the context of linear regression models, where the power posterior in Eq.~\ref{eq:powerposterior} is used to evaluate the plausibility of model parameters in relation to the data, rather than reflecting a latent variable posterior. However, their findings may not hold directly for KAEM and generative modeling. Inspired by the cyclical annealing schedules proposed by \citet{fu2019cyclicalannealingschedulesimple} for mitigating KL vanishing in VAEs, we adopt a similar cosine-based strategy to anneal the power-law exponent $p$ during training:
\begin{equation}
    p(i) = p_{\text{start}} + (p_{\text{end}} - p_{\text{start}}) \cdot \frac{1}{2}\left(1 - \cos\!\left(2\pi \left(N_{\text{cycles}} + \tfrac{1}{2}\right) \cdot \frac{i-1}{N_{\text{updates}}}\right)\right),
    \label{eq:p-anneal}
\end{equation}
where $i$ is the current parameter update, $N_{\text{updates}}$ is the total number of parameter updates, and $N_{\text{cycles}}$ controls the number of cosine annealing cycles. Setting $N_{\text{cycles}}=0$ yields a single half-cosine that monotonically transitions $p$ from $p_{\text{start}}$ to $p_{\text{end}}$ over training.

\begin{figure}[H]
    \centering
    \includegraphics[width=0.6\linewidth]{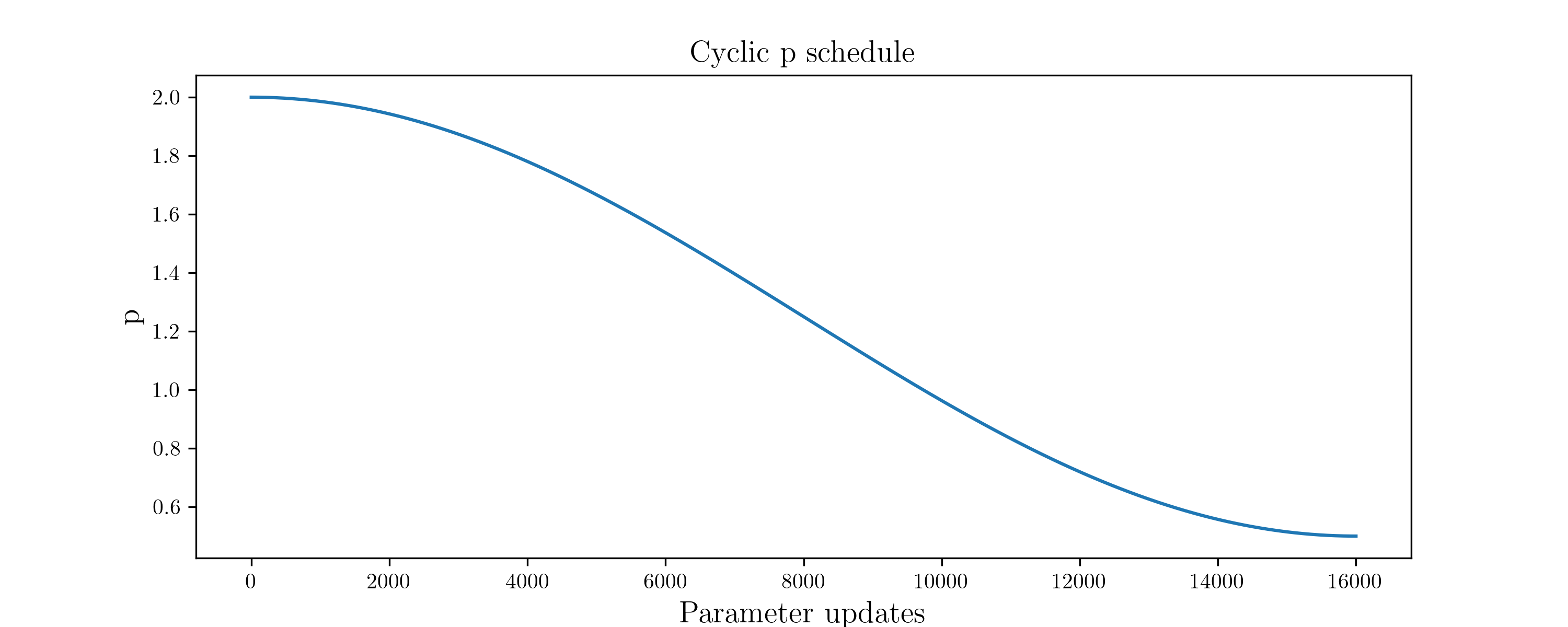}
    \caption{Annealing schedule for the power-law exponent $p$ over training, decreasing from $p_{\text{start}}=2$ to $p_{\text{end}}=0.5$ with $N_{\text{cycles}}=0$.}
    \label{fig:p-schedule}
\end{figure}

Initializing with $p_{\text{start}} > 1$ concentrates trapezoid bins closer to $t=0$, focusing early learning on low-variance, smoother power posteriors where global features emerge between adjacent power posteriors. As training progresses, KAEM learns these global features sufficiently well, reducing the change in density between adjacent power posteriors of a $p > 1$ schedule. At this stage, $p(i)$ transitions toward $p_{\text{end}} \leq 1$, shifting focus toward capturing more high-variance, data-informed features of the posterior landscape, where training is expected to enhance the model's ability to generate finer details. In this study, we set $p_{\text{start}}=2$ and $p_{\text{end}}=0.5$ with $N_{\text{cycles}}=0$, yielding the schedule shown in Fig.~\ref{fig:p-schedule}.

\subsection{Choosing hyperparameters}
\begin{table}[H]
\centering
\caption{Tempering hyperparameters.}
\label{tab:hyperparameters}
\renewcommand{\arraystretch}{1.5} 
\begin{tabular}{|p{0.2\textwidth}|p{0.75\textwidth}|}
\hline
\textbf{Hyperparameter} & \textbf{Description} \\ \hline
${N_t}>1$ & Number of power posteriors transitioning between prior and posterior. Increasing improves exploration but also increases cost due to the need for more ULA iterations. Typically, this is memory-bound and guided by hardware availability, such as \( N_{t} = 10 \). \\ \hline
$N_{\text{local}}>1$ & Number of ULA iterations per power posterior. Although increasing this value improves the quality of samples at each stage, it also raises the computational cost, much like $N_t$. It can be kept at a small value, such as \( N_{\text{local}} = 40 \), (based on the recommendations of \citet{nijkamp2019learningnonconvergentnonpersistentshortrun, nijkamp2019anatomymcmcbasedmaximumlikelihood} regarding short-run Markov chain Monte Carlo). \\ \hline
$p_{\text{start}}$ & Initial power-law exponent. Values $p > 1$ cluster trapezoid bins closer to $t=0$, focusing early training on smoother power posteriors where global features emerge. We set $p_{\text{start}}=2$. \\ \hline
$p_{\text{end}}$ & Final power-law exponent. Values $p < 1$ cluster bins closer to $t=1$, shifting focus to high-variance, data-informed features for finer details. We set $p_{\text{end}}=0.5$. \\ \hline
$N_{\text{cycles}}$ & Number of cosine annealing cycles (Eq.~\ref{eq:p-anneal}). Setting $N_{\text{cycles}}=0$ yields a single monotonic transition from $p_{\text{start}}$ to $p_{\text{end}}$. \\ \hline
\end{tabular}
\end{table}

\section{Derivations and useful identities}
\label{app:derivations}

\subsubsection{Useful proof}
\label{sec:useful-proof}

The following proof for a gradient with respect to an arbitrary set of parameters, $\boldsymbol{\theta}$, proves useful for other derivations:
\begin{align}
    \mathbb{E}_{p_{\boldsymbol{\theta}}(\boldsymbol{x})}\left[ \nabla_{\boldsymbol{\theta}} \log p_{\boldsymbol{\theta}}(\boldsymbol{x}) \right] 
    &= \mathbb{E}_{p_{\boldsymbol{\theta}}(\boldsymbol{x})}\left[ \frac{\nabla_{\boldsymbol{\theta}} p_{\boldsymbol{\theta}}(\boldsymbol{x})}{p_{\boldsymbol{\theta}}(\boldsymbol{x})} \right] = \int_{\mathcal{X}} \frac{\nabla_{\boldsymbol{\theta}} p_{\boldsymbol{\theta}}(\boldsymbol{x})}{p_{\boldsymbol{\theta}}(\boldsymbol{x})} \, \, \, p_{\boldsymbol{\theta}}(\boldsymbol{x}) d\boldsymbol{x} \notag\\[0.5em]
    &= \int_{\mathcal{X}} \nabla_{\boldsymbol{\theta}} p_{\boldsymbol{\theta}}(\boldsymbol{x}) \, d\boldsymbol{x} 
    = \nabla_{\boldsymbol{\theta}} \int_{\mathcal{X}} p_{\boldsymbol{\theta}}(\boldsymbol{x}) \, d\boldsymbol{x} = \nabla_{\boldsymbol{\theta}} 1 = 0.
    \label{eq:useful-proof}
\end{align}

\subsubsection{Marginal likelihood}
\label{sec:marginal-llhood}

\begin{align}
  \log \left( p_{f, \Phi} (\boldsymbol{x}) \right) &= \mathbb{E}_{p_{f, \Phi}(\boldsymbol{z} \mid \boldsymbol{x})}\left[\log \left(  p_{f, \Phi} (\boldsymbol{x},\boldsymbol{z}) \right)\right] = \mathbb{E}_{p_{f, \Phi}(\boldsymbol{z} \mid \boldsymbol{x})} \left[\log \left(   p_{f} \left(\boldsymbol{z}\right) \cdot p_{\Phi} \left(\boldsymbol{x}\mid\boldsymbol{z} \right) \right) \right] \notag \\ &= \mathbb{E}_{p_{f, \Phi}(\boldsymbol{z} \mid \boldsymbol{x})}\left[\log \left(   p_{f} \left(\boldsymbol{z}\right) \right) + \log \left( p_{\Phi} \left(\boldsymbol{x}\mid\boldsymbol{z} \right) \right) \right].
\end{align}

\subsubsection{Maximum likelihood learning gradient}
\label{sec:MLE-lr}

This proof demonstrates that the gradient with respect to an arbitrary set of parameters, $\theta$, is independent of sampling:
\begin{align}
    \mathbb{E}_{p(\boldsymbol{z} \mid \boldsymbol{x})}\left[ \nabla_{\boldsymbol{\theta}} \log p(\boldsymbol{x}, \boldsymbol{z}) \right]
    &= \mathbb{E}_{p(\boldsymbol{z} \mid \boldsymbol{x})}\left[ \nabla_{\boldsymbol{\theta}} \log p(\boldsymbol{z} \mid \boldsymbol{x}) + \nabla_{\boldsymbol{\theta}} \log p(\boldsymbol{x}) \right] \notag \\
    &= \mathbb{E}_{p(\boldsymbol{z} \mid \boldsymbol{x})}\left[ \nabla_{\boldsymbol{\theta}} \log p(\boldsymbol{z} \mid \boldsymbol{x}) \right] + \mathbb{E}_{p(\boldsymbol{z} \mid \boldsymbol{x})}\left[\nabla_{\boldsymbol{\theta}} \log p(\boldsymbol{x}) \right] \notag \\
    &= 0 + \nabla_{\boldsymbol{\theta}} \log p(\boldsymbol{x}),
\end{align}
where Eq.~\ref{eq:useful-proof} has been used to zero the expected posterior. This allows gradient flow to be disregarded in sampling procedures as shown in Eq.~\ref{eq:learning-grad}.

\subsubsection{Contrastive divergence learning gradient for univariate log-prior}
\label{app:prior-grad}
The learning gradient attributed to the univariate KAEM prior in Eq.~\ref{eq:exp-tilt} can be derived into a contrastive divergence framework, which is more typical of energy-based modeling literature. The prior model is solely dependent on $f$ in Eq.~\ref{eq:exp-tilt}. We can rewrite Eq.~\ref{eq:log-prior} as:

\begin{align}
    \nabla_{f} \log \left( p_{f} \left(\boldsymbol{z}\right) \right) &= \nabla_{f} \sum_{q=1}^{2n_z+1} \sum_{p=1}^{n_z} \log \left( p_{q,p}(z) \right) = \nabla_{f} \sum_{q=1}^{2n_z+1} \sum_{p=1}^{n_z} f_{q,p}(z) + \log \left( \pi_0(z) \right) - \log Z_{q,p}
    \label{eq:prior-grad-1}
\end{align}
where \(z = \int_{\mathcal{Z}} \exp(f_{q,p}(z_p)) \, d\pi_0(z_p)\) is the normalization constant. We can simplify the constant using Eq.~\ref{eq:useful-proof}:
\begin{align}
    \mathbb{E}_{p_{f}(\boldsymbol{z})} 
    \Bigg[ \nabla_{f} \log p_{f}\left(\boldsymbol{z}\right) \Bigg] 
    &= \mathbb{E}_{p_{f}(\boldsymbol{z})} 
    \Bigg[ \nabla_{f} \sum_{q=1}^{2n_z+1} \sum_{p=1}^{n_z} f_{q,p}(z) + \log \left( \pi_0(z) \right) - \log Z_{q,p} \Bigg] = 0 \notag \\[0.5em]
    &= \mathbb{E}_{p_{f}(\boldsymbol{z})} 
    \Bigg[ \nabla_{f} \sum_{q=1}^{2n_z+1} \sum_{p=1}^{n_z} f_{q,p}(z) \Bigg] -  \nabla_{f} \sum_{q=1}^{2n_z+1} \sum_{p=1}^{n_z} \log Z_{q,p} = 0 \notag \\[0.5em]
    \therefore \nabla_{f} \sum_{q=1}^{2n_z+1} \sum_{p=1}^{n_z} \log Z_{q,p} &= \mathbb{E}_{p_{f}(\boldsymbol{z})} 
    \Bigg[ \nabla_{f} \sum_{q=1}^{2n_z+1} \sum_{p=1}^{n_z} f_{q,p}(z) \Bigg]
\end{align}

Therefore the log-prior gradient of Eq.~\ref{eq:prior-grad-1} can be evaluated without quadrature normalization as:
\begin{align}
    \nabla_{f} \log p_{f}\left(\boldsymbol{z}\right) &= \nabla_{f} \sum_{q=1}^{2n_z+1} \sum_{p=1}^{n_z} f_{q,p}(z) -  \mathbb{E}_{p_{f}(\boldsymbol{z})} 
    \Bigg[ \nabla_{f} \sum_{q=1}^{2n_z+1} \sum_{p=1}^{n_z} f_{q,p}(z) \Bigg],
\end{align}
and the learning gradient attributed to the prior is finalized into a contrastive divergence format:
\begin{align}
    \mathbb{E}_{p_{f, \Phi}(\boldsymbol{z} \mid \boldsymbol{x})} \Bigg[
        \nabla_{f} 
        \left[ \log p_{f}\left(\boldsymbol{z}\right) \right] 
    \Bigg] = \; &\mathbb{E}_{p_{f, \Phi}(\boldsymbol{z} \mid \boldsymbol{x})} \Bigg[ \nabla_{f} \sum_{q=1}^{2n_z+1} \sum_{p=1}^{n_z} f_{q,p}(z) \Bigg] - \mathbb{E}_{p_{f}(\boldsymbol{z})} 
    \Bigg[ \nabla_{f} \sum_{q=1}^{2n_z+1} \sum_{p=1}^{n_z} f_{q,p}(z) \Bigg].
\end{align}

\subsubsection{Deriving the Thermodynamic Integral}
\label{sec:TI-derived}

Starting from the power posterior:
\begin{equation}
p_{f, \Phi}(\boldsymbol{z} \mid \boldsymbol{x}, t) = \frac{p_{\Phi} \left(\boldsymbol{x}\mid\boldsymbol{z}\right)^t \, \, p_{f}(\boldsymbol{z})}{Z_{t}},
  \quad \text{with parition} \quad Z_{t} = \mathbb{E}_{p_{f}(\boldsymbol{z})}\left[ p_{\Phi} \left(\boldsymbol{x}\mid\boldsymbol{z}\right)^t \right]
\label{eq:power-post-rep}
\end{equation}
Now, differentiating the log-partition function with respect to \( t \):
\begin{align}
  \frac{\partial}{\partial t} \log(Z_t) &= \frac{1}{Z_t} \frac{\partial}{\partial t} Z_t = \frac{1}{Z_t}  \mathbb{E}_{p_{f}(\boldsymbol{z})}\left[ \frac{\partial}{\partial t}  p_{\Phi} \left(\boldsymbol{x}\mid\boldsymbol{z}\right)^t \right] \notag \\ &= \frac{1}{Z_t}  \mathbb{E}_{p_{f}(\boldsymbol{z})}\left[ \log p_{\Phi} \left(\boldsymbol{x}\mid\boldsymbol{z}\right)  \cdot p_{\Phi} \left(\boldsymbol{x}\mid\boldsymbol{z}\right)^t \right]\
= \mathbb{E}_{p_{f, \Phi}(\boldsymbol{z} \mid \boldsymbol{x}, t)} \left[ \log p_{\Phi} \left(\boldsymbol{x}\mid\boldsymbol{z}\right) \right]
\label{eq:dt-partition}
\end{align}
Integrating both sides from 0 to 1:
\begin{equation}
\int_0^1 \frac{\partial}{\partial t} \log(Z_t) \, dt = \int_0^1 \mathbb{E}_{p_{f, \Phi}(\boldsymbol{z} \mid \boldsymbol{x}, t)} \left[ \log p_{\Phi} \left(\boldsymbol{x}\mid\boldsymbol{z}\right) \right] \, dt
\end{equation}
Using the fundamental theorem of calculus, \citep{larson2008calculus}:
\begin{equation}
\int_0^1 \frac{\partial}{\partial t} \log(Z_t) \, dt = \log(Z_{t=1}) - \log(Z_{t=0}) = \int_0^1 \mathbb{E}_{p_{f, \Phi}(\boldsymbol{z} \mid \boldsymbol{x}, t)} \left[ \log p_{\Phi} \left(\boldsymbol{x}\mid\boldsymbol{z}\right) \right] \, dt
\end{equation}
The limits of the integrals can be expressed as: 
\begin{align}
    \log(Z_{t=0}) &= \log\Big(\mathbb{E}_{p_{f}(\boldsymbol{z})}\left[ p_{\Phi} \left(\boldsymbol{x}\mid\boldsymbol{z}\right)^0 \right]\Big) = \log(1) = 0 \notag \\
    \log(Z_{t=1}) &= \log \Big( \mathbb{E}_{p_{f, \Phi}(\boldsymbol{z} \mid \boldsymbol{x})} \left[ \log p_{\Phi} \left(\boldsymbol{x}\mid\boldsymbol{z}\right) \right] \Big) = \log p_{f, \Phi}(\boldsymbol{x})
\end{align}
Therefore, we obtain the log-marginal likelihood expressed as the Thermodynamic Integral:
\begin{align}
 \log(Z_{t=1}) - \log(Z_{t=0}) = \log p_{f, \Phi}(\boldsymbol{x}) = \int_0^1 \mathbb{E}_{p_{f, \Phi}(\boldsymbol{z} \mid \boldsymbol{x}, t)} \left[ \log p_{\Phi} \left(\boldsymbol{x}\mid\boldsymbol{z}\right) \right] \, dt
\end{align}

\subsubsection{Discretizing the Thermodynamic Integral}
\label{sec:disc-ti}

Here, $t_k$ denotes the tempering at the $k$-th index of the schedule, and $N_{\text{t}}$ is the number of temperatures. As derived by \citet{CALDERHEAD20094028}, Eq.~\ref{eq:thermo} can then be evaluated using:
\begin{equation}
\log \left( p_{f, \Phi} (\boldsymbol{x}) \right) = \frac{1}{2} \sum_{k=1}^{N_t} \Delta t_k (E_{k-1} + E_{k}) + \frac{1}{2} \sum_{k=1}^{N_t} D_{\text{KL}}(p_{t_{k-1}}||p_{t_k}) - D_{\text{KL}}(p_{t_k}||p_{t_{k-1}}) 
\label{eq:disc}
\end{equation}
Where:
\begin{align}
\Delta t_k &= t_k - t_{k-1}, \quad \quad
E_k = \mathbb{E}_{p_{f, \Phi}(\boldsymbol{z} \mid \boldsymbol{x},t_k)}\left[ \log \left( p_{\Phi} \left(\boldsymbol{x}\mid\boldsymbol{z} \right) \right) \right] ,\notag \\[0.5em]
D_{\text{KL}}(p_{t_{k-1}}||p_{t_k}) &= \mathbb{E}_{p_{f, \Phi}(\boldsymbol{z} \mid \boldsymbol{x},t_{k-1})} \Bigg[  \log \Bigg( \frac{p_{f, \Phi}(\boldsymbol{z} \mid \boldsymbol{x},t_{k-1})}{p_{f, \Phi}(\boldsymbol{z} \mid \boldsymbol{x},t_k)} \Bigg) \Bigg]
\label{eq:kl-div}
\end{align}
This reflects the trapezium rule for numerical integration, supported by bias correction to provide an estimate of the Riemann integral across $t$ that is especially robust to error. The learning gradient derived from the discretized Thermodynamic Integral is unchanged from the MLE gradient in Eq.~\ref{eq:learning-grad}, as demonstrated under Sec.~\ref{sec:TI-learning-grad}.

\subsubsection{Learning gradient for discretized Thermodynamic Integral and power posterior partition function}
\label{sec:TI-learning-grad}

We can derive the learning gradient of the discretized Thermodynamic Integral in Eq.~\ref{eq:disc} by reducing it down into a sum of normalizing constants. Starting from the definitions attributed to the KL divergence terms:
\begin{align}
D_{\text{KL}}(p_{t_{k-1}} \| p_{t_k}) 
&=\mathbb{E}_{p_{f, \Phi}(\boldsymbol{z} \mid \boldsymbol{x}, t_{k-1})}\Big[\log p_{f, \Phi}(\boldsymbol{z} \mid \boldsymbol{x}, t_{k-1}) - \log p_{f, \Phi}(\boldsymbol{z} \mid \boldsymbol{x}, t_{k}) \Big]\notag \\[0.5em]
\text{and} \quad D_{\text{KL}}(p_{t_k} \| p_{t_{k-1}}) &= \mathbb{E}_{p_{f, \Phi}(\boldsymbol{z} \mid \boldsymbol{x}, t_{k})}\Big[\log p_{f, \Phi}(\boldsymbol{z} \mid \boldsymbol{x}, t_{k}) - \log p_{f, \Phi}(\boldsymbol{z} \mid \boldsymbol{x}, t_{k-1}) \Big] \notag \\
&= - \mathbb{E}_{p_{f, \Phi}(\boldsymbol{z} \mid \boldsymbol{x}, t_{k})}\Big[\log p_{f, \Phi}(\boldsymbol{z} \mid \boldsymbol{x}, t_{k-1}) - \log p_{f, \Phi}(\boldsymbol{z} \mid \boldsymbol{x}, t_{k}) \Big]
\label{eq:kl-div-subtract}
\end{align}
We can leverage Eq.~\ref{eq:powerposterior} to quantify the log-difference in terms of the change in likelihood contribution plus some normalization constant:
\begin{align}
    \log p_{f, \Phi}(\boldsymbol{z} \mid \boldsymbol{x}, t_{k-1}) &- \log p_{f, \Phi}(\boldsymbol{z} \mid \boldsymbol{x}, t_{k}) = - \Delta t_k \cdot \log p_{\Phi}(\boldsymbol{x} \mid \boldsymbol{z}) + \log \Bigg( \frac{Z_{t_k}}{Z_{t_{k-1}}} \Bigg)
\end{align} 
The definitions in Eq.~\ref{eq:kl-div} can be used to express the KL divergences in terms of the expected likelihoods:
\begin{align}
    \therefore D_{\text{KL}}(p_{t_{k-1}} \| p_{t_k})  &= - \Delta t_k \cdot \mathbb{E}_{p_{f, \Phi}(\boldsymbol{z} \mid \boldsymbol{x}, t_{k-1})}\Big[ \log p_{\Phi}(\boldsymbol{x} \mid \boldsymbol{z}) \Big] + \log \Bigg( \frac{Z_{t_k}}{Z_{t_{k-1}}} \Bigg) = -\Delta t_k \cdot E_{k-1} + \log \Bigg( \frac{Z_{t_k}}{Z_{t_{k-1}}} \Bigg)\\
    \text{and} \quad D_{\text{KL}}(p_{t_k} \| p_{t_{k-1}}) &= \Delta t_k \cdot \mathbb{E}_{p_{f, \Phi}(\boldsymbol{z} \mid \boldsymbol{x}, t_{k})}\Big[ \log p_{\Phi}(\boldsymbol{x} \mid \boldsymbol{z}) \Big] - \log \Bigg( \frac{Z_{t_k}}{Z_{t_{k-1}}} \Bigg) = \Delta t_k \cdot E_{k} - \log \Bigg( \frac{Z_{t_k}}{Z_{t_{k-1}}} \Bigg)   
\end{align}
Substituting back into Eq.~\ref{eq:disc}, we eliminate the expected likelihood terms:
\begin{align}
    \frac{1}{2}\sum_{k=1}^{N_t} D_{\text{KL}}(p_{t_{k-1}} \| p_{t_k}) &- D_{\text{KL}}(p_{t_k} \| p_{t_{k-1}}) = - \frac{1}{2}\sum_{k=1}^{N_t} \Delta t_k (E_{k-1} + E_{k}) - 2 \log \Bigg( \frac{Z_{t_k}}{Z_{t_{k-1}}} \Bigg) \notag \\[0.5em]
    \therefore \log \left( p_{f, \Phi} (\boldsymbol{x}) \right) &= \frac{1}{2} \sum_{k=1}^{N_t} \Delta t_k (E_{k-1} + E_{k}) + \frac{1}{2} \sum_{k=1}^{N_t} D_{\text{KL}}(p_{t_{k-1}}||p_{t_k}) - D_{\text{KL}}(p_{t_k}||p_{t_{k-1}}) \notag \\
    &= \frac{1}{2} \sum_{k=1}^{N_t} \Delta t_k (E_{k-1} + E_{k}) - \Bigg[ \frac{1}{2}\sum_{k=1}^{N_t} \Delta t_k (E_{k-1} + E_{k}) - 2 \log \Bigg( \frac{Z_{t_k}}{Z_{t_{k-1}}} \Bigg) \Bigg] \notag \\
    &= \sum_{k=1}^{N_t} \log \Bigg( \frac{Z_{t_k}}{Z_{t_{k-1}}} \Bigg) = \sum_{k=1}^{N_t}\log(Z_{t_k})-\log(Z_{t_{k-1}})
    \label{eq:derived-dkl+trapz}
\end{align}
Following a similar method to Sec.~\ref{app:prior-grad} we can use Eq.~\ref{eq:useful-proof} to express the learning gradient attributed to each normalization constant in a simpler manner:
\begin{align}
    &\mathbb{E}_{p_{f, \Phi}(\boldsymbol{z} \mid \boldsymbol{x}, t_{k})}\Big[\nabla_{f,\Phi} \log p_{f, \Phi}(\boldsymbol{z} \mid \boldsymbol{x}, t_{k}) \Big] \notag \\ = \, &\mathbb{E}_{p_{f, \Phi}(\boldsymbol{z} \mid \boldsymbol{x}, t_{k})}\Big[ t_k \cdot \nabla_{\Phi } \log p_{\Phi}(\boldsymbol{x} \mid \boldsymbol{z}) + \nabla_{f} \log p_{f}(\boldsymbol{z}) \Big] - \nabla_{f,\Phi} \log(Z_{t_k}) = 0 \notag \\[0.5em]
    \therefore &\nabla_{f,\Phi} \log(Z_{t_k})  = \mathbb{E}_{p_{f, \Phi}(\boldsymbol{z} \mid \boldsymbol{x}, t_{k})}\Big[ t_k \cdot \nabla_{\Phi } \log p_{\Phi}(\boldsymbol{x} \mid \boldsymbol{z}) + \nabla_{f} \log p_{f}(\boldsymbol{z}) \Big] 
    \label{eq:partition-is-MLE}
\end{align}
Telescoping the sum in Eq.~\ref{eq:derived-dkl+trapz}, and noting that $k=1$ corresponds to the power posterior equaling the prior, and $k=N_t$ to the posterior, Eq.~\ref{eq:derived-dkl+trapz} yields KAEM's learning gradient:
\begin{align}
    \nabla_{f,\Phi}\log \left( p_{f, \Phi} (\boldsymbol{x}) \right) &= \sum_{k=1}^{N_t} \nabla_{f,\Phi}\log(Z_{t_k})-\nabla_{f,\Phi}\log(Z_{t_{k-1}}) = \nabla_{f,\Phi}\log(Z_{t_{N_t}}) - \nabla_{f,\Phi}\log(Z_{t_{0}}) \notag \\
    &= \mathbb{E}_{p_{f, \Phi}(\boldsymbol{z} \mid \boldsymbol{x})} \Bigg[
        \nabla_{f} 
         \log p_{f}\left(\boldsymbol{z}\right)  + \nabla_{\Phi } 
         \log p_{ \Phi } \left(\boldsymbol{x} \mid \boldsymbol{z}\right)  
    \Bigg] - \underbrace{\mathbb{E}_{p(\boldsymbol{z} \mid f)} \Bigg[
        \nabla_{f} 
        \left[ \log p_{f}\left(\boldsymbol{z}\right) \right]
    \Bigg]}_{= 0 \; \text{(Eq.~\ref{eq:useful-proof})}}.
\end{align}

\subsubsection{Importance sampling form for Steppingstone estimator}
\label{sec:SS-IS}

The partition ratios for the Steppingstone estimator are typically approximated using importance sampling:
\begin{align}
  \frac{Z_{t_k}}{Z_{t_{k-1}}} &= \frac{\int p_{\Phi}(\boldsymbol{x} \mid \boldsymbol{z})^{t_k} \, p_{f}(\boldsymbol{z})\,d\boldsymbol{z}}{\int p_{\Phi}(\boldsymbol{x} \mid \boldsymbol{z})^{t_{k-1}} \, p_{f}(\boldsymbol{z}) \,d\boldsymbol{z}}
  = \int p_{\Phi}(\boldsymbol{x} \mid \boldsymbol{z})^{\Delta t_k} \, \frac{p_{\Phi}(\boldsymbol{x} \mid \boldsymbol{z})^{t_{k-1}} \, p_{f}(\boldsymbol{z})}{Z_{t_{k-1}}} \,d\boldsymbol{z}
  = \mathbb{E}_{p_{f, \Phi}(\boldsymbol{z} \mid \boldsymbol{x},t_{k-1})}\left[ p_{\Phi} \left(\boldsymbol{x}\mid\boldsymbol{z} \right)^{\Delta t_k} \right]
\end{align}
Therefore, the log-partition ratios are estimated with the Monte Carlo estimator:
\begin{align}
  \log \left( \frac{\hat{Z}_{t_k}}{Z_{t_{k-1}}} \right) &\approx \log \left( \frac{1}{N} \sum_{s=1}^N p_\Phi(\boldsymbol{x} \mid \boldsymbol{z}^{(s)})^{\Delta t_k} \right)
  = \mathrm{logsumexp}_{s} \Big( \Delta t_k \log p_\Phi(\boldsymbol{x} \mid \boldsymbol{z}^{(s)}) \Big) - \log N,
\end{align}
where $\boldsymbol{z}^{(s)} \sim p_{f, \Phi}(\boldsymbol{z} \mid \boldsymbol{x}, t_{k-1})$. This form is analytically accurate, but results in learning being concentrated on single samples, similar to the variance problems previously discussed in Sec.~\ref{sec:importance-sampling-theory}.

\section{Supplementary experiment details}

\subsection{Hyperparameters tables}

\begin{table}[H]
    \centering
    \small
    \caption{Hyperparameters for NIST datasets in Sec.~\ref{sec:experiment}.}
    \label{tab:hp-nist}
    \begin{tabular}{|c|c|c|}
        \hline
        \multicolumn{3}{|c|}{\textbf{EBM Prior}} \\
        \hline
        layer\_widths & $P,Q$ & 40, 81 \\
        kan\_type & (-) & RBF \\
        base\_activation & (-) & Relu \\
        layer\_norm & (-) & false \\
        grid\_size & $N_{\text{grid}}$ & 20 \\
        grid\_update\_ratio & $\alpha_{\text{grid}}$ & 0.05 \\
        \multirow{3}{*}{Initial grid\_range (before grid updating)} & \multirow{3}{*}{$[z_{\text{min}}, z_{\text{max}}]$} & Uniform: $[0,1]$ \\
        & & Gaussian: $[-1.5,1.5]$ \\
        & & No ref.: $[-1.2,1.2]$ \\
        KAN basis $\varepsilon$ init. & $\varepsilon$ & 0.1 \\
        KAN basis $\mu$ scaling & $\mu$ & 1.0 \\
        Basis activation scaling & $\sigma_b$ & 1.0 \\
        Basis KAN scaling & $\sigma_s$ & 1.0 \\
        Initial RBF variance & $\tau_0$ & 1.0 \\
        Train variance & $\tau_{\text{trainable}}$ & true \\
        GaussQuad\_nodes & $N_{\text{quad}}$ & 200 \\
        Use mixture prior & (-) & false \\
        \hline
        \multicolumn{3}{|c|}{\textbf{KAN Likelihood}} \\
        \hline
        hidden\_widths & $q,2q+1$ & 81, 162 \\
        All other KAN hparams & \multicolumn{2}{|c|}{Identical to EBM Prior} \\
        generator\_noise & $\sigma_{\epsilon}$ & 0.1 \\
        generator\_variance & $\sigma_{\text{llhood}}$ & 0.1 \\
        output\_activation & $\sigma_{\text{act}}$ & sigmoid \\
        Importance sampling resampler & resampler & residual \\
        ESS threshold & $\gamma$ & 0.5 \\
        \hline
        \multicolumn{3}{|c|}{\textbf{Grid updating}} \\
        \hline
        update\_prior\_grid & (-) & true \\
        update\_llhood\_grid & (-) & true \\
        num\_grid\_updating\_samples & (-) & 100 \\
        grid\_update\_frequency & (-) & 100 \\
        grid\_update\_decay & (-) & 0.999 \\
        \hline
        \multicolumn{3}{|c|}{\textbf{Training}} \\
        \hline
        batch\_size & $N_{x}$ & 100 \\
        importance\_sample\_size & $N_{s}$ & 100 \\
        Training examples & $N_{\text{train}}$ &50,000 \\
        Testing examples & $N_{\text{test}}$ & 100 \\
        num\_generated\_samples & (-) & 3,000 \\
        Number of epochs & $N_{\text{epochs}}$ & 10 \\
        eps & $\epsilon$ & 0.001 \\
        \hline
        \multicolumn{3}{|c|}{\textbf{Optimizer}} \\
        \hline
        type & (-) & adam \\
        learning\_rate & $\eta_{LR}$ & 0.001 \\
        betas & (-) & (0.9, 0.999) \\
        \hline
    \end{tabular}
\end{table}

\begin{table}[H]
    \centering
    \small
    \caption{Hyperparameters for SVHN, CIFAR10, and CelebA in Sec.~\ref{sec:experiment}.}
    \label{tab:hp-cnn}
    \begin{tabular}{|c|c|c|}
        \hline
        \multicolumn{3}{|c|}{\textbf{EBM Prior}} \\
        \hline
        layer\_widths & $P,Q$ & See Sec.~\ref{sec:arch} \\
        kan\_type & (-) & Morlet Wavelet \\
        base\_activation & (-) & None \\
        layer\_norm & (-) & false \\
        grid\_size & $N_{\text{grid}}$ & N/A \\
        KAN basis $\varepsilon$ init. & $\varepsilon$ & 0.1 \\
        KAN basis $\mu$ scaling & $\mu$ & 1.0 \\
        Basis activation scaling & $\sigma_b$ & 1.0 \\
        Basis KAN scaling & $\sigma_s$ & 1.0 \\
        Initial Wavelet bandwidth. & $\tau_0$ & 1 \\
        Train bandwidth & $\tau_{\text{trainable}}$ & true \\
        GaussQuad\_nodes & $N_{\text{quad}}$ & 200 \\
        Use mixture prior & (-) & true \\
        Initial prior domain for ITS & (-) & $[-12.0,12.0]$ \\
        \hline
        \multicolumn{3}{|c|}{\textbf{KAN Likelihood}} \\
        \hline
        cnn\_widths & (-) & See Sec.~\ref{sec:arch} \\
        All other KAN hparams & \multicolumn{2}{|c|}{N/A, a standard CNN was used} \\
        generator\_noise & $\sigma_{\epsilon}$ & 0 \\
        generator\_variance & $\sigma_{\epsilon}$ & 0.3 \\
        output\_activation & $\sigma_{\text{act}}$ & sigmoid \\
        Importance sampling resampler & resampler & Not used \\
        ESS threshold & $\gamma$ & N/A \\
        VGG16 perceptual loss scale, perceptual\_scale & (-) & N/A \\
        \hline 
        \multicolumn{3}{|c|}{\textbf{Grid updating}} \\
        \hline
        update\_prior\_grid & (-) & N/A \\
        update\_llhood\_grid & (-) & N/A \\
        \hline
        \multicolumn{3}{|c|}{\textbf{Mixture}} \\
        \hline
        use\_mixture\_prior & (-) & true \\
        train\_proportions & (-) & false \\
        \hline
        \multicolumn{3}{|c|}{\textbf{Training}} \\
        \hline
        batch\_size & $N_{x}$ & 50 \\
        importance\_sample\_size & $N_{s}$ & N/A \\
        Training examples & $N_{\text{train}}$ & 40,000 \\
        Testing examples & $N_{\text{test}}$ & 100 \\
        num\_generated\_samples & (-) & 20,000 \\
        Number of epochs & $N_{\text{epochs}}$ & 100 \\
        eps & $\epsilon$ & 0.0001 \\
        Use VGG16 perceptual loss, use\_perceptual\_loss & (-) & false \\
        \hline 
        \multicolumn{3}{|c|}{\textbf{Optimizer}} \\
        \hline
        type & (-) & adam \\
        gen\_learning\_rate & $\eta_{\text{gen}}$ & 0.0001 \\
        ebm\_learning\_rate & $\eta_{\text{ebm}}$ & 0.00002 \\
        betas & (-) & (0.9, 0.999) \\
        \hline
        \multicolumn{3}{|c|}{\textbf{POSTERIOR\_ULA}} \\
        \hline
        iters & $N_{\text{local}}$ & 40 \\
        ula\_eta & $\eta$ & 0.01 \\
        \hline
        \multicolumn{3}{|c|}{\textbf{THERMODYNAMIC\_INTEGRATION}} \\
        \hline
        iters & $N_{\text{local}}$ & (see posterior ULA) \\
        num\_temps & $N_{t}$ & 10 \\
        p\_start & $p_{\text{start}}$ & 2 \\
        p\_end & $p_{\text{end}}$ & 0.5 \\
        num\_cycles & $N_{\text{cycles}}$ & 0 \\
        exchange\_type & (-) & DEO \\
        \hline
    \end{tabular}
\end{table}

\subsection{SVHN, CIFAR10, \& CelebA Architectures}
\label{app:arch}

\subsubsection{KAEM Architectures}

\begin{table}[H]
    \centering
    \small
    \caption{The dense Morelet Wavelet KAN network used as the mixture prior model}
    \begin{tabular}{|c|c|c|}
        \hline
        \textbf{Layer} & \textbf{Operation} & \textbf{Output Dim.} \\
        \hline
        Input & Sample $\bm{z}$ & $P=40$ \\
        Energy functions & Dense($40$) & $Q=81$ \\
        \hline
    \end{tabular}
    \caption*{\small \texttt{sizes=[$n_z=40,2n_z+1=81$]}}
    \label{tab:mixture-dense}
\end{table}

\begin{table}[H]
    \centering
    \small
    \caption{CNN Generator Architecture for SVHN ($32\times32$)}
    \begin{tabular}{|c|c|c|c|}
        \hline
        \textbf{Layer} & \textbf{Operation} & \textbf{Output Shape} & \textbf{Stride} \\
        \hline
        Input & $\bm{z} \to$ Dense$(Q, 512)$, LeakyReLU $\to$ reshape & $2 \times 2 \times 128$ & -- \\
        1 & 4$\times$4 convT($128$), LeakyReLU & $4 \times 4 \times 128$ & 2 \\
        2 & 4$\times$4 convT($64$), LeakyReLU & $8 \times 8 \times 64$ & 2 \\
        3 & 4$\times$4 convT($32$), LeakyReLU & $16 \times 16 \times 32$ & 2 \\
        4 & 4$\times$4 convT(3), Sigmoid & $32 \times 32 \times 3$ & 2 \\
        \hline
    \end{tabular}
    \caption*{\small Config: \texttt{hidden\_feature\_dims=128,64,32, kernel=4,4,4,4, strides=2/2/2/2, paddings=1/1/1/1, activation=LeakyReLU, batchnorm=False, projection=True}}
    \label{tab:svhn-gen}
\end{table}

\begin{table}[H]
    \centering
    \small
    \caption{CNN Generator Architecture for CIFAR10 ($32\times32$)}
    \begin{tabular}{|c|c|c|c|}
        \hline
        \textbf{Layer} & \textbf{Operation} & \textbf{Output Shape} & \textbf{Stride} \\
        \hline
        Input & $\bm{z} \to$ Dense$(Q, 1024)$, LeakyReLU $\to$ reshape & $2 \times 2 \times 256$ & -- \\
        1 & 4$\times$4 convT($256$), LeakyReLU & $4 \times 4 \times 256$ & 2 \\
        2 & 4$\times$4 convT($128$), LeakyReLU & $8 \times 8 \times 128$ & 2 \\
        3 & 4$\times$4 convT($64$), LeakyReLU & $16 \times 16 \times 64$ & 2 \\
        4 & 4$\times$4 convT(3), Sigmoid & $32 \times 32 \times 3$ & 2 \\
        \hline
    \end{tabular}
    \caption*{\small Config: \texttt{hidden\_feature\_dims=256,128,64, kernel=4,4,4,4, strides=2/2/2/2, paddings=1/1/1/1, activation=LeakyReLU, batchnorm=False, projection=True}}
    \label{tab:cifar-gen}
\end{table}

\begin{table}[H]
    \centering
    \small
    \caption{CNN Generator Architecture for CelebA ($64\times64$)}
    \begin{tabular}{|c|c|c|c|}
        \hline
        \textbf{Layer} & \textbf{Operation} & \textbf{Output Shape} & \textbf{Stride} \\
        \hline
        Input & $\bm{z} \to$ Dense$(Q, 1024)$, LeakyReLU $\to$ reshape & $2 \times 2 \times 256$ & -- \\
        1 & 4$\times$4 convT($256$), LeakyReLU & $4 \times 4 \times 256$ & 2 \\
        2 & 4$\times$4 convT($128$), LeakyReLU & $8 \times 8 \times 128$ & 2 \\
        3 & 4$\times$4 convT($64$), LeakyReLU & $16 \times 16 \times 64$ & 2 \\
        4 & 4$\times$4 convT($32$), LeakyReLU & $32 \times 32 \times 32$ & 2 \\
        5 & 4$\times$4 convT(3), Sigmoid & $64 \times 64 \times 3$ & 2 \\
        \hline
    \end{tabular}
    \caption*{\small Config: \texttt{hidden\_feature\_dims=256,128,64,32, kernel=4,4,4,4,4, strides=2/2/2/2/2, paddings=1/1/1/1/1, activation=LeakyReLU, batchnorm=False, projection=True}}
    \label{tab:celeba-gen}
\end{table}

\subsubsection{Baseline VAE architectures}

\begin{table}[H]
    \centering
    \small
    \caption{VAE Encoder Architecture for SVHN ($32\times32$). Trained with KL weight, $\beta = 0.01$, Adam optimizer at learning rate $1\times10^{-4}$ and betas (0.9, 0.999).}
    \begin{tabular}{|c|c|c|c|}
        \hline
        \textbf{Layer} & \textbf{Operation} & \textbf{Output Shape} & \textbf{Stride} \\
        \hline
        Input & Image $\bm{x}$ & $32 \times 32 \times 3$ & -- \\
        1 & 4$\times$4 conv($16$), LeakyReLU & $16 \times 16 \times 16$ & 2 \\
        2 & 4$\times$4 conv($32$), LeakyReLU & $8 \times 8 \times 32$ & 2 \\
        3 & 4$\times$4 conv($64$), LeakyReLU & $4 \times 4 \times 64$ & 2 \\
        4 & 4$\times$4 conv($128$), LeakyReLU & $2 \times 2 \times 128$ & 2 \\
        5 & Flatten + Dense & $\bm{\mu}, \log\bm{\sigma}^2 \in \mathbb{R}^{81}$ & -- \\
        \hline
    \end{tabular}
    \caption*{\small Config: \texttt{channels=16,32,64,128, kernels=4,4,4,4, strides=2/2/2/2, paddings=1/1/1/1, latent\_dim=81, batchnorm=False}}
    \label{tab:svhn-vae-enc}
\end{table}

\begin{table}[H]
    \centering
    \small
    \caption{VAE Decoder Architecture for SVHN ($32\times32$)}
    \begin{tabular}{|c|c|c|c|}
        \hline
        \textbf{Layer} & \textbf{Operation} & \textbf{Output Shape} & \textbf{Stride} \\
        \hline
        Input & Latent vector $\bm{z}$ & $81$ & -- \\
        1 & Dense$(81, 512)$, LeakyReLU + Reshape & $2 \times 2 \times 128$ & -- \\
        2 & 4$\times$4 convT($128$), LeakyReLU & $4 \times 4 \times 128$ & 2 \\
        3 & 4$\times$4 convT($64$), LeakyReLU & $8 \times 8 \times 64$ & 2 \\
        4 & 4$\times$4 convT($32$), LeakyReLU & $16 \times 16 \times 32$ & 2 \\
        5 & 4$\times$4 convT($3$), Sigmoid & $32 \times 32 \times 3$ & 2 \\
        \hline
    \end{tabular}
    \caption*{\small Config: \texttt{channels=128,64,32,16, kernels=4,4,4,4, strides=2/2/2/2, paddings=1/1/1/1, latent\_dim=81, batchnorm=False}}
    \label{tab:svhn-vae-dec}
\end{table}

\begin{table}[H]
    \centering
    \small
    \caption{VAE Encoder Architecture for CIFAR10 ($32\times32$)}
    \begin{tabular}{|c|c|c|c|}
        \hline
        \textbf{Layer} & \textbf{Operation} & \textbf{Output Shape} & \textbf{Stride} \\
        \hline
        Input & Image $\bm{x}$ & $32 \times 32 \times 3$ & -- \\
        1 & 4$\times$4 conv($32$), LeakyReLU & $16 \times 16 \times 32$ & 2 \\
        2 & 4$\times$4 conv($64$), LeakyReLU & $8 \times 8 \times 64$ & 2 \\
        3 & 4$\times$4 conv($128$), LeakyReLU & $4 \times 4 \times 128$ & 2 \\
        4 & 4$\times$4 conv($256$), LeakyReLU & $2 \times 2 \times 256$ & 2 \\
        5 & Flatten + Dense & $\bm{\mu}, \log\bm{\sigma}^2 \in \mathbb{R}^{81}$ & -- \\
        \hline
    \end{tabular}
    \caption*{\small Config: \texttt{channels=32,64,128,256, kernels=4,4,4,4, strides=2/2/2/2, paddings=1/1/1/1, latent\_dim=81, batchnorm=False}}
    \label{tab:cifar-vae-enc}
\end{table}

\begin{table}[H]
    \centering
    \small
    \caption{VAE Decoder Architecture for CIFAR10 ($32\times32$)}
    \begin{tabular}{|c|c|c|c|}
        \hline
        \textbf{Layer} & \textbf{Operation} & \textbf{Output Shape} & \textbf{Stride} \\
        \hline
        Input & Latent vector $\bm{z}$ & $81$ & -- \\
        1 & Dense$(81, 1024)$, LeakyReLU + Reshape & $2 \times 2 \times 256$ & -- \\
        2 & 4$\times$4 convT($256$), LeakyReLU & $4 \times 4 \times 256$ & 2 \\
        3 & 4$\times$4 convT($128$), LeakyReLU & $8 \times 8 \times 128$ & 2 \\
        4 & 4$\times$4 convT($64$), LeakyReLU & $16 \times 16 \times 64$ & 2 \\
        5 & 4$\times$4 convT($3$), Sigmoid & $32 \times 32 \times 3$ & 2 \\
        \hline
    \end{tabular}
    \caption*{\small Config: \texttt{channels=256,128,64,32, kernels=4,4,4,4, strides=2/2/2/2, paddings=1/1/1/1, latent\_dim=81, batchnorm=False}}
    \label{tab:cifar-vae-dec}
\end{table}

\begin{table}[H]
    \centering
    \small
    \caption{VAE Encoder Architecture for CelebA ($64\times64$). Trained with KL weight, $\beta = 0.01$, Adam optimizer at learning rate $1\times10^{-4}$.}
    \begin{tabular}{|c|c|c|c|}
        \hline
        \textbf{Layer} & \textbf{Operation} & \textbf{Output Shape} & \textbf{Stride} \\
        \hline
        Input & Image $\bm{x}$ & $64 \times 64 \times 3$ & -- \\
        1 & 4$\times$4 conv($16$), LeakyReLU & $32 \times 32 \times 16$ & 2 \\
        2 & 4$\times$4 conv($32$), LeakyReLU & $16 \times 16 \times 32$ & 2 \\
        3 & 4$\times$4 conv($64$), LeakyReLU & $8 \times 8 \times 64$ & 2 \\
        4 & 4$\times$4 conv($128$), LeakyReLU & $4 \times 4 \times 128$ & 2 \\
        5 & 4$\times$4 conv($256$), LeakyReLU & $2 \times 2 \times 256$ & 2 \\
        6 & Flatten + Dense & $\bm{\mu}, \log\bm{\sigma}^2 \in \mathbb{R}^{81}$ & -- \\
        \hline
    \end{tabular}
    \caption*{\small Config: \texttt{channels=16,32,64,128,256, kernels=4,4,4,4,4, strides=2/2/2/2/2, paddings=1/1/1/1/1, latent\_dim=81, batchnorm=False}}
    \label{tab:celeba-vae-enc}
\end{table}

\begin{table}[H]
    \centering
    \small
    \caption{VAE Decoder Architecture for CelebA ($64\times64$)}
    \begin{tabular}{|c|c|c|c|}
        \hline
        \textbf{Layer} & \textbf{Operation} & \textbf{Output Shape} & \textbf{Stride} \\
        \hline
        Input & Latent vector $\bm{z}$ & $81$ & -- \\
        1 & Dense$(81, 1024)$, LeakyReLU + Reshape & $2 \times 2 \times 256$ & -- \\
        2 & 4$\times$4 convT($256$), LeakyReLU & $4 \times 4 \times 256$ & 2 \\
        3 & 4$\times$4 convT($128$), LeakyReLU & $8 \times 8 \times 128$ & 2 \\
        4 & 4$\times$4 convT($64$), LeakyReLU & $16 \times 16 \times 64$ & 2 \\
        5 & 4$\times$4 convT($32$), LeakyReLU & $32 \times 32 \times 32$ & 2 \\
        6 & 4$\times$4 convT($3$), Sigmoid & $64 \times 64 \times 3$ & 2 \\
        \hline
    \end{tabular}
    \caption*{\small Config: \texttt{channels=256,128,64,32,16, kernels=4,4,4,4,4, strides=2/2/2/2/2, paddings=1/1/1/1/1, latent\_dim=81, batchnorm=False}}
    \label{tab:celeba-vae-dec}
\end{table}

\subsubsection{Baseline neural latent EBM architectures}

\begin{table}[H]
    \centering
    \small
    \caption{Neural latent EBM \citep{pang2020learninglatentspaceenergybased} generator architecture for SVHN ($32\times32$). Posterior sampling: 40 ULA steps with step size $0.01$ and Gaussian observation $\sigma_x = 0.3$. Prior sampling: 60 ULA steps with step size $0.16$. Noise scale $1.0$. Prior $\mathcal{N}(0,1)$. Optimizers: Adam, $1\times10^{-4}$ for the generator and $2\times10^{-5}$ for the energy network, matching \citet{pang2020learninglatentspaceenergybased}.}
    \begin{tabular}{|c|c|c|c|}
        \hline
        \textbf{Layer} & \textbf{Operation} & \textbf{Output Shape} & \textbf{Stride} \\
        \hline
        Input & Latent vector $\bm{z}$ & $81$ & -- \\
        1 & Dense$(81, 512)$ + LeakyReLU + Reshape & $2 \times 2 \times 128$ & -- \\
        2 & 4$\times$4 convT($128$), LeakyReLU & $4 \times 4 \times 128$ & 2 \\
        3 & 4$\times$4 convT($64$), LeakyReLU & $8 \times 8 \times 64$ & 2 \\
        4 & 4$\times$4 convT($32$), LeakyReLU & $16 \times 16 \times 32$ & 2 \\
        5 & 4$\times$4 convT($3$), Sigmoid & $32 \times 32 \times 3$ & 2 \\
        \hline
    \end{tabular}
    \caption*{\small Config: \texttt{generator\_channels=128,64,32,16, kernels=4,4,4,4, strides=2/2/2/2, paddings=1/1/1/1, latent\_dim=81}}
    \label{tab:svhn-nlebm-gen}
\end{table}

\begin{table}[H]
    \centering
    \small
    \caption{Neural latent EBM \citep{pang2020learninglatentspaceenergybased} energy network for SVHN ($32\times32$).}
    \begin{tabular}{|c|c|c|}
        \hline
        \textbf{Layer} & \textbf{Operation} & \textbf{Output Shape} \\
        \hline
        Input & Latent vector $\bm{z}$ & $81$ \\
        1 & Dense$(81, 200)$, Swish & $200$ \\
        2 & Dense$(200, 200)$, Swish & $200$ \\
        3 & Dense$(200, 1)$ & $1$ \\
        \hline
    \end{tabular}
    \caption*{\small Config: \texttt{energy\_widths=200,200,1}}
    \label{tab:svhn-nlebm-ebm}
\end{table}

\begin{table}[H]
    \centering
    \small
    \caption{Neural latent EBM \citep{pang2020learninglatentspaceenergybased} generator architecture for CIFAR10 ($32\times32$). Posterior sampling: 40 ULA steps with step size $0.01$ and Gaussian observation $\sigma_x = 0.3$. Prior sampling: 60 ULA steps with step size $0.16$. Reference prior $\mathcal{N}(0,1)$. Optimizers: Adam, $1\times10^{-4}$ for the generator and $2\times10^{-5}$ for the energy network, matching \citet{pang2020learninglatentspaceenergybased}.}
    \begin{tabular}{|c|c|c|c|}
        \hline
        \textbf{Layer} & \textbf{Operation} & \textbf{Output Shape} & \textbf{Stride} \\
        \hline
        Input & Latent vector $\bm{z}$ & $81$ & -- \\
        1 & Dense$(81, 1024)$ + LeakyReLU + Reshape & $2 \times 2 \times 256$ & -- \\
        2 & 4$\times$4 convT($256$), LeakyReLU & $4 \times 4 \times 256$ & 2 \\
        3 & 4$\times$4 convT($128$), LeakyReLU & $8 \times 8 \times 128$ & 2 \\
        4 & 4$\times$4 convT($64$), LeakyReLU & $16 \times 16 \times 64$ & 2 \\
        5 & 4$\times$4 convT($3$), Sigmoid & $32 \times 32 \times 3$ & 2 \\
        \hline
    \end{tabular}
    \caption*{\small Config: \texttt{generator\_channels=256,128,64,32, kernels=4,4,4,4, strides=2/2/2/2, paddings=1/1/1/1, latent\_dim=81}}
    \label{tab:cifar-nlebm-gen}
\end{table}

\begin{table}[H]
    \centering
    \small
    \caption{Neural latent EBM \citep{pang2020learninglatentspaceenergybased} energy network for CIFAR10 ($32\times32$).}
    \begin{tabular}{|c|c|c|}
        \hline
        \textbf{Layer} & \textbf{Operation} & \textbf{Output Shape} \\
        \hline
        Input & Latent vector $\bm{z}$ & $81$ \\
        1 & Dense$(81, 200)$, Swish & $200$ \\
        2 & Dense$(200, 200)$, Swish & $200$ \\
        3 & Dense$(200, 1)$ & $1$ \\
        \hline
    \end{tabular}
    \caption*{\small Config: \texttt{energy\_widths=200,200,1}}
    \label{tab:cifar-nlebm-ebm}
\end{table}

\begin{table}[H]
    \centering
    \small
    \caption{Neural latent EBM \citep{pang2020learninglatentspaceenergybased} generator architecture for CelebA ($64\times64$). Posterior sampling: 40 ULA steps with step size $0.01$ and Gaussian observation $\sigma_x = 0.3$. Prior sampling: 60 ULA steps with step size $0.16$. Reference prior $\mathcal{N}(0,1)$. Optimizers: Adam, $1\times10^{-4}$ for the generator and $2\times10^{-5}$ for the energy network, matching \citet{pang2020learninglatentspaceenergybased}.}
    \begin{tabular}{|c|c|c|c|}
        \hline
        \textbf{Layer} & \textbf{Operation} & \textbf{Output Shape} & \textbf{Stride} \\
        \hline
        Input & Latent vector $\bm{z}$ & $81$ & -- \\
        1 & Dense$(81, 1024)$ + LeakyReLU + Reshape & $2 \times 2 \times 256$ & -- \\
        2 & 4$\times$4 convT($256$), LeakyReLU & $4 \times 4 \times 256$ & 2 \\
        3 & 4$\times$4 convT($128$), LeakyReLU & $8 \times 8 \times 128$ & 2 \\
        4 & 4$\times$4 convT($64$), LeakyReLU & $16 \times 16 \times 64$ & 2 \\
        5 & 4$\times$4 convT($32$), LeakyReLU & $32 \times 32 \times 32$ & 2 \\
        6 & 4$\times$4 convT($3$), Sigmoid & $64 \times 64 \times 3$ & 2 \\
        \hline
    \end{tabular}
    \caption*{\small Config: \texttt{generator\_channels=256,128,64,32,16, kernels=4,4,4,4,4, strides=2/2/2/2/2, paddings=1/1/1/1/1, latent\_dim=81}}
    \label{tab:celeba-nlebm-gen}
\end{table}

\begin{table}[H]
    \centering
    \small
    \caption{Neural latent EBM \citep{pang2020learninglatentspaceenergybased} energy network for CelebA ($64\times64$).}
    \begin{tabular}{|c|c|c|}
        \hline
        \textbf{Layer} & \textbf{Operation} & \textbf{Output Shape} \\
        \hline
        Input & Latent vector $\bm{z}$ & $81$ \\
        1 & Dense$(81, 200)$, Swish & $200$ \\
        2 & Dense$(200, 200)$, Swish & $200$ \\
        3 & Dense$(200, 1)$ & $1$ \\
        \hline
    \end{tabular}
    \caption*{\small Config: \texttt{energy\_widths=200,200,1}}
    \label{tab:celeba-nlebm-ebm}
\end{table}

\subsubsection{Baseline DDPM architectures}

The DDPM baseline \citep{ho2020denoisingdiffusionprobabilisticmodels} predicts the noise component at a sampled diffusion timestep with a UNet conditioned on a sinusoidal time embedding. Training uses $T = 1000$ forward steps with a linear $\beta$ schedule from $\beta_{\text{start}} = 10^{-4}$ to $\beta_{\text{end}} = 2 \times 10^{-2}$. At inference, we use the full unstrided schedule of $T_s = T = 1000$ ancestral denoising steps. At each step we recover $\hat{\bm{x}}_0$ from the predicted noise, clip $\hat{\bm{x}}_0$ to $[-1, 1]$, and form the posterior mean of $q(\bm{x}_{t-1} \mid \bm{x}_t, \hat{\bm{x}}_0)$ with posterior variance $\tilde{\beta}_t = \beta_t(1 - \bar{\alpha}_{t-1})/(1 - \bar{\alpha}_t)$, matching the reference DDPM implementation. All blocks use GELU activations, GroupNorm with at most $8$ groups, and an additive time-embedding projection between conv layers.

\begin{table}[H]
    \centering
    \small
    \caption{DDPM UNet architecture for SVHN and CIFAR10 ($32\times32$). Channels $[32, 64, 128, 256]$, $3{\times}3$ kernels, time embedding dim $128$ (projected to $512$) for SVHN and $256$ (projected to $1024$) for CIFAR10. Optimizer: Adam, learning rate $1\times10^{-4}$.}
    \begin{tabular}{|c|c|c|}
        \hline
        \textbf{Stage} & \textbf{Operation} & \textbf{Output Shape} \\
        \hline
        Input & Noisy image $\bm{x}_t$, time $t$ & $32 \times 32 \times 3$ \\
        Time embed & Sinusoidal $\to$ Dense, GELU $\to$ Dense & $4 \cdot \text{time\_dim}$ \\
        Init conv & $3{\times}3$ Conv($3 \to 32$), GELU & $32 \times 32 \times 32$ \\
        Down 1 & DownBlock($32 \to 64$), stride-$2$ Conv & $16 \times 16 \times 64$ \\
        Down 2 & DownBlock($64 \to 128$), stride-$2$ Conv & $8 \times 8 \times 128$ \\
        Down 3 & DownBlock($128 \to 256$), stride-$2$ Conv & $4 \times 4 \times 256$ \\
        Middle & $3{\times}3$ Conv($256$), GroupNorm, $3{\times}3$ Conv($256$) & $4 \times 4 \times 256$ \\
        Up 1 & ConvT($256$, s$=2$), concat skip, $3{\times}3$ Conv($512 \to 128$) & $8 \times 8 \times 128$ \\
        Up 2 & ConvT($128$, s$=2$), concat skip, $3{\times}3$ Conv($256 \to 64$) & $16 \times 16 \times 64$ \\
        Up 3 & ConvT($64$, s$=2$), concat skip, $3{\times}3$ Conv($128 \to 32$) & $32 \times 32 \times 32$ \\
        Final conv & $1{\times}1$ Conv($32 \to 3$) & $32 \times 32 \times 3$ \\
        \hline
    \end{tabular}
    \caption*{\small Each DownBlock and UpBlock applies two $3{\times}3$ Conv layers with GroupNorm and a Dense time-embedding projection added in between.}
    \label{tab:svhn-ddpm}
\end{table}

\begin{table}[H]
    \centering
    \small
    \caption{DDPM UNet architecture for CelebA ($64\times64$). Channels $[16, 32, 64, 128, 256]$, $3{\times}3$ kernels, time embedding dim $512$ (projected to $2048$). Optimizer: Adam, learning rate $1\times10^{-4}$.}
    \begin{tabular}{|c|c|c|}
        \hline
        \textbf{Stage} & \textbf{Operation} & \textbf{Output Shape} \\
        \hline
        Input & Noisy image $\bm{x}_t$, time $t$ & $64 \times 64 \times 3$ \\
        Time embed & Sinusoidal $\to$ Dense$(512, 2048)$, GELU $\to$ Dense$(2048, 2048)$ & $2048$ \\
        Init conv & $3{\times}3$ Conv($3 \to 16$), GELU & $64 \times 64 \times 16$ \\
        Down 1 & DownBlock($16 \to 32$), stride-$2$ Conv & $32 \times 32 \times 32$ \\
        Down 2 & DownBlock($32 \to 64$), stride-$2$ Conv & $16 \times 16 \times 64$ \\
        Down 3 & DownBlock($64 \to 128$), stride-$2$ Conv & $8 \times 8 \times 128$ \\
        Down 4 & DownBlock($128 \to 256$), stride-$2$ Conv & $4 \times 4 \times 256$ \\
        Middle & $3{\times}3$ Conv($256$), GroupNorm, $3{\times}3$ Conv($256$) & $4 \times 4 \times 256$ \\
        Up 1 & ConvT($256$, s$=2$), concat skip, $3{\times}3$ Conv($512 \to 128$) & $8 \times 8 \times 128$ \\
        Up 2 & ConvT($128$, s$=2$), concat skip, $3{\times}3$ Conv($256 \to 64$) & $16 \times 16 \times 64$ \\
        Up 3 & ConvT($64$, s$=2$), concat skip, $3{\times}3$ Conv($128 \to 32$) & $32 \times 32 \times 32$ \\
        Up 4 & ConvT($32$, s$=2$), concat skip, $3{\times}3$ Conv($64 \to 16$) & $64 \times 64 \times 16$ \\
        Final conv & $1{\times}1$ Conv($16 \to 3$) & $64 \times 64 \times 3$ \\
        \hline
    \end{tabular}
    \caption*{\small Each DownBlock and UpBlock applies two $3{\times}3$ Conv layers with GroupNorm and a Dense time-embedding projection added in between.}
    \label{tab:celeba-ddpm}
\end{table}

\section{Generated images}
\label{app:images}

\subsection{MNIST \& FMNIST}
\label{app:images-nist}

\begin{figure}[H]
    \centering

    \begin{subfigure}{0.32\textwidth}
        \centering
        \includegraphics[width=\linewidth]{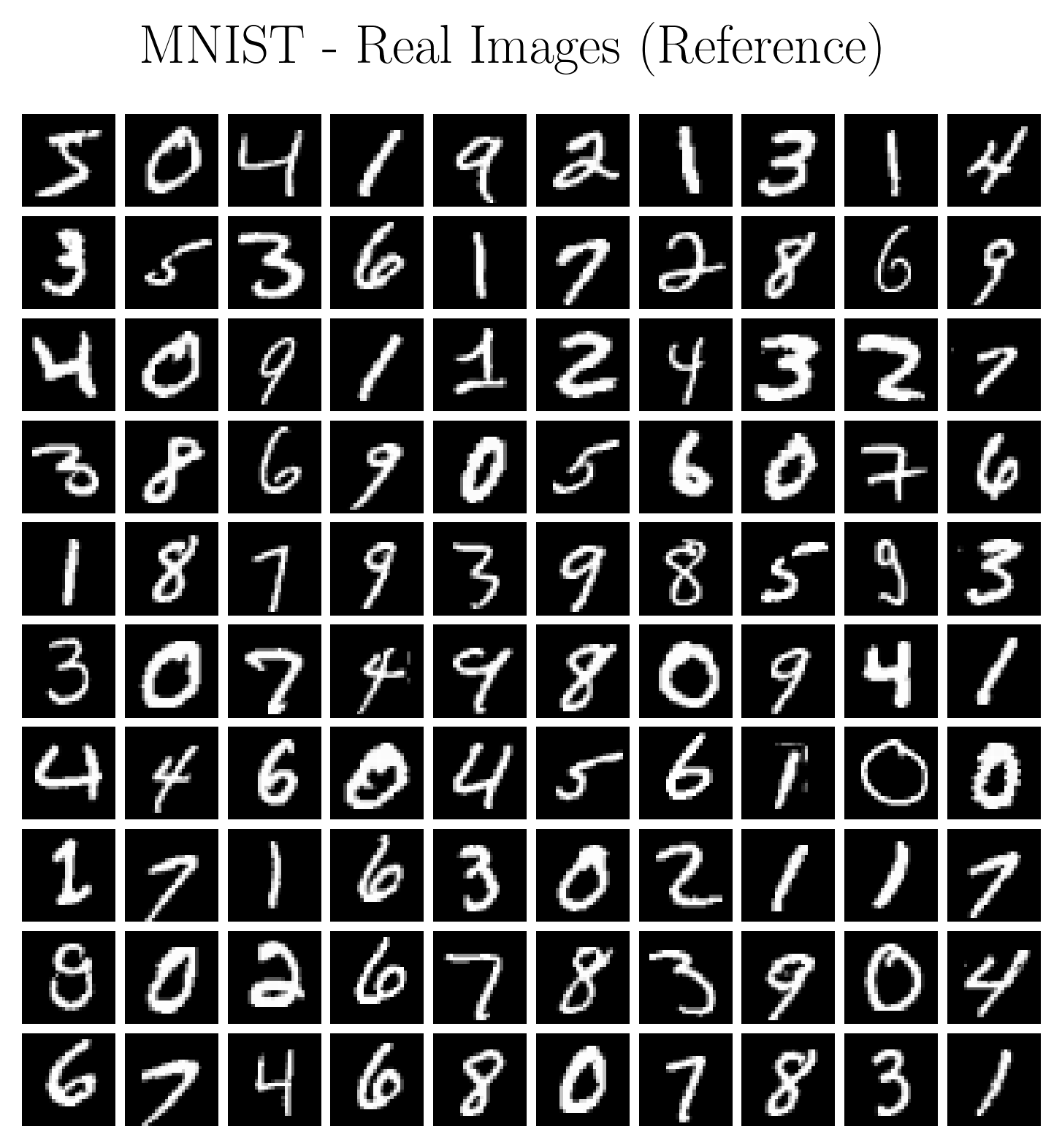}
        \caption{Real samples}
    \end{subfigure}
    \hfill
    \begin{subfigure}{0.32\textwidth}
        \centering
        \includegraphics[width=\linewidth]{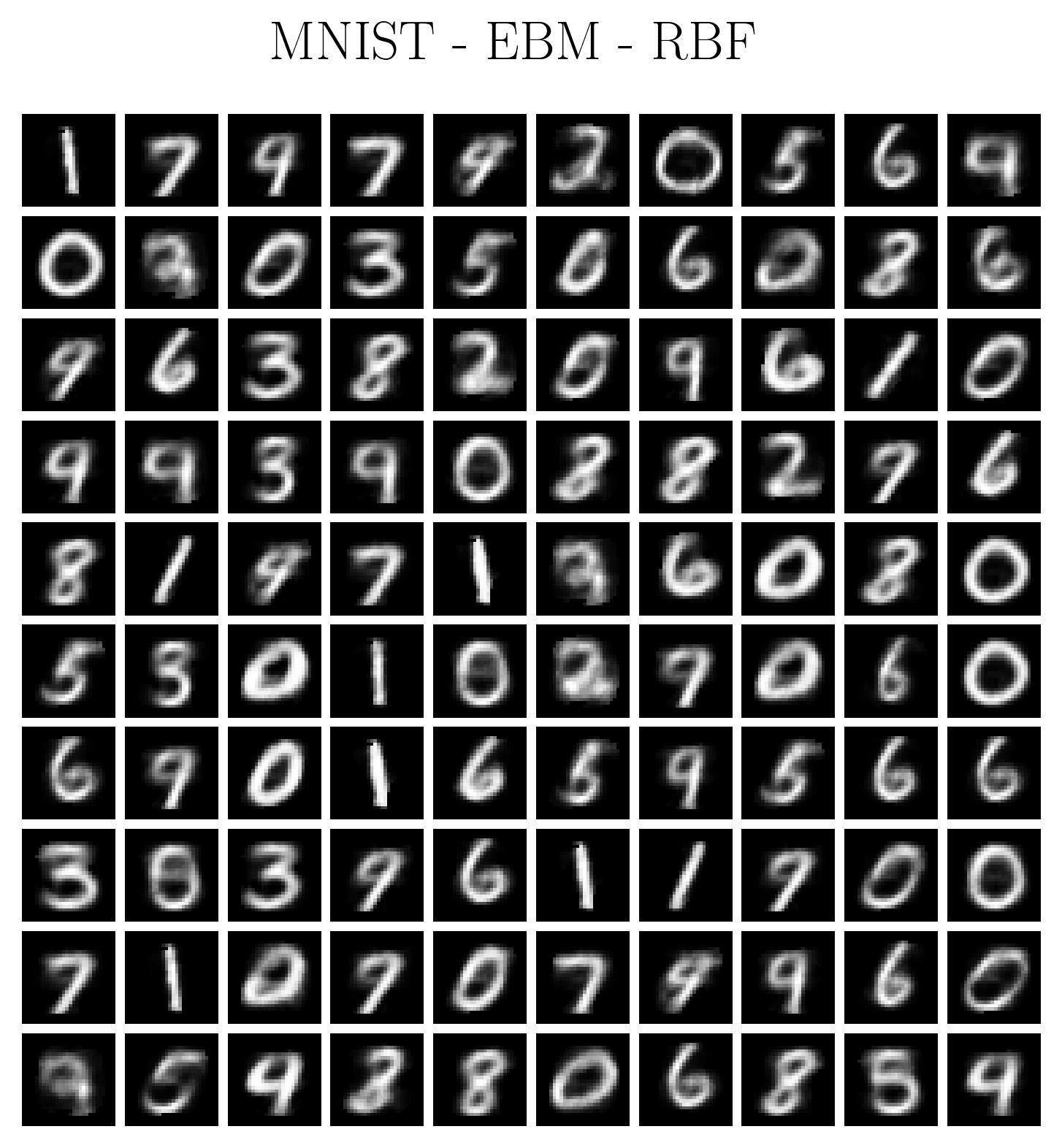}
        \caption{No ref. prior}
    \end{subfigure}

    \vspace{0.5cm}

    \begin{subfigure}{0.32\textwidth}
        \centering
        \includegraphics[width=\linewidth]{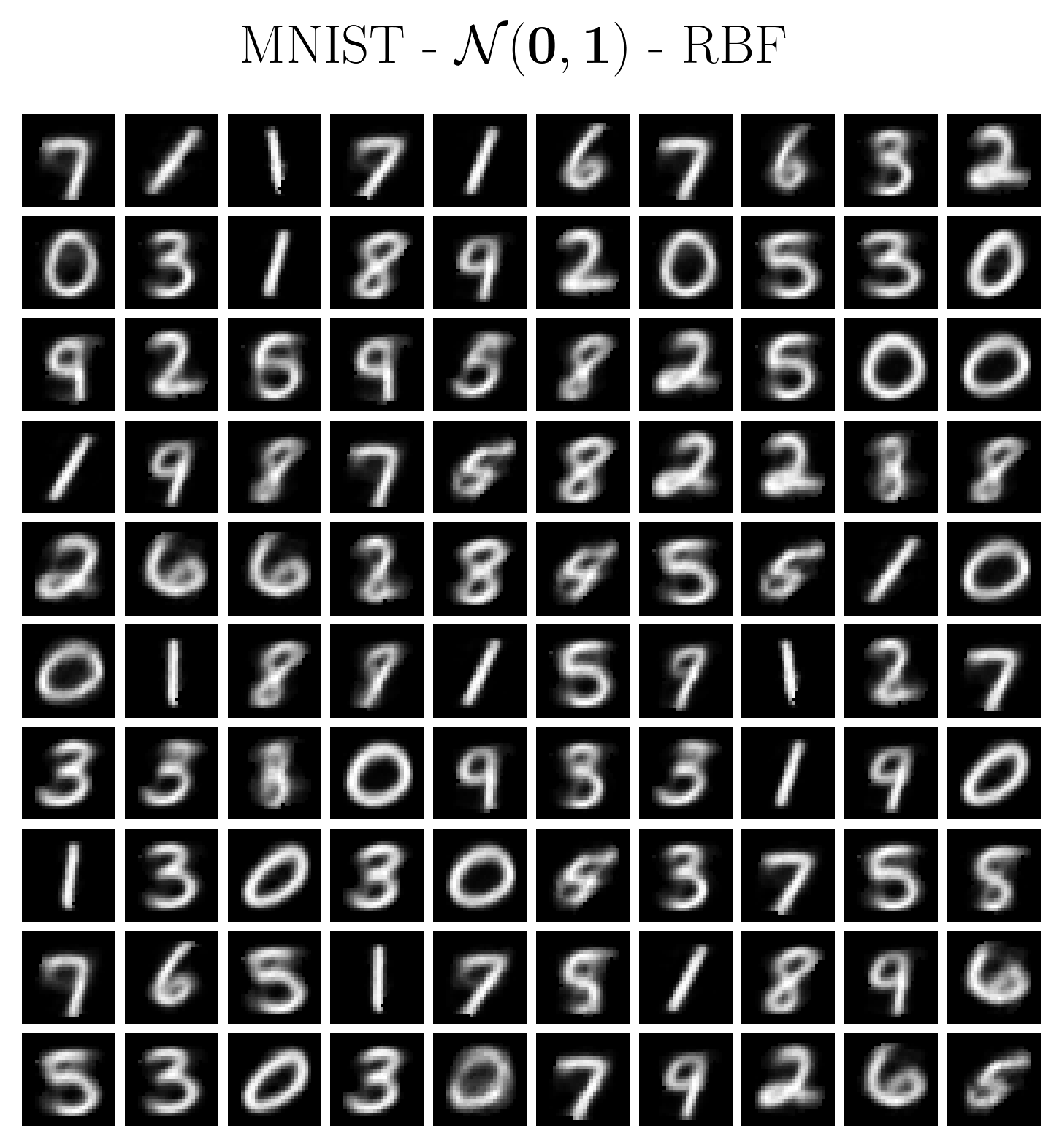}
        \caption{Gaussian ref. prior}
    \end{subfigure}
    \hfill
    \begin{subfigure}{0.32\textwidth}
        \centering
        \includegraphics[width=\linewidth]{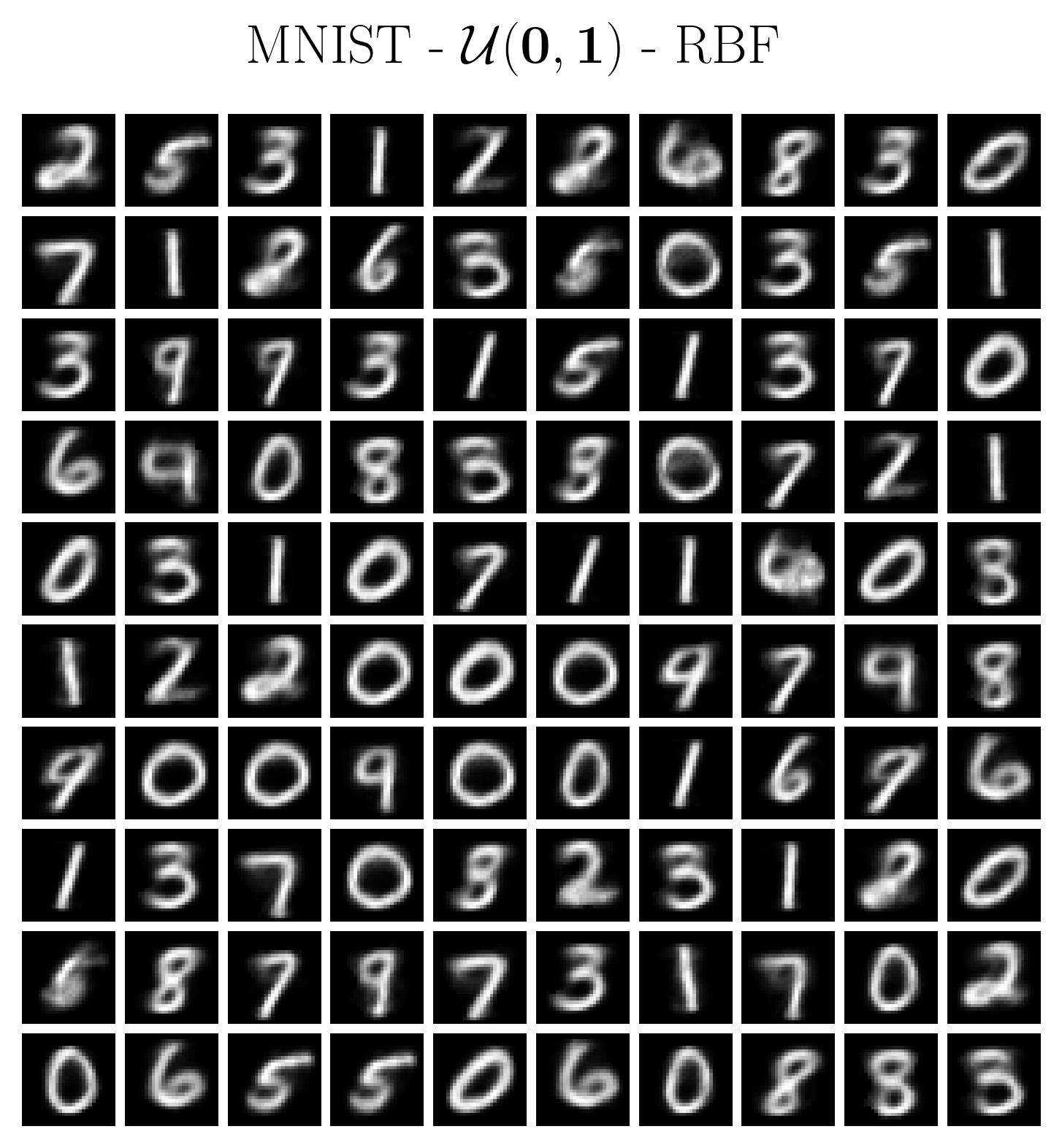}
        \caption{Uniform ref. prior}
    \end{subfigure}

    \caption{Generated MNIST samples}
    \label{fig:mnist}
\end{figure}

\begin{figure}[H]
    \centering

    \begin{subfigure}{0.32\textwidth}
        \centering
        \includegraphics[width=\linewidth]{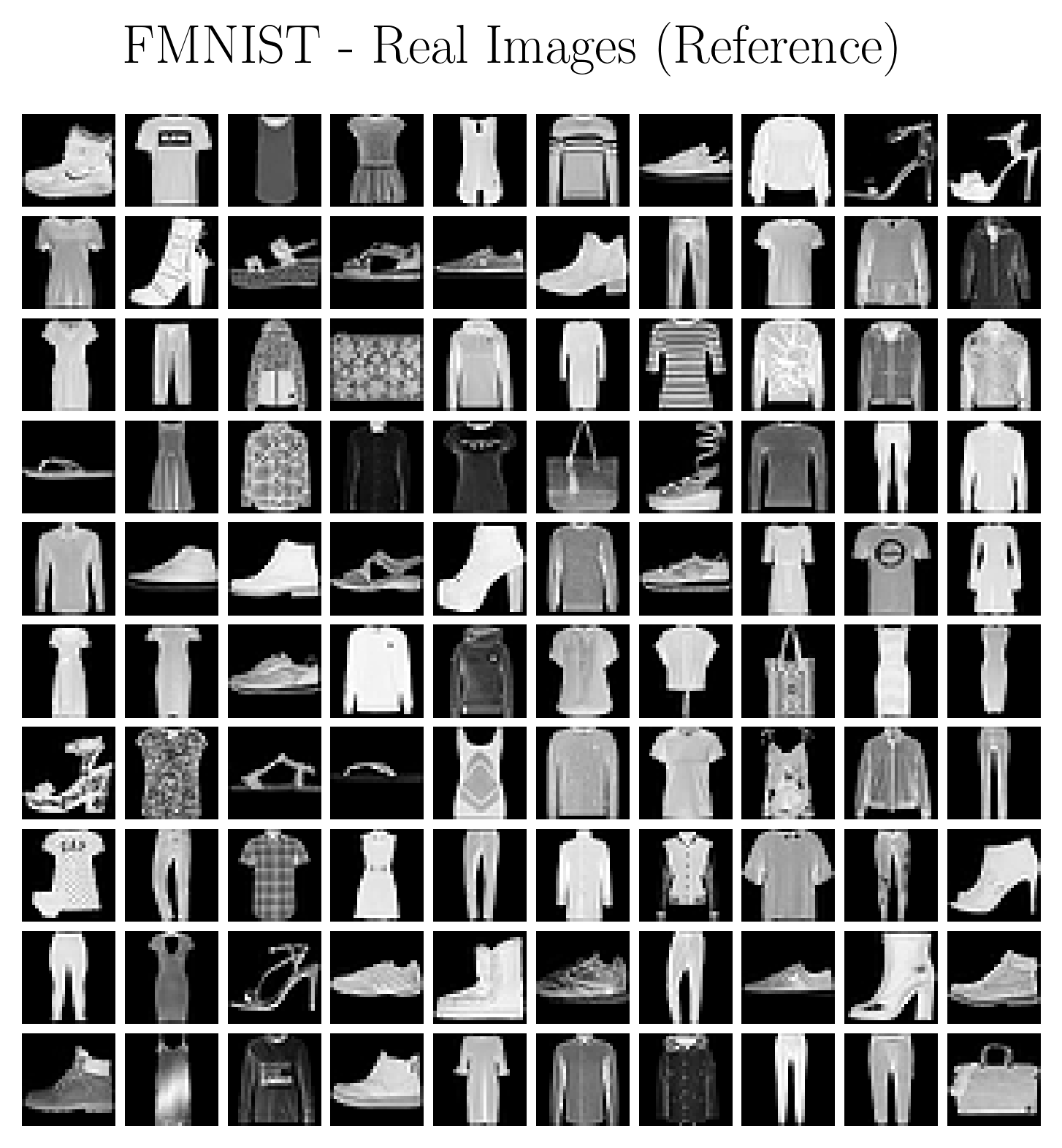}
        \caption{Real samples}
    \end{subfigure}
    \hfill
    \begin{subfigure}{0.32\textwidth}
        \centering
        \includegraphics[width=\linewidth]{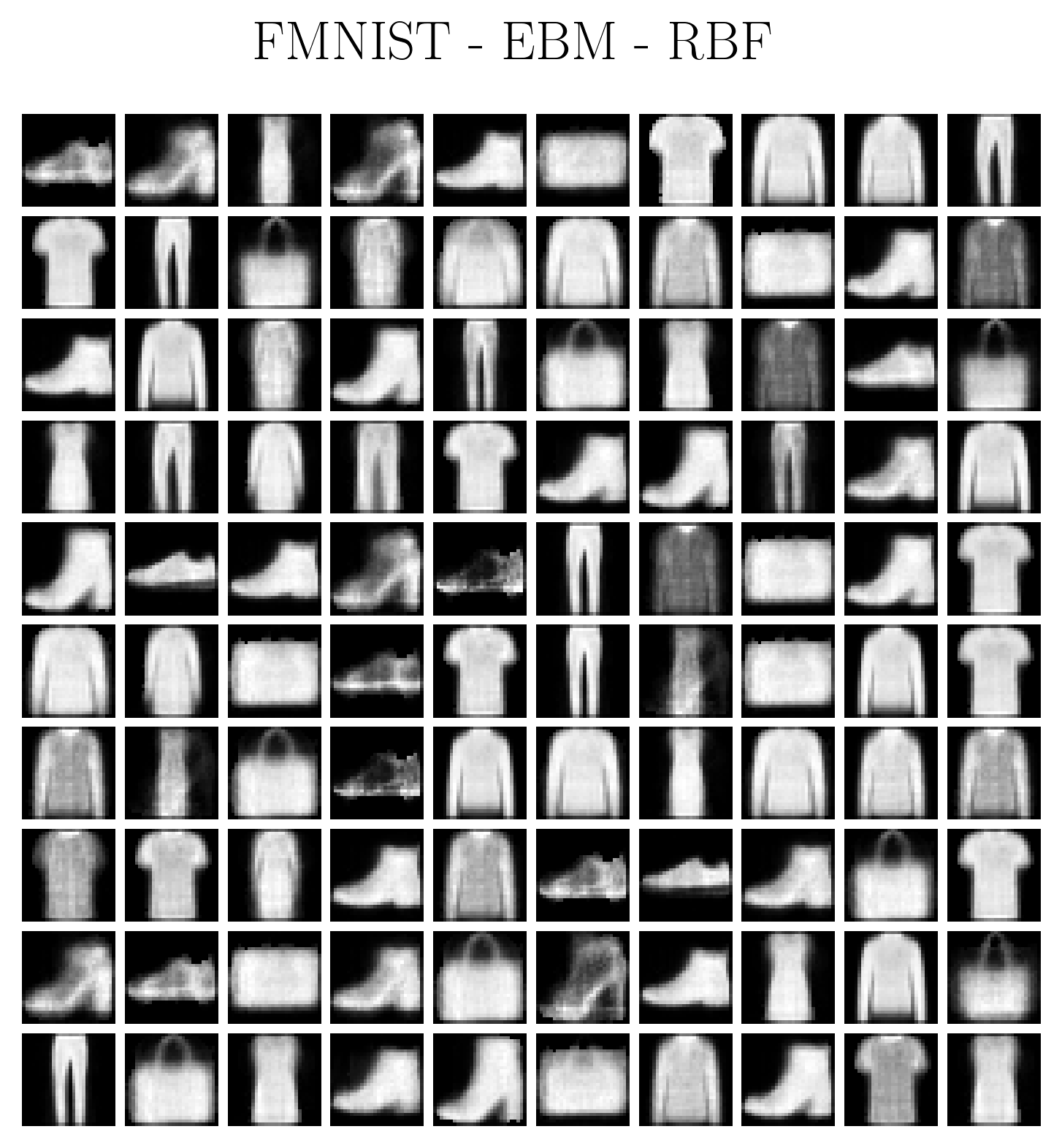}
        \caption{No ref. prior}
    \end{subfigure}

    \vspace{0.5cm}

    \begin{subfigure}{0.32\textwidth}
        \centering
        \includegraphics[width=\linewidth]{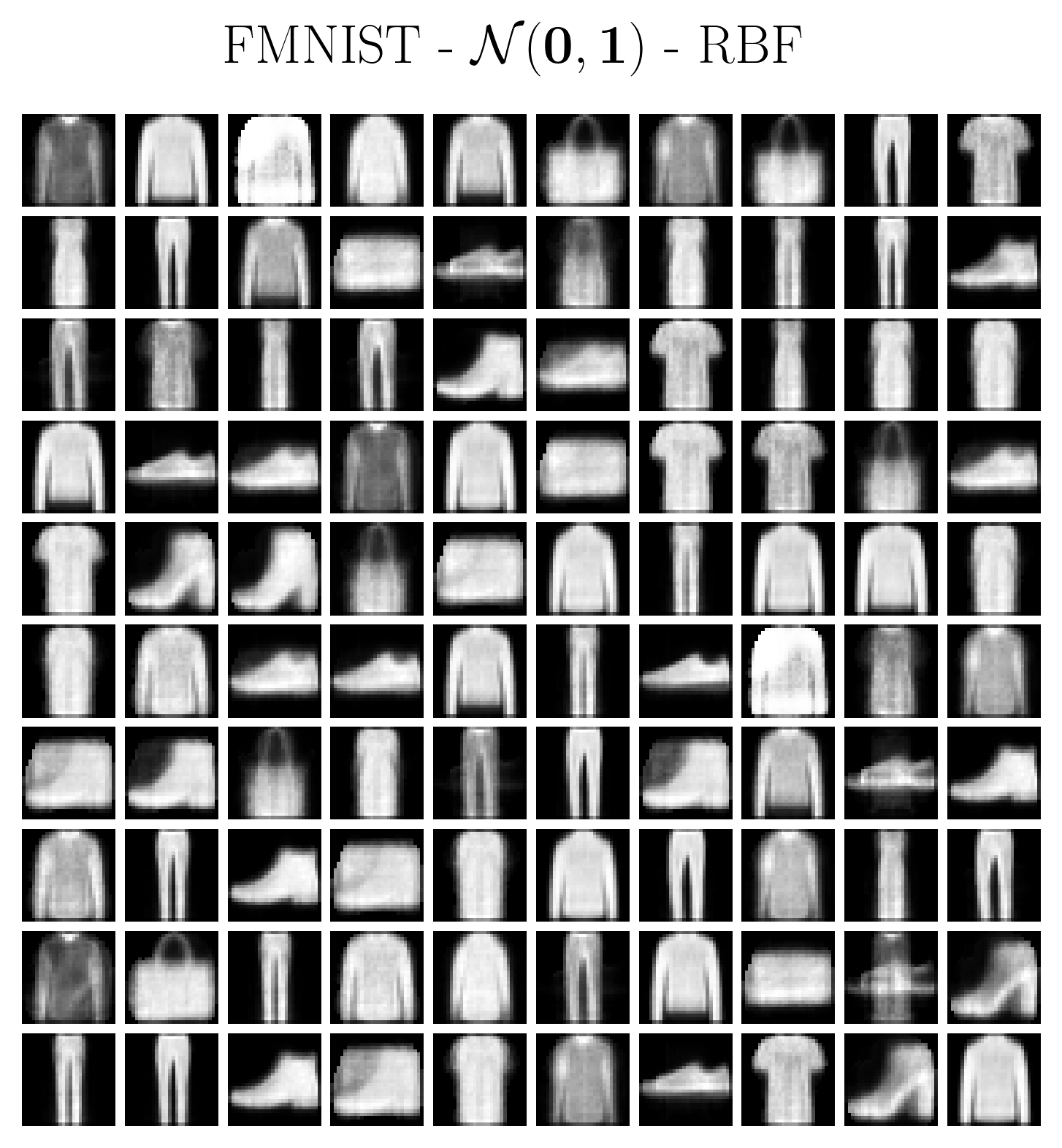}
        \caption{Gaussian ref. prior}
    \end{subfigure}
    \hfill
    \begin{subfigure}{0.32\textwidth}
        \centering
        \includegraphics[width=\linewidth]{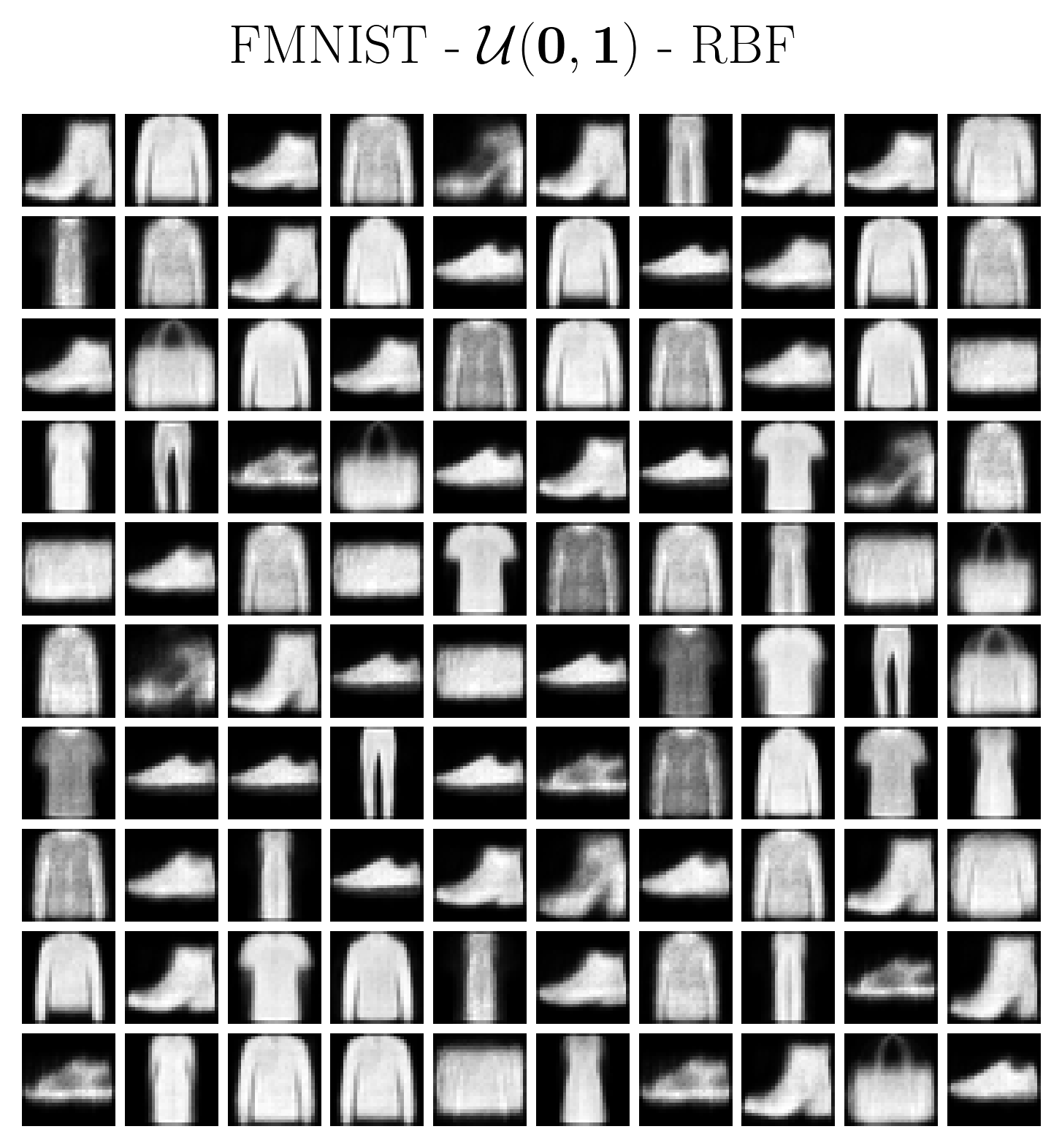}
        \caption{Uniform ref. prior}
    \end{subfigure}

    \caption{Generated FMNIST samples}
    \label{fig:fmnist}
\end{figure}

\subsection{SVHN, CIFAR10, \& CelebA}
\label{app:images-cif}

\begin{figure}[H]
    \centering

    \begin{subfigure}{0.32\textwidth}
        \centering
        \includegraphics[width=\linewidth]{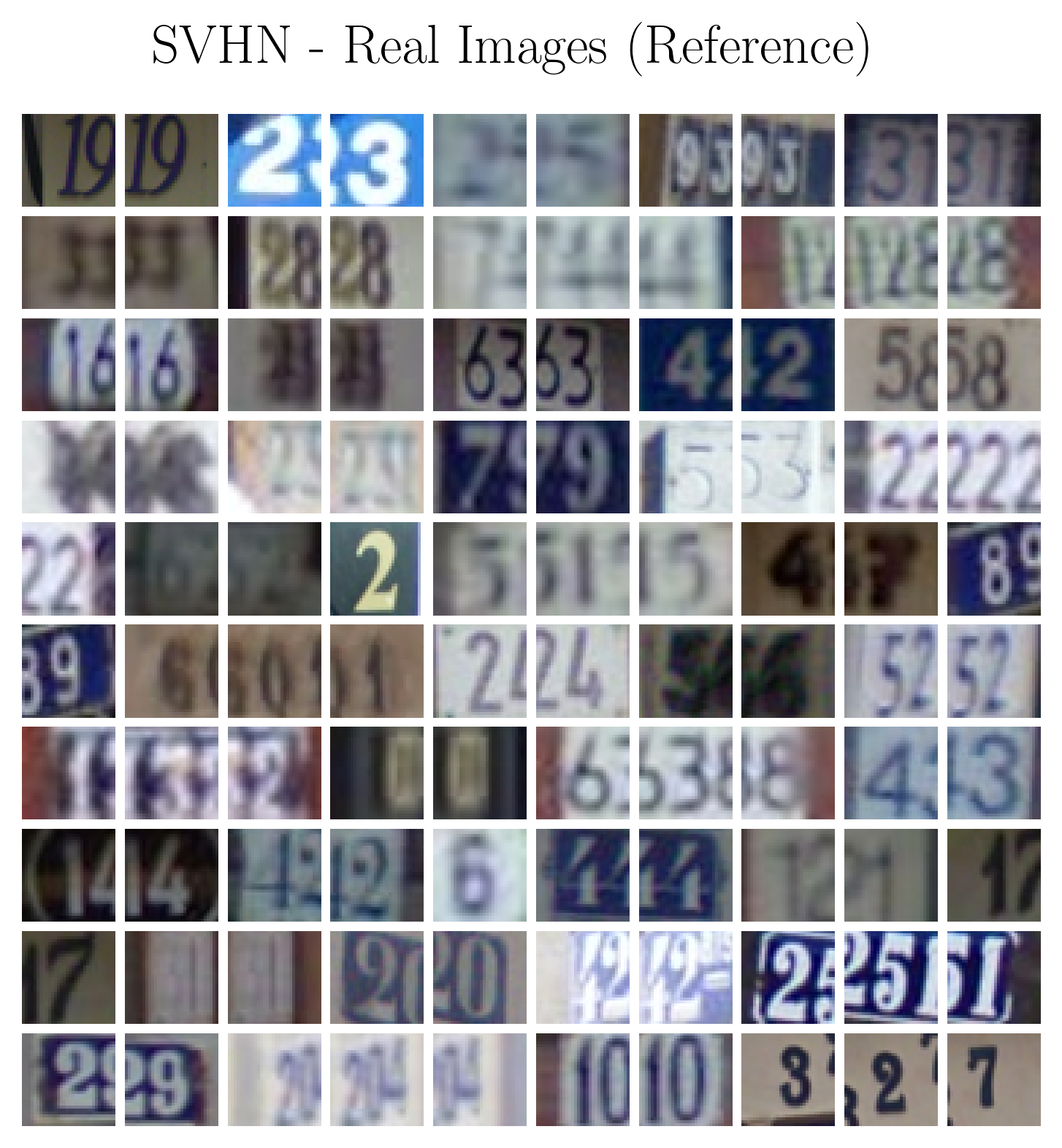}
        \caption{Real samples}
    \end{subfigure}
    \hfill
    \begin{subfigure}{0.32\textwidth}
        \centering
        \includegraphics[width=\linewidth]{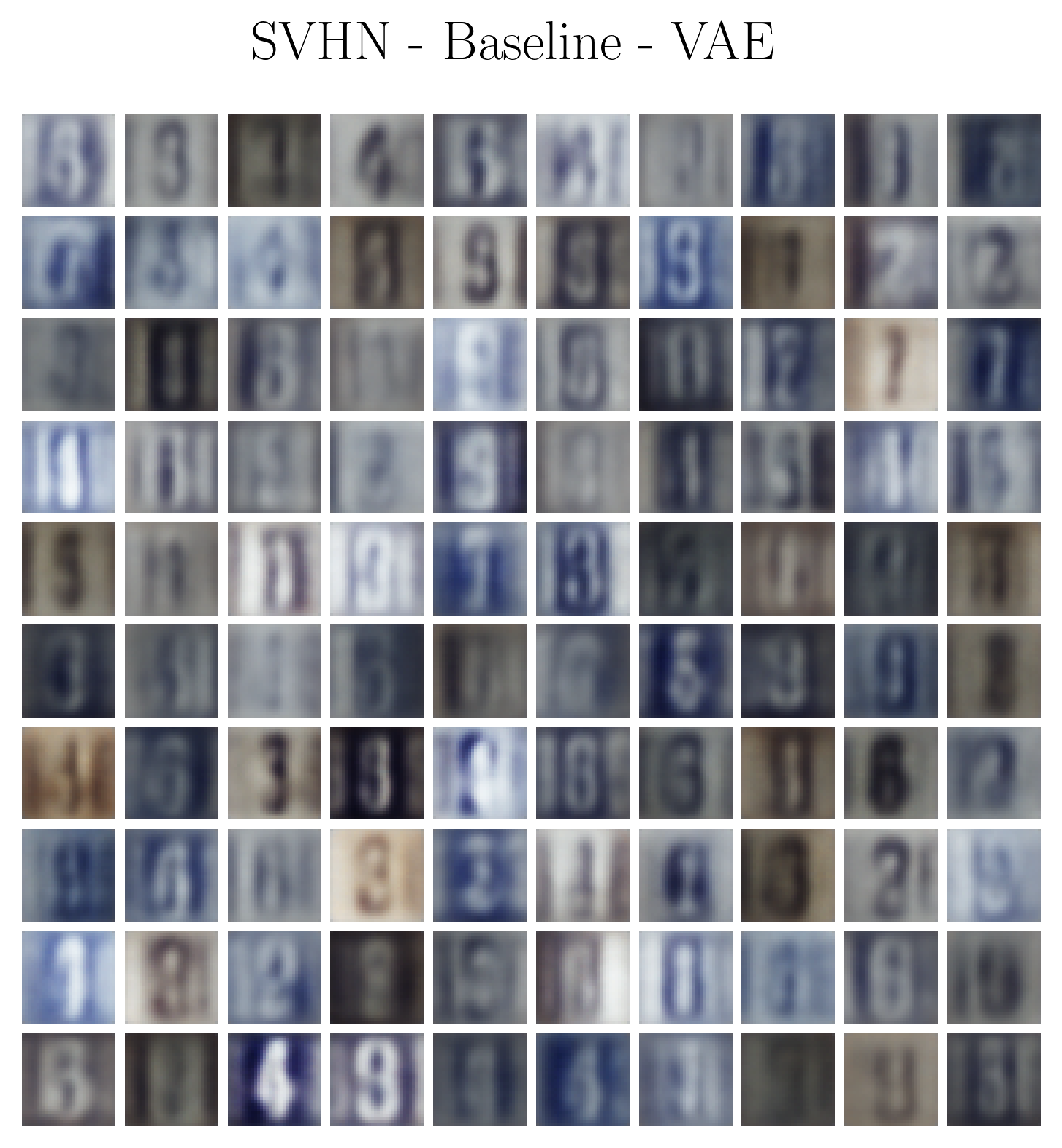}
        \caption{VAE baseline}
    \end{subfigure}
    \hfill
    \begin{subfigure}{0.32\textwidth}
        \centering
        \includegraphics[width=\linewidth]{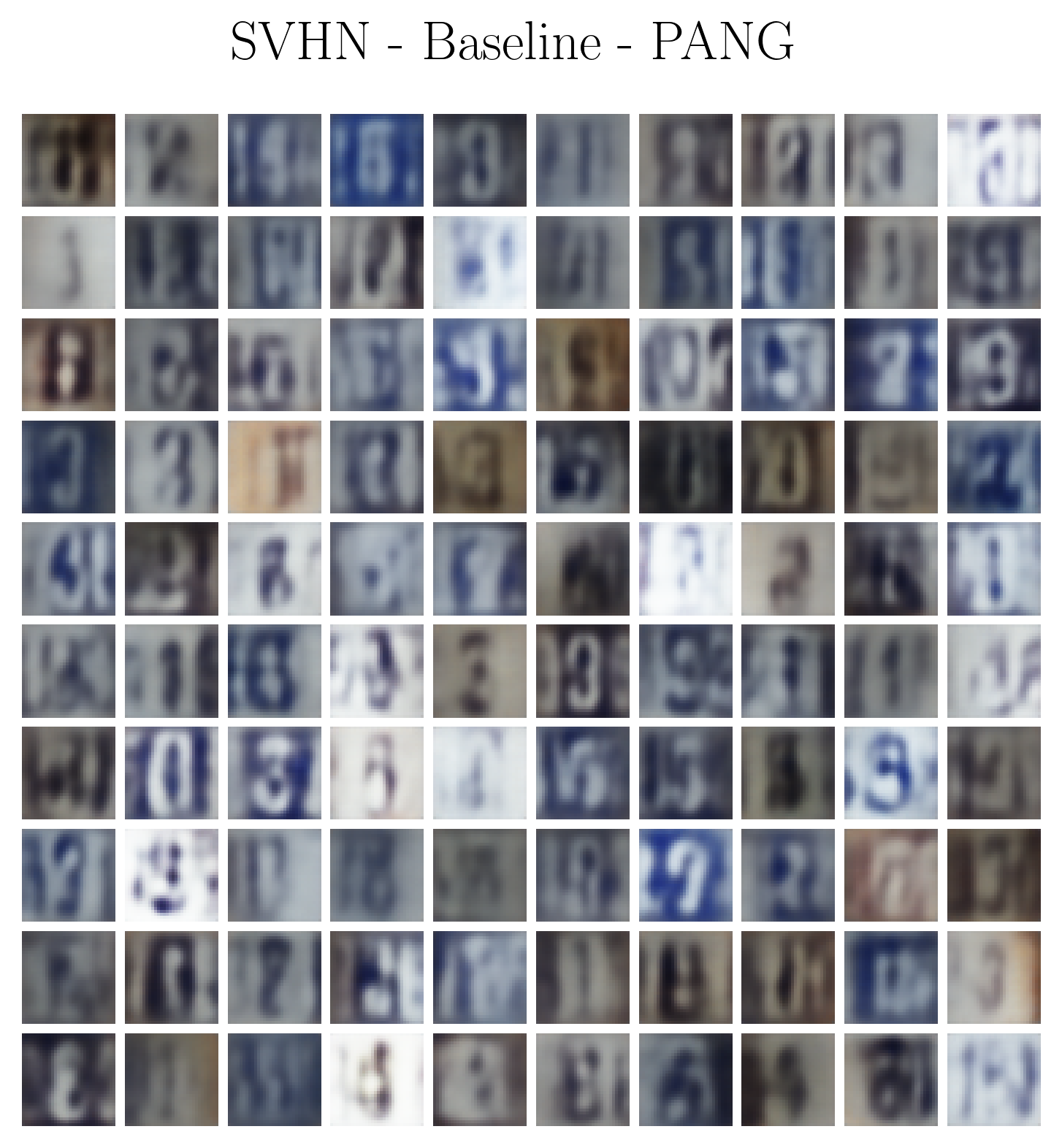}
        \caption{Neural latent EBM baseline}
    \end{subfigure}

    \vspace{0.5cm}

    \begin{subfigure}{0.32\textwidth}
        \centering
        \includegraphics[width=\linewidth]{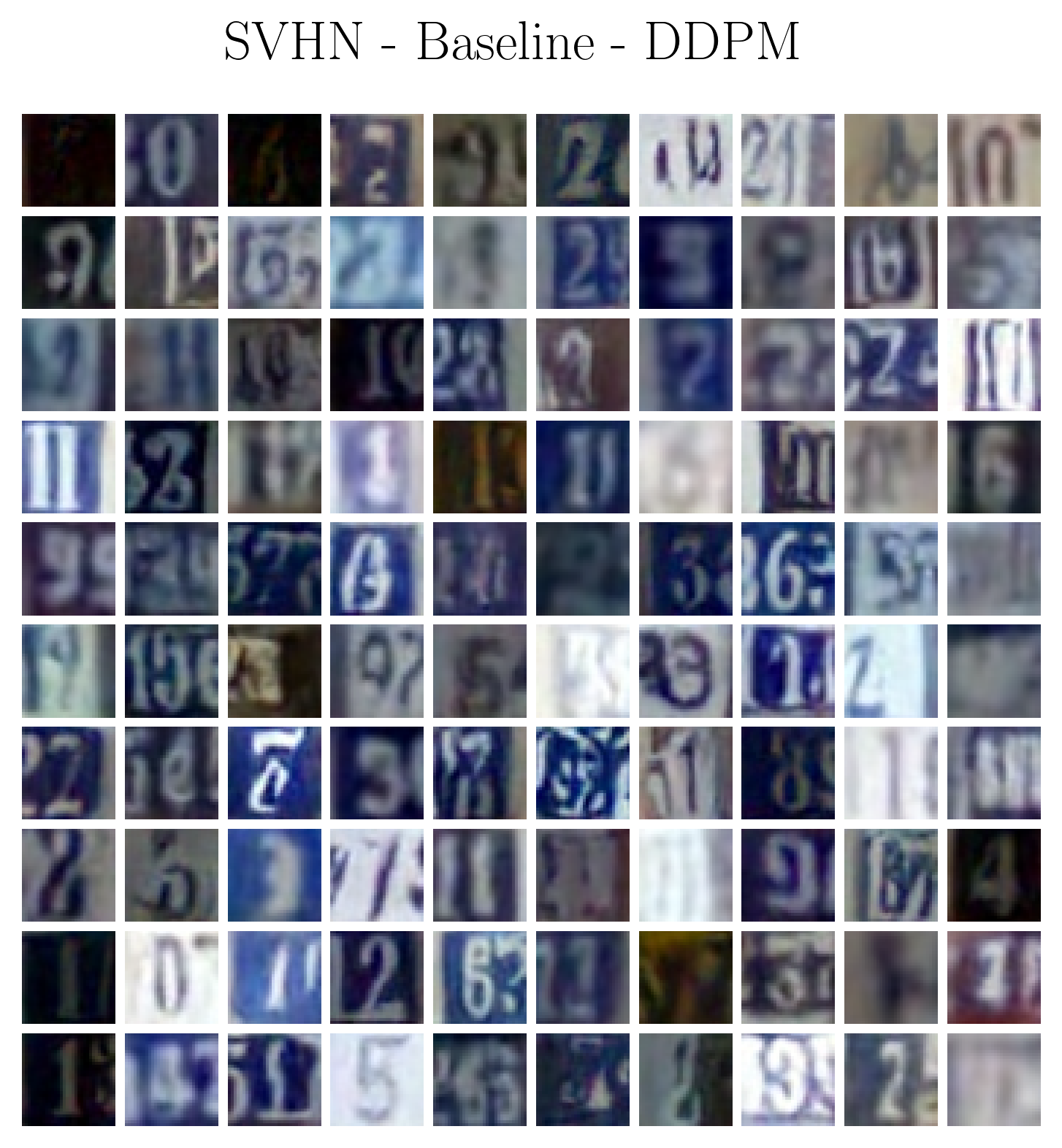}
        \caption{DDPM baseline}
    \end{subfigure}
    \hfill
    \begin{subfigure}{0.32\textwidth}
        \centering
        \includegraphics[width=\linewidth]{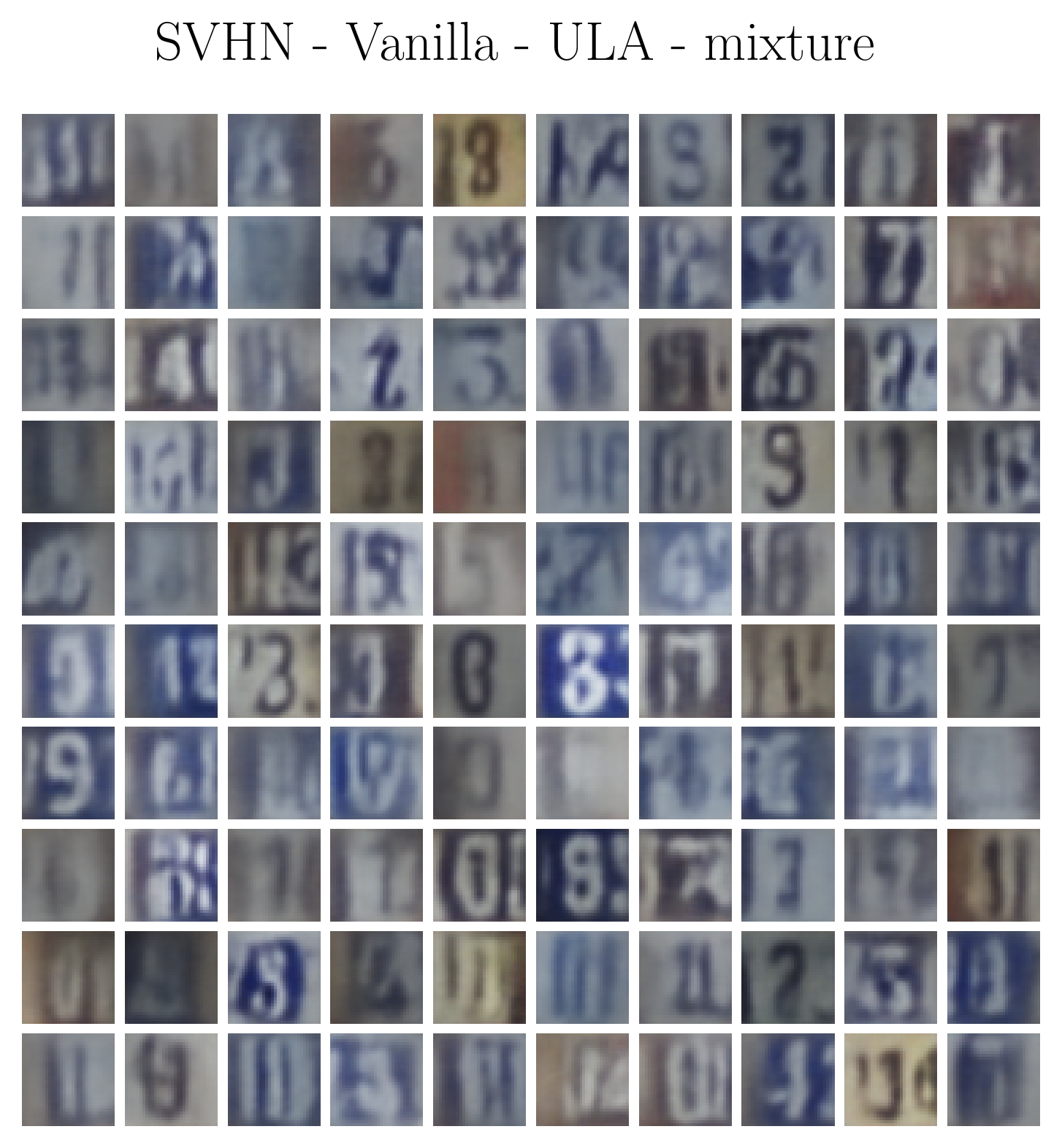}
        \caption{KAEM (MLE)}
    \end{subfigure}
    \hfill
    \begin{subfigure}{0.32\textwidth}
        \centering
        \includegraphics[width=\linewidth]{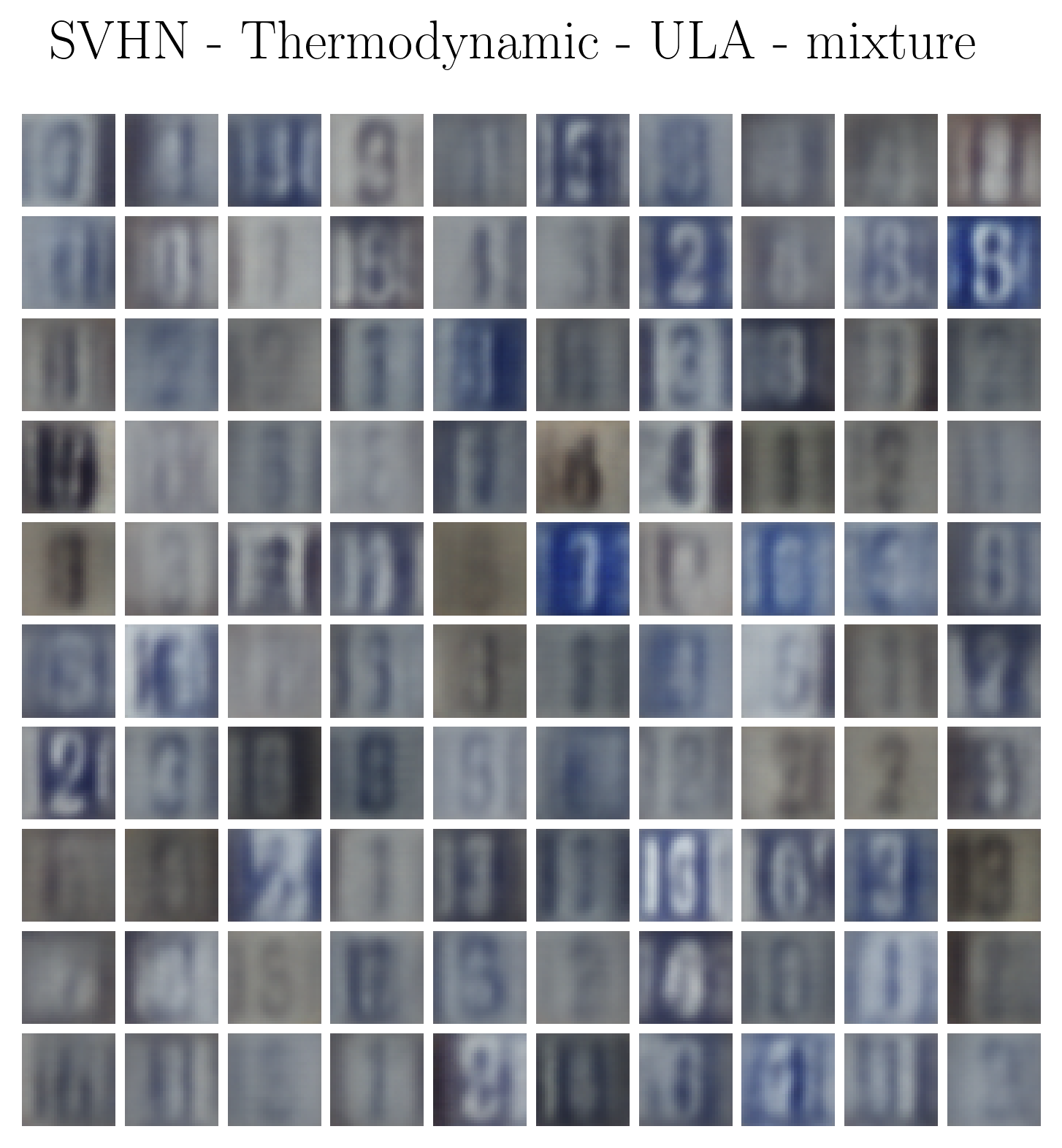}
        \caption{KAEM (Thermo)}
    \end{subfigure}

    \caption{Generated SVHN samples.}
    \label{fig:svhn}
\end{figure}

\begin{figure}[H]
    \centering

    \begin{subfigure}{0.32\textwidth}
        \centering
        \includegraphics[width=\linewidth]{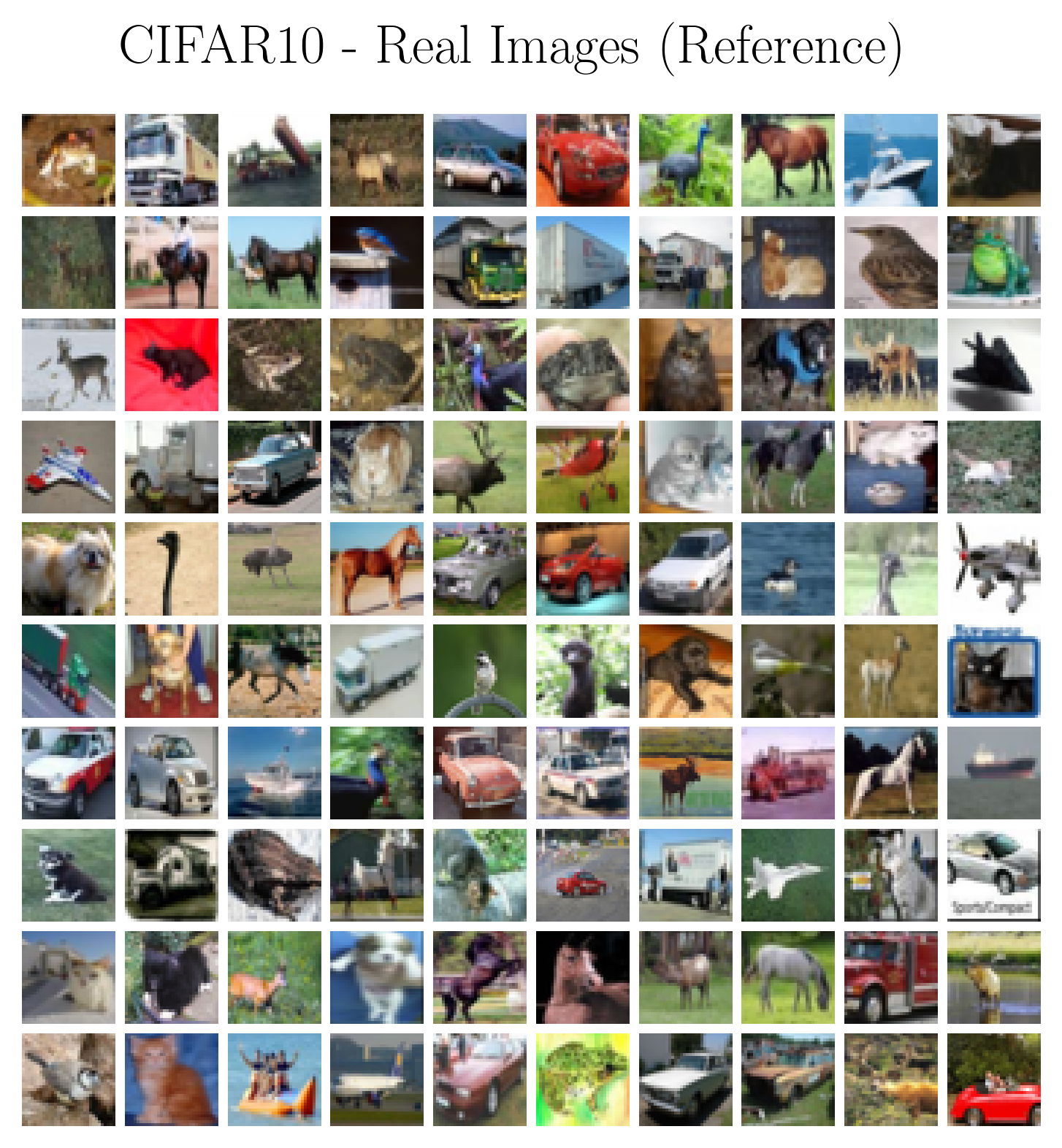}
        \caption{Real samples}
    \end{subfigure}
    \hfill
    \begin{subfigure}{0.32\textwidth}
        \centering
        \includegraphics[width=\linewidth]{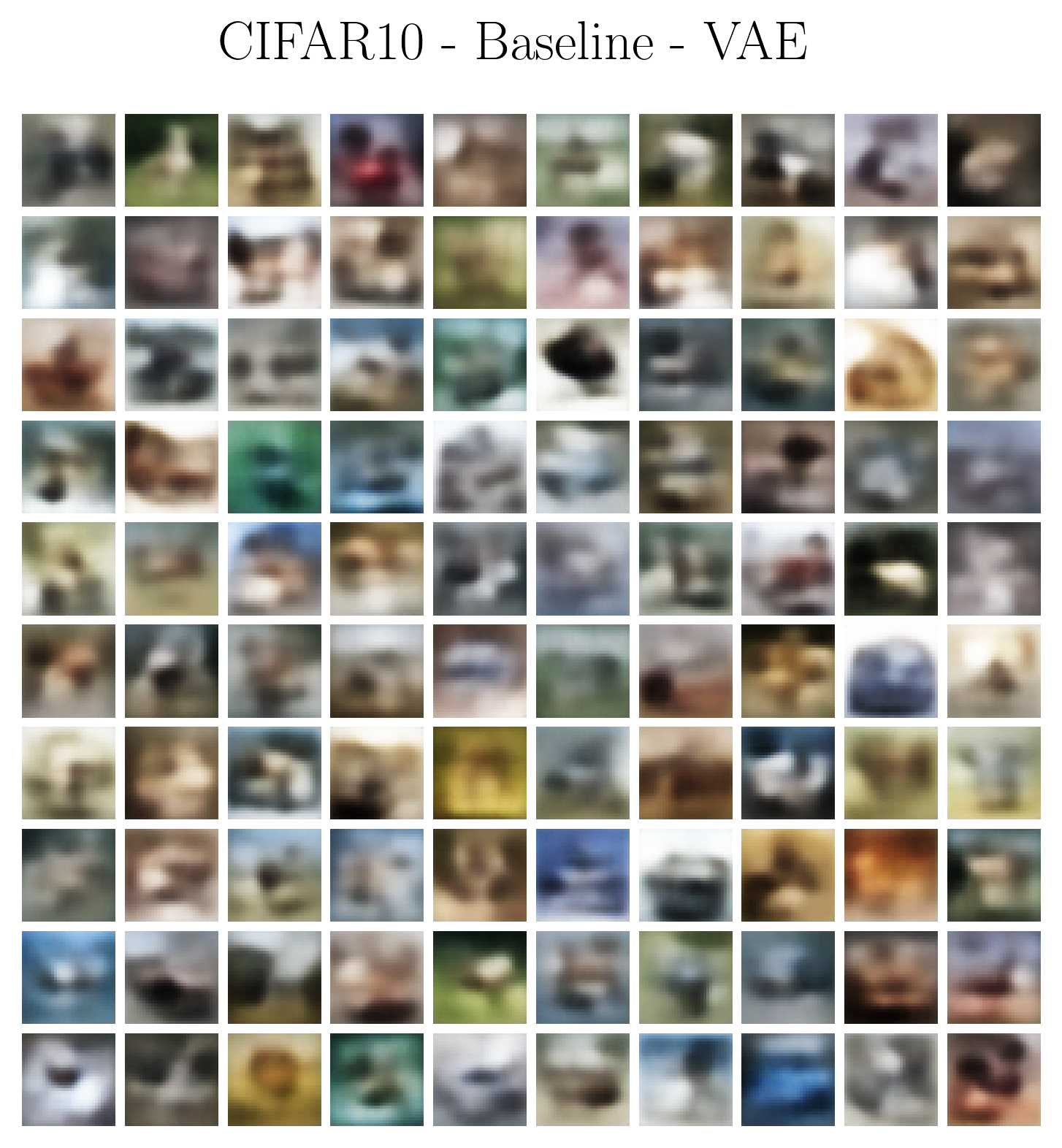}
        \caption{VAE baseline}
    \end{subfigure}
    \hfill
    \begin{subfigure}{0.32\textwidth}
        \centering
        \includegraphics[width=\linewidth]{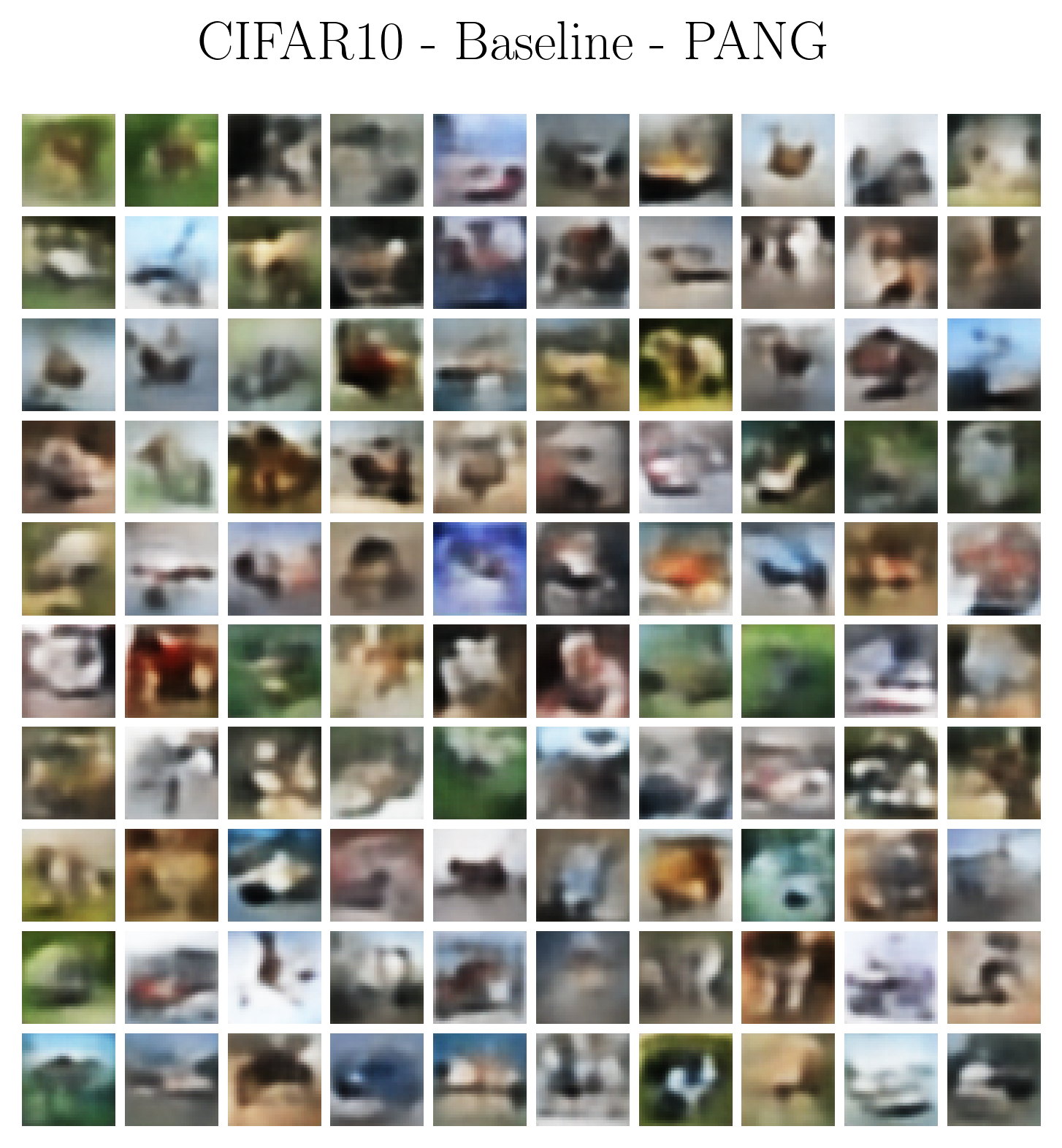}
        \caption{Neural latent EBM baseline}
    \end{subfigure}

    \vspace{0.5cm}

    \begin{subfigure}{0.32\textwidth}
        \centering
        \includegraphics[width=\linewidth]{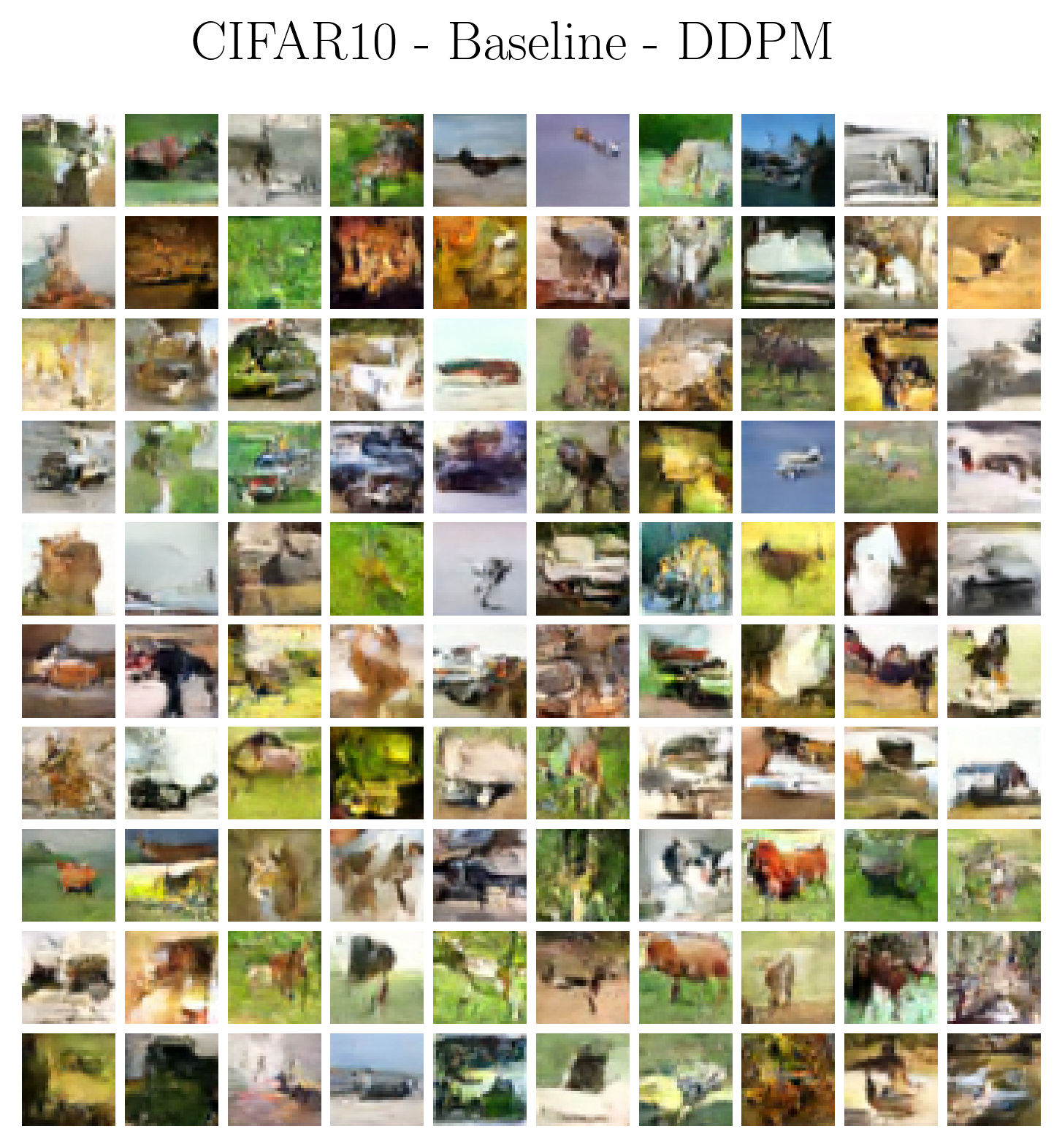}
        \caption{DDPM baseline}
    \end{subfigure}
    \hfill
    \begin{subfigure}{0.32\textwidth}
        \centering
        \includegraphics[width=\linewidth]{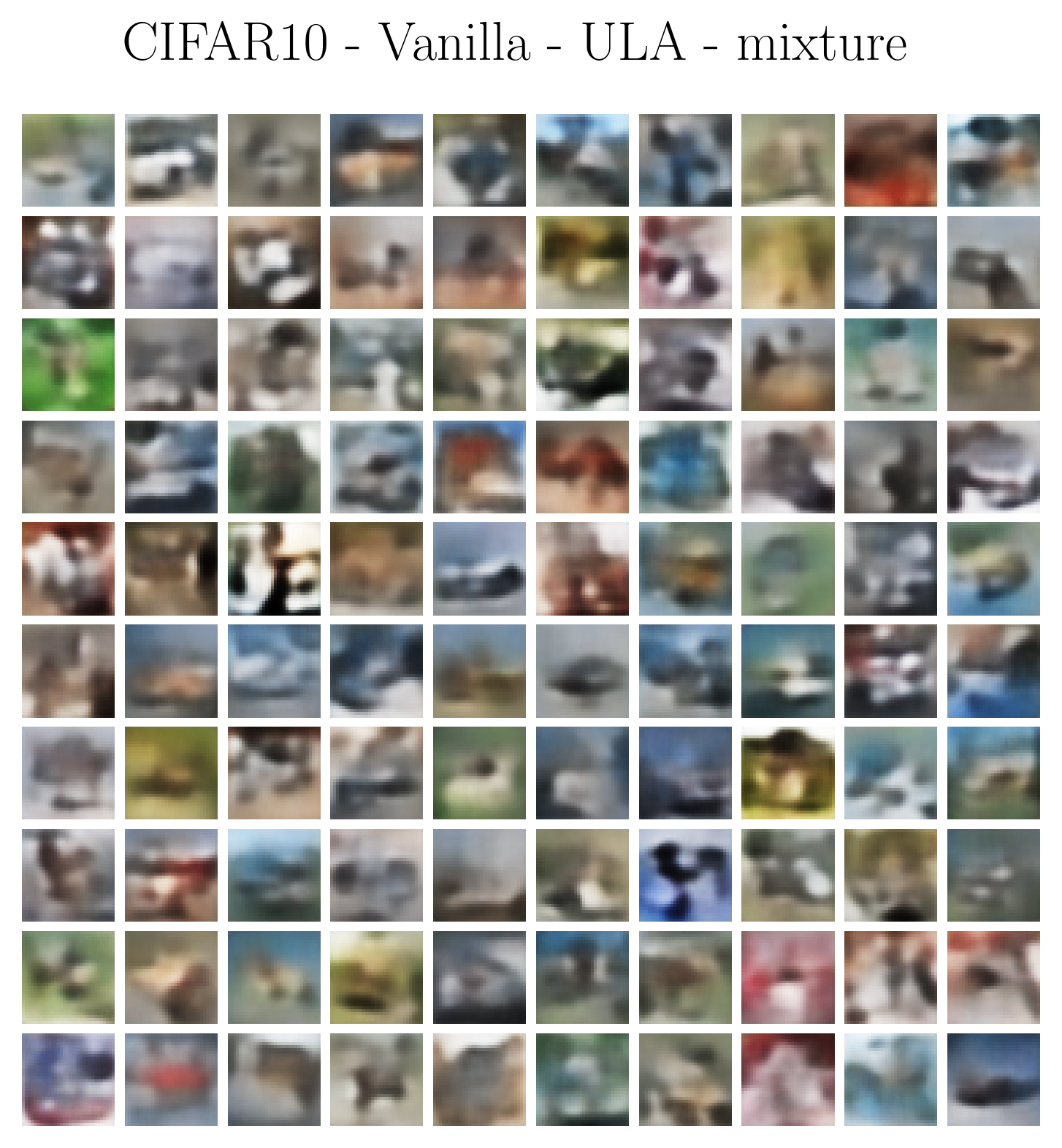}
        \caption{KAEM (MLE)}
    \end{subfigure}
    \hfill
    \begin{subfigure}{0.32\textwidth}
        \centering
        \includegraphics[width=\linewidth]{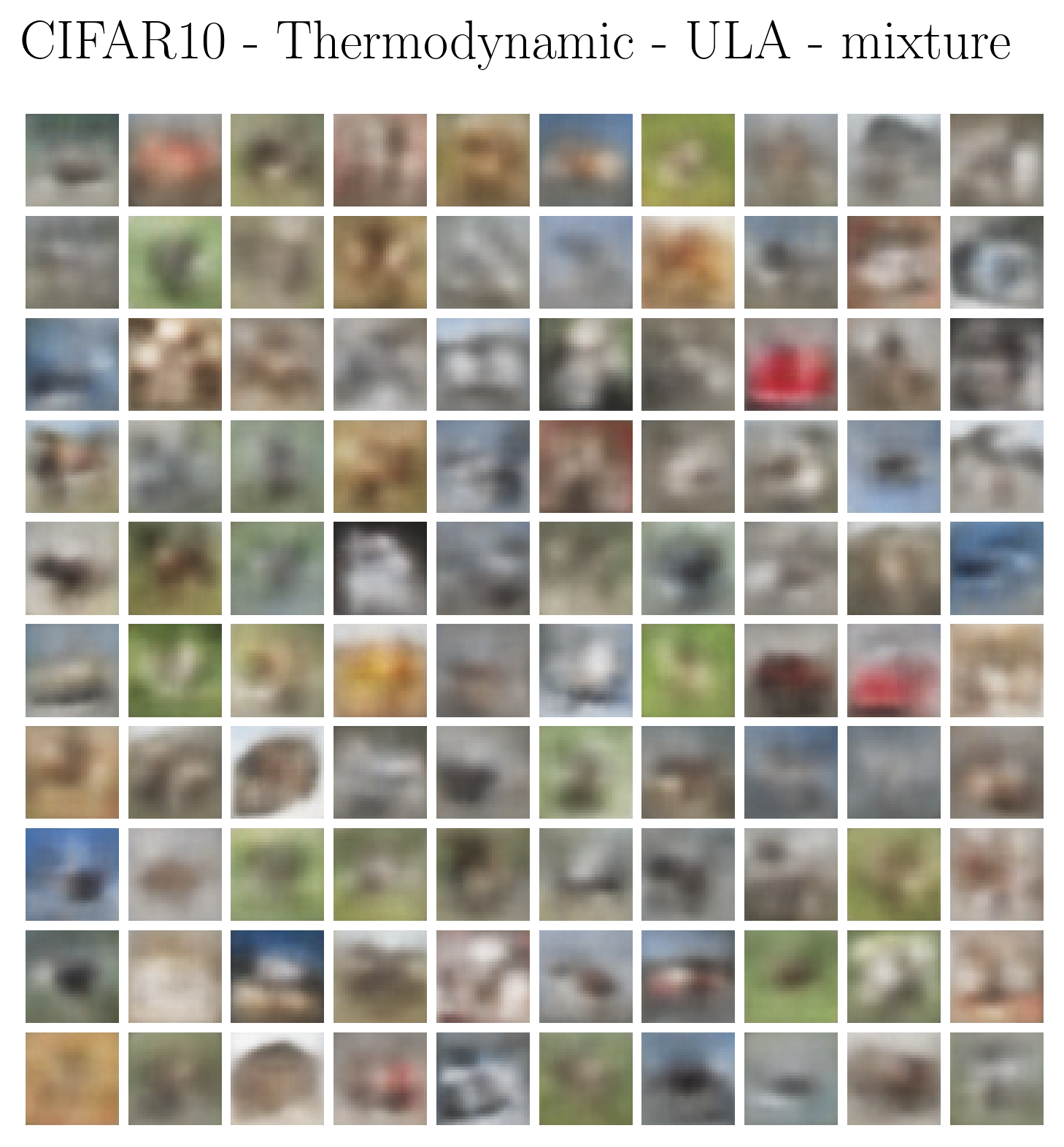}
        \caption{KAEM (Thermo)}
    \end{subfigure}

    \caption{Generated CIFAR10 samples.}
    \label{fig:cifar10}
\end{figure}

\begin{figure}[H]
    \centering

    \begin{subfigure}{0.32\textwidth}
        \centering
        \includegraphics[width=\linewidth]{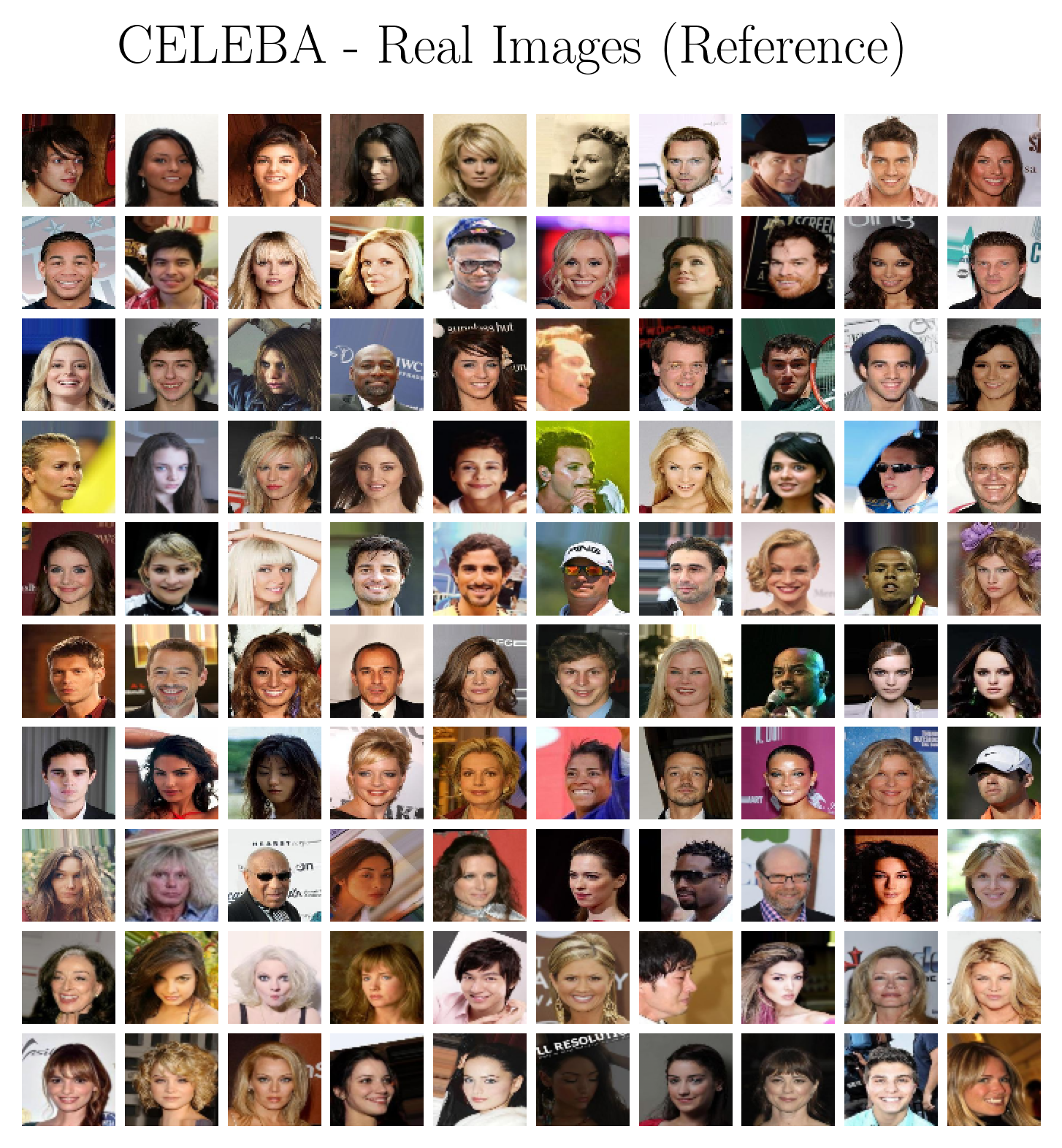}
        \caption{Real samples}
    \end{subfigure}
    \hfill
    \begin{subfigure}{0.32\textwidth}
        \centering
        \includegraphics[width=\linewidth]{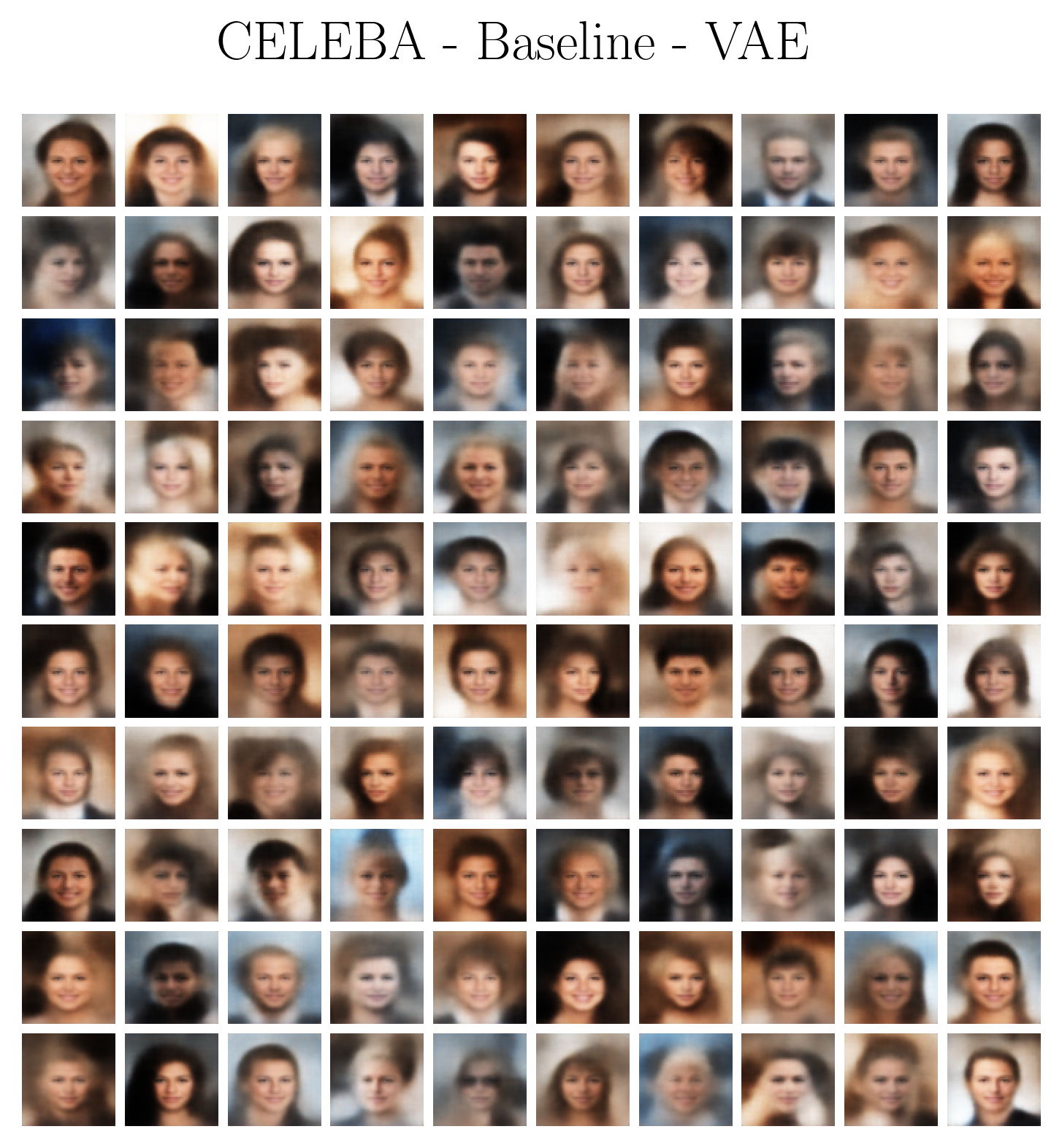}
        \caption{VAE baseline}
    \end{subfigure}
    \hfill
    \begin{subfigure}{0.32\textwidth}
        \centering
        \includegraphics[width=\linewidth]{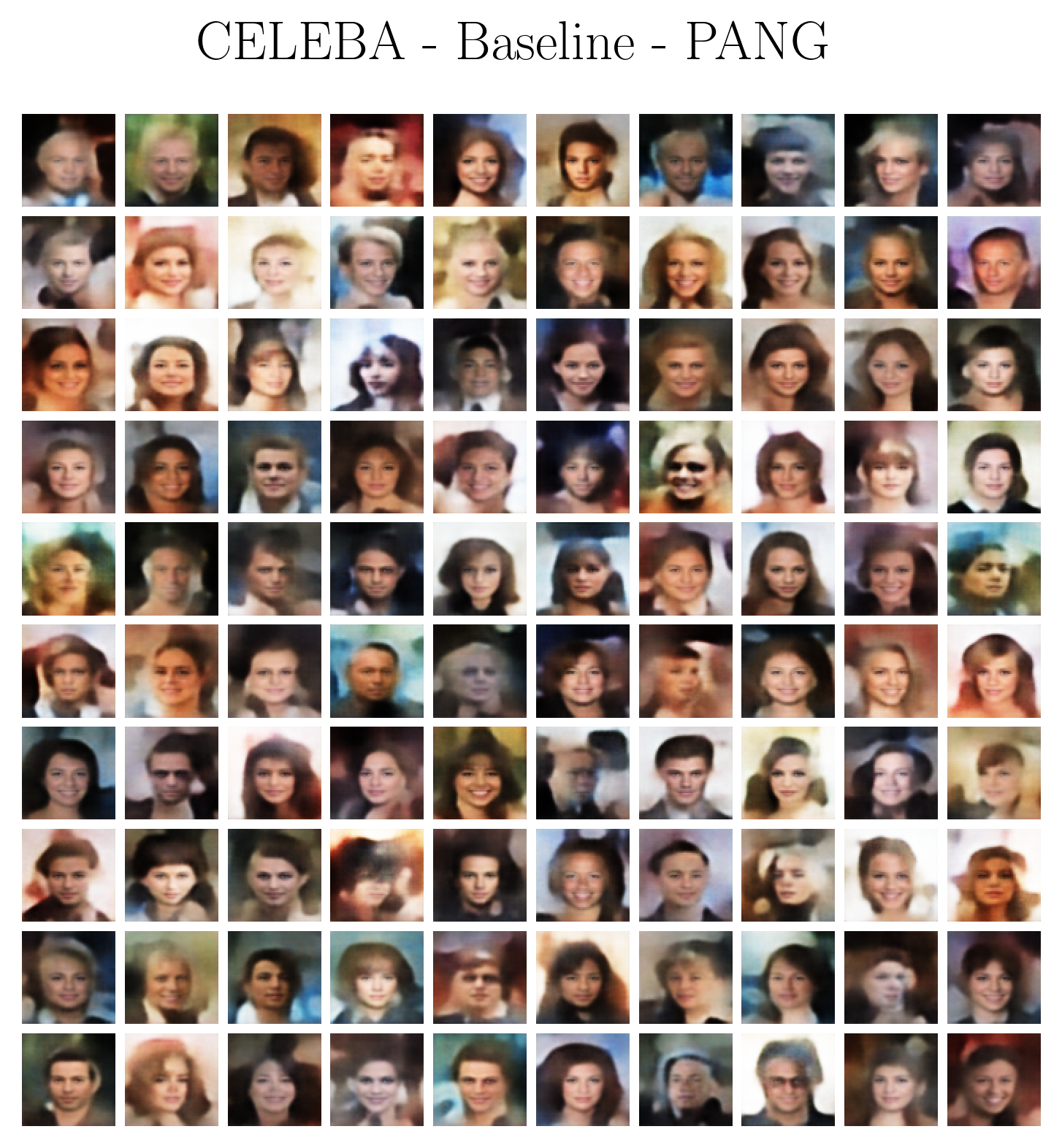}
        \caption{Neural latent EBM baseline}
    \end{subfigure}

    \vspace{0.5cm}

    \begin{subfigure}{0.32\textwidth}
        \centering
        \includegraphics[width=\linewidth]{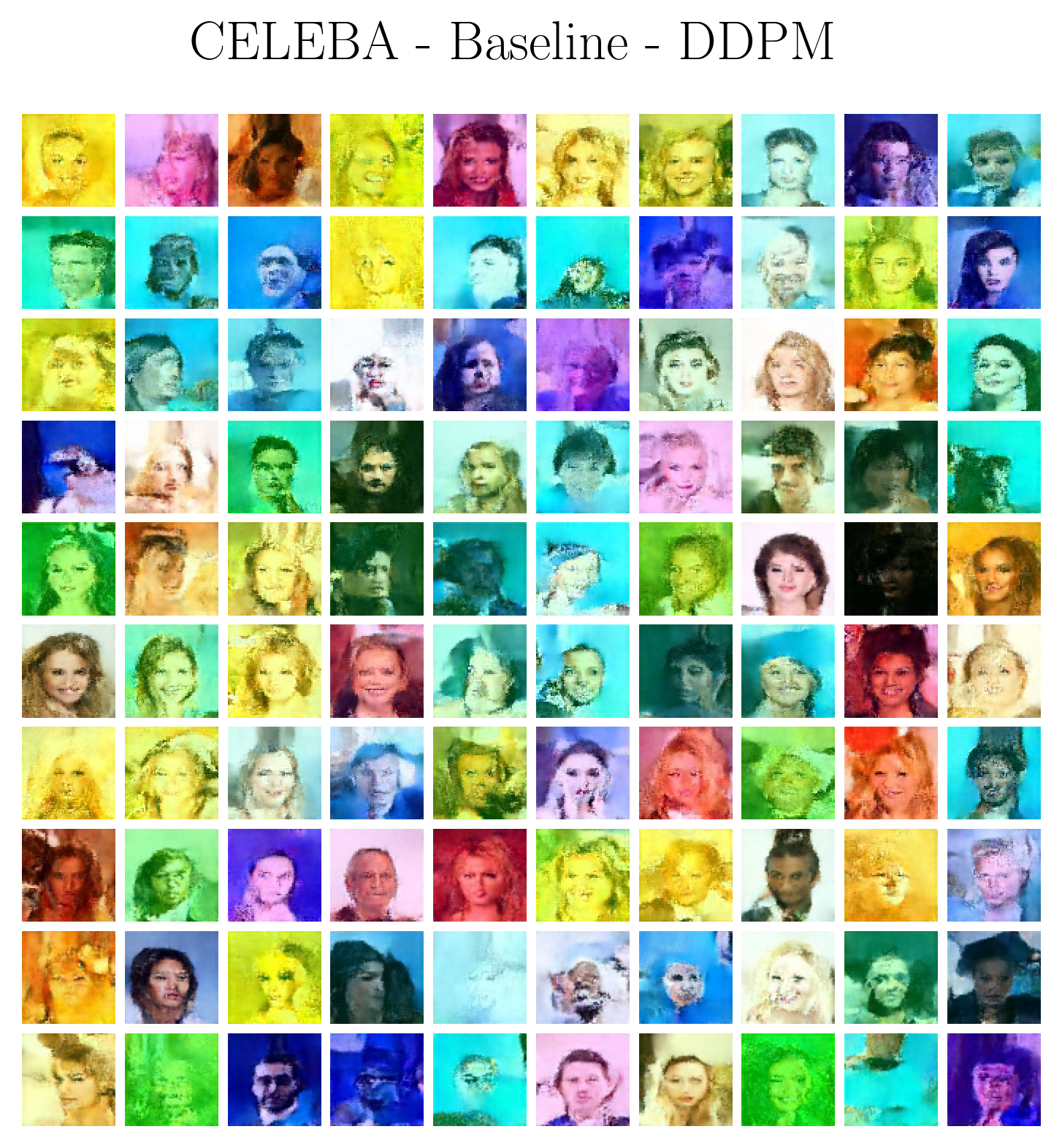}
        \caption{DDPM baseline}
    \end{subfigure}
    \hfill
    \begin{subfigure}{0.32\textwidth}
        \centering
        \includegraphics[width=\linewidth]{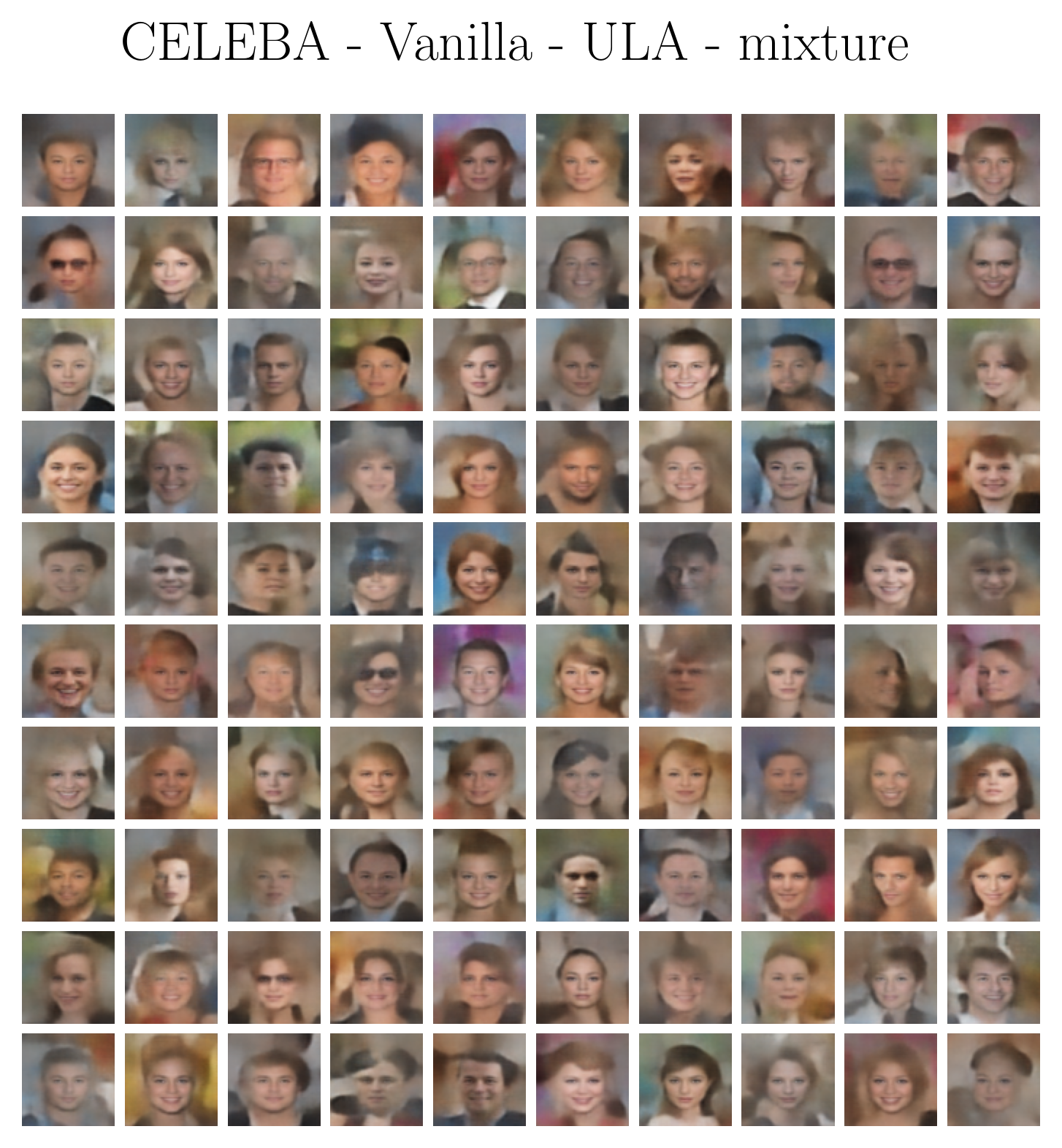}
        \caption{KAEM (MLE)}
    \end{subfigure}
    \hfill
    \begin{subfigure}{0.32\textwidth}
        \centering
        \includegraphics[width=\linewidth]{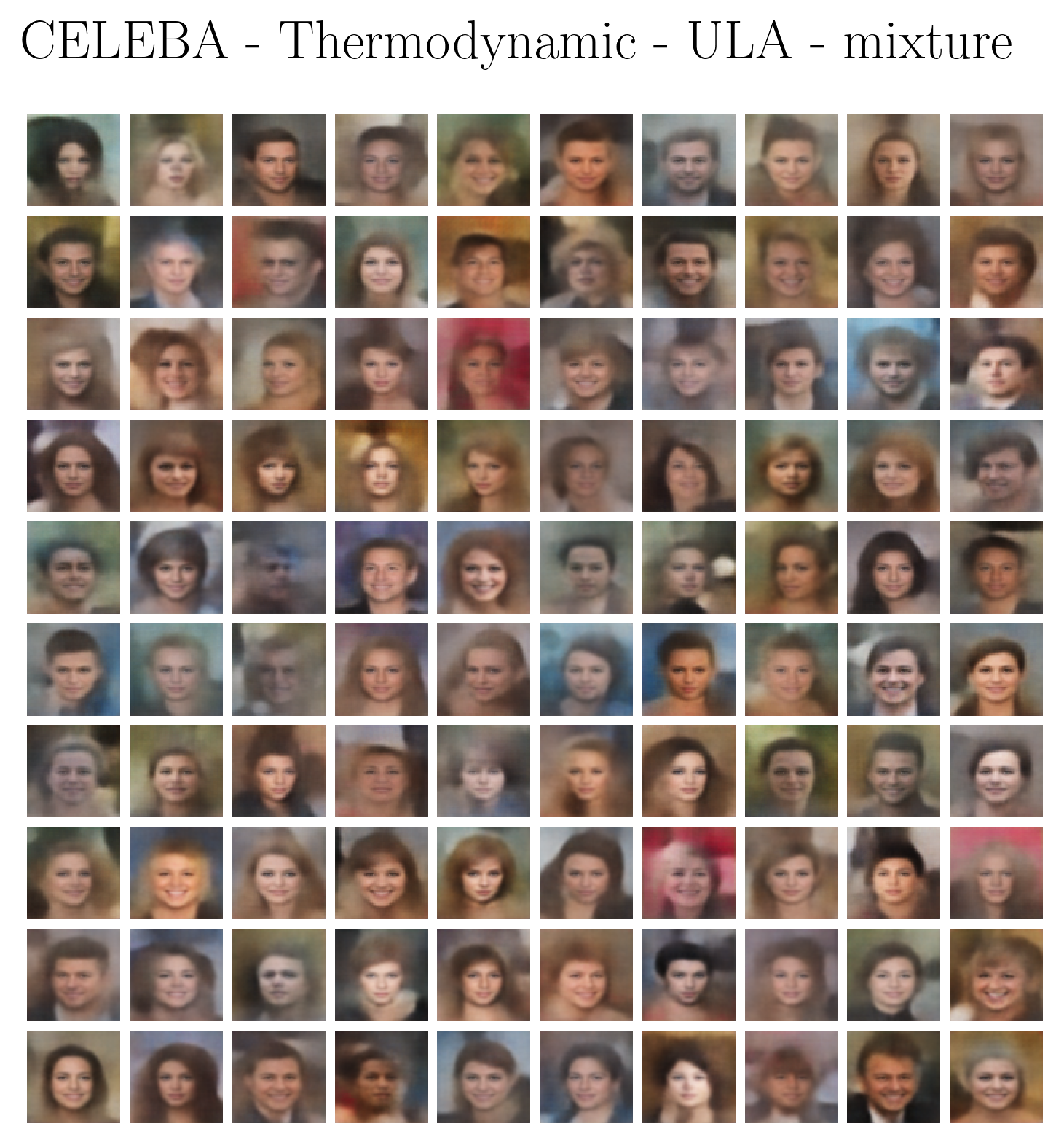}
        \caption{KAEM (Thermo)}
    \end{subfigure}

    \caption{Generated CelebA samples.}
    \label{fig:celeba}
\end{figure}

%
%
%
%
%
%
%

\end{document}